# Subjective Perspectives within Learned Representations Predict High-Impact Innovation

Likun Cao[†], Rui Pan[‡], James Evans[†§]
[†]University of Chicago, [‡]Northwestern University, [§]Santa Fe Institute

**Existing studies of innovation emphasize the power of social structures to shape innovation capacity[1–4]. Emerging machine learning approaches, however, enable us to model innovators' personal perspectives and interpersonal innovation opportunities as a function of their prior experience[5,6]. We theorize and then quantify subjective perspectives and their interaction based on innovator positions within the geometric space of concepts inscribed by dynamic machine-learned language representations[7]. Using data on millions of scientists, inventors, screenplay writers, entrepreneurs, and Wikipedia contributors across their respective creative domains, here we show that measured subjective perspectives predict which ideas individuals and groups will creatively attend to and successfully combine in the future. Across all cases and time periods we examine, when perspective diversity is decomposed as the difference between collaborators' perspectives on their creation, and background diversity as the difference between their experiences, the former consistently anticipates creative achievement while the latter portends its opposite. We analyze a natural experiment and simulate creative collaborations between AI agents designed with various perspective and background diversity[8–10], which support our observational findings. We explore mechanisms underlying these findings and identify how successful collaborators leverage common language to weave together diverse experiences obtained through trajectories of prior work. These perspectives converge and provoke one another to innovate. We examine the significance of these findings for team formation and research policy.**

Innovation has emerged as a focus of research across the social sciences. Economists increasingly characterize it in terms of endogenous growth[11–13], the natural outcome of directed investment. Robust legal frameworks[14,15], supportive institutions[16], diverse and egalitarian team structures[17–19], and efficient information flows[20,21] boost creativity and collaboration, promoting the quantity and quality of new knowledge and the inventions that flow from it. In parallel, a burgeoning psychological literature has examined individual determinants of creativity, ranging from background experiences to micro-behaviors that foster innovation[22–25]. Nevertheless, little is known about how situated innovators perceive inventive opportunities, to which ideas and opportunities they attend, and what they ultimately combine to generate successful innovations. This study has been largely limited by the lack of large-scale data on situated innovator experiences and methods for modeling the complex, high-dimensional distribution of individual perceptions and creative opportunities.

Digital trace and archival data, combined with emergent machine learning methods, enable us to examine the relationship between situated, subjective positions of individual innovators and how those positions condition what innovators can sense and combine in future creative efforts. Moreover, it enables us to examine how innovator subjectivities intersect in more and less successful collaborations across various cultural domains. Here we measure situated subjectivity as a function of individuals' prior creative experience in five domains of innovation: scientific discoveries, patented inventions, entrepreneurial investment, movie screenplays, and Wikipedia articles. We first utilize dynamic word embeddings to construct five conceptual spaces[26] for scientific, technological, financial, cultural, and Wikipedia domains, respectively. We then situate millions of innovators across those spaces based on their prior creative outputs, and characterize their positions and simulated subjective perceptions with respect to each idea in terms of their geometric angle in conceptual space.

This design enables us to observe how innovator-innovation relationships emerge and evolve over time. It also enables us to investigate how diverse subjectivities combine within creative collaborations to generate more or less successful innovations in terms of attention and assessment, which we measure with paper and patent citations, movie ratings, startup fundraising success, and Wikipedia article quality. Our statistical analysis demonstrates that the geometric perspective from innovators' positions within conceptual space consistently predicts their creative movements. Confluences of ideas generate higher innovator uptake and recombination. Innovators are more likely to adopt ideas uniquely visible to them, determined by their angle of perception, and abandon less visible ideas. And when innovators collaborate, their collective success is systematically characterized by decreasing distance between collaborators, but greater diversity in the angle of their perspectives with respect to collaborative outputs.

Across the cultural domains we study, innovations can be arranged according to whether we study the original artifact (e.g., papers, patents, and Wikipedia articles) or summaries (e.g., film and new venture descriptions), and whether the topic of innovation was selected (Wikipedia articles and new venture descriptions) or more freely designed by the innovators (e.g., films, papers, patents). Collaborative processes and outcomes naturally vary across these domains, but all five allow us to measure team composition, the perspective of team members based on prior experience, machine-learned representation of the focal innovation, and team performance as evaluated by the innovation's primary audience (see Fig. S1). Similar results across these diverse contexts further corroborate our rich representations of subjective perspective and their power in predicting the success of combinations across a range of creative collaborations.

**Idea Movements in Conceptual Space Predict Innovative Opportunities**

We represent the shifting landscape of ideas with dynamic word embeddings. Word embeddings produce dense, continuous models that represent words as high-dimensional vectors to enable the measurement of semantic distances between them[27–29]. Distance between word vectors in embedding space proxies for complex relationships between referenced concepts[26] (see Fig. 1a-b). Substantial research has demonstrated the capacity of learned embeddings to robustly replicate even subtle semantic relationships, including analogies and metaphor[28], which trace empirical shifts in society and culture[30–32]. As such, embeddings represent a powerful instrument from which to measure changing cultural associations[33–35]. The newest large language model representations layer multiple, contextual embeddings and can be even more semantically precise, but they rely on imperfectly dated web-scale data and cannot yet be estimated across dynamic periods in the past without containing shadows of the future[36,37].

In this work, we employ dynamic word embeddings to naturally account for changes in concept relationships over historical time, allowing for comparison between periods[38]. We utilize large-scale corpora to represent the space of concepts associated with cultural production in distinct innovation domains: science, technology, Wikipedia, film, and entrepreneurial finance. We use (1) Microsoft Academic Graph to build a dynamic scientific space within which we project articles and conference papers; (2) the US Patent & Trademark database to build an evolving technology space within which we project patented inventions; (3) ProQuest business news to build a changing space of business discourse that forms the background for new venture creation, within which we project Crunchbase descriptions of new companies; (4) the Google Books' fiction and nonfiction corpora to build a dynamic literary space that forms the background for film production, within which we project IMDB film summaries; and (5) English-language Wikipedia to create an encyclopedic knowledge space within which we project individual Wikipedia articles.

In each temporal word embedding space, we apply density-peak clustering[39] to track the location and movement of each semantic cluster. If a subspace has more in-flowing, convergent clusters and fewer out-flowing, divergent clusters, we hypothesize that the area will support more creative conflict, leading to more innovation. In Extended Data Fig. 1, we show that in all five domains, the subspaces where fresh ideas flow host a substantially higher number of new innovations, with correlation between concept in-flow and novelty emergence ranging between 0.11 in the case of entrepreneurial finance and 0.34 in Wikipedia article creation (all $p$<0.001). This relationship stays robust with alternative parameter settings (Figs. S2-S6), and passes Granger causality tests (see Table S1). The most prosperous subfields are located on the frontier of current knowledge, as shown at the boundaries of the space, as depicted in Fig. 1c.

We then position innovators within conceptual landscapes as a function of their prior work experience to simulate their subjective perspective and predict how they view, move, and collaboratively create new cultural products. Fig. 1 illustrates how we decompose and categorize the diversity of team perspectives and their relationship to innovative success. We assess the *perspective diversity* of collaborators as the average (angular) difference between their individual perspectives on their creative products $\bar{\theta}_p$. Such diversity may lead to conflict, complementarity, and collective creativity. We validate perspective diversity by showing that innovators who hold similar perspectives across their projects tend to make similar contributions (see SI2.4 and Fig. S7). We then assess collaborators' *background diversity* as the average cosine distance between their past experiences $\bar{\theta}_b$. Greater background diversity may decrease collaborators' ability to effectively communicate with one another and limit their creative capacity. We validate background diversity by measuring a strong and significant positive correlation (Spearman's $\rho$= 0.7, $p$<0.001) between innovators' unshared past experiences and their background diversity (see SI2.4 and Fig. S7). In Fig. 1e, we see how perspective and background diversity naturally categorize teams into high/low on both, or high on one and low on the other. Teams across science, technology, entrepreneurship, film, and Wikipedia exhibit a broader perspective diversity distribution compared to a relatively narrow background diversity distribution (Fig. 1f). This suggests that innovators with backgrounds too far apart fail to launch collaborations.

**Results**

When we position innovators within conceptual landscapes based on prior experience, their subjective perspectives predict the likelihood of adopting new concepts in future work. We assess the subjective salience of ideas as a function of increasing idea proximity and the degree to which idea movement is visible when projected onto scholars' personal horizons. Subjective visibility increases the likelihood that an innovator adopts a new idea as it moves closer, with the estimated rate of increase ranging from 3% in movies to 154% in scientific publications (regression model $\beta$s capturing the effect of proximity $\times$ angle on innovator adoption of semantic concepts, with full specification in SI Table S2: $\beta$ = 0.024, $p$ = 0.002, $df$ = 6.687$\times 10^7$ for scientists, $\beta$ = 0.013, $p$ = 0.022, $df$ = 1.150$\times 10^8$ for inventors, $\beta$ = 0.019, $p$ = 0.0015, $df$ = 1.003 $\times 10^7$ for startup investors; film and Wikipedia cases positive but non-significant; also see Extended Data Fig. 2).

Subjective position is even more important in predicting the success of creative collaborations. We measure success through the innovation's primary audience—citations for papers and patents, fundraising for startups, ratings for movies, and assessed quality for Wikipedia articles. But while success is measured differently across domains, the effect is the same. Perspective diversity positively and substantially anticipates the likelihood of creative success in all areas for all time periods investigated. Diverse perspectives, however, are counterbalanced by fragmented experience. As shown in Fig. 2 and Extended Data Fig. 3, higher perspective diversity (*pd*) and lower background diversity (*bd*) are significant and substantial factors in anticipating performance success across creative contexts (regression model $\beta$s with full specifications in SI Tables S3 and S4: science, $\beta_{pd}$= 0.504, $p<10^{-4}$, $\beta_{bd}$ = -1.107, $p<10^{-6}$, $df$ = 2.041$\times 10^7$; technology, $\beta_{pd}$= 0.099, $p<0.01$, $\beta_{bd}$= -0.916, $p<10^{-10}$, $df$ = 2.525$\times 10^6$; entrepreneurship, $\beta_{pd}$= 0.438, $p<10^{-9}$, $\beta_{bd}$= -1.268, $p<10^{-54}$, $df$ = 1.017$\times 10^4$; film, $\beta_{pd}$= 1.081, $p<10^{-28}$, $\beta_{bd}$ = -1.711, $p<10^{-14}$, $df$ = 2.977$\times 10^4$; and Wikipedia, $\beta_{pd}$= 0.603, $p<10^{-1024}$, $\beta_{bd}$ = -2.139, $p<10^{-1024}$, $df$ = 5.058$\times 10^6$). These effects persist when we control for knowledge domain and year (see Tables S3 and S4), team size (see Extended Data Fig. 4), and the achievement of the same innovator when changing teams (see Extended Data Table 1).

Next, we explore mechanisms underlying these observational findings. To gain insight into team collaboration processes, we analyze self-reported contribution statements from 89,575 papers published in the multidisciplinary journals *Science*, *Nature*, *PNAS*, and *PLOS ONE*. We cluster research roles expressed in contribution statements according to the researchers who performed them (see SI2.7). In this scheme, roles co-expressed by many shared researchers are likely to cluster together. Clustered research roles cleave unambiguously into leading the research or directly or indirectly supporting it (see Fig. 3a)[18]. We examine how an individual's contribution to team diversity is associated with their role in the collaboration. Results show that when individuals bring new perspectives to a team, they are more likely to take on central, leadership roles such as "conceive", "initiate", or "direct" the research (regression model $\beta$s with full specifications in SI 2.10 and results in Extended Data Table 2: $\beta_{mpd}$ = 0.094 where *mpd* indicates the marginal perspective diversity contributed, $p<10^{-20}$, $df$ = 5.236$\times 10^5$). In contrast, those who contribute to background diversity without introducing new perspectives are more likely to take on only marginal tasks, such as "discuss" or "edit" the manuscript" ($\beta_{mbd}$ = 0.033 where *mbd* indicates the marginal background diversity contributed, $p<10^{-60}$, $df$ = 5.236$\times 10^5$) (see Fig. 3b, Extended Data Table 2 and Fig. 7). This pattern remains consistent even after controlling for author fixed effects (lead role model, $\beta_{mpd}$ = 0.040, $P<10^{-14}$; indirect support model, $\beta_{mbd}$ = 0.025, $p<10^{-16}$, both $df$ = 2.530$\times 10^5$), suggesting that the same individual may be assigned different responsibilities depending on the unique value they contribute to each collaboration (see Supplementary Tables S5, and Extended Data Fig. 7-8).

Following this logic, teams with high perspective diversity may benefit from more active and engaged participation by each member, leading to better integration of individual expertise. We test this hypothesis using a knowledge integration index as a mediator of success. We construct our integration index from taxonomies uniquely designed for each of the five datasets (see Methods and Supplementary Fig. S8). Fixed-effects regression models reveal how teams with diverse perspectives but not overly diverse experiences succeed in part by focusing on shared expertise across members' prior work, as shown in Fig. 3c. Teams with high perspective and low background diversity mobilize their overlapping understanding in the new, collaborative project (with $\beta_{pd} = 0.172$, $p<10^{-23}$, and $\beta_{bd} = -0.634$, $p<10^{-518}$, $df = 2.041\times10^{7}$ for science, and $\beta_{pd} = 0.280$, $p<10^{-142}$, and $\beta_{bd} = -0.902$, $p<10^{-308}$, $df = 2.525\times10^{6}$ for technology, for example; see Fig. 3c and Supplementary Table S6), which is less likely to stray and speculate in new domains that none, one or only a few members have previous experience (with $\beta_{pd} = -0.294$, $p<10^{-26}$, $\beta_{bd} = 0.761$, $p<10^{-79}$, $df = 2.041\times10^{7}$ for science, and $\beta_{pd} = -0.313$, $p<10^{-95}$, $\beta_{bd} = 0.930$, $p<10^{-154}$, $df = 2.525\times10^{6}$ for technology; see Fig. 3d and Supplementary Table S7).

To identify whether the pattern of different perspectives but similar backgrounds and the integration mechanisms observed above were robust to selection or success bias, we (1) evaluated a natural experiment on Wikipedia, and (2) ran an AI agent-based simulation of creative collaboration in science. The natural experiment involved Wikipedia contributions during periods in which editors self-selected to contribute to Eastern European pages versus campaigns in which banners broadly recruited contributors from across Wikipedia and the web to edit the same subjects. Results suggest a causal effect of perspective and background diversity on page quality ($\beta_{pd} = 0.136$, $p = 0.026$, $df = 924$; $\beta_{bd} = -0.188$, $p = 0.002$, $df = 826$; see Extended Data Fig. 5).

We also simulated creative collaborations between Large Language Model (LLM) agents designed with distinct configurations of perspective and background diversity[8–10]. Specifically, we constructed 720 Large Language Model (LLM) agents by simulating specific human scientists into 240 three-agent teams configured across all four binarized combinations of perspective and background diversity (i.e., high-high, high-low, low-high, low-low). Figs. 4a and b graphs histograms that demonstrate, as with our observational data and the Wikipedia natural experiment, simulated scientific collaborations with high perspective and low background diversity consistently performed better in writing quality and scientific significance (e.g., originality, importance) as assessed by advanced reasoning models, which we validated against human reviewers (see SI4.2.5, Table S14). These simulation results replicate using LLMs of different architectures and parameter sizes (essays generated by these "optimal" teams in Gemma have a 95% bootstrapped CI for scientific quality of [7.695, 8.133], nonoverlapping with the next highest CI of [7.181,7.676], and a 95% CI for writing score of [7.610, 8.076], overlapping but higher than the next highest CI of [7.571, 7.914]; similar patterns of effect for Mistral-NeMo and QwQ-based experiments; see Fig. 4, Extended Data Fig. 6, Fig. S11, and Table S15).

The integration and speculation dynamics, which we observed in all empirical contexts, also emerge organically in our LLM-based team simulations. We analyze how information from each agent's assigned background becomes incorporated into their iterative essay drafts. Results are presented in Fig. 4c-e. For all three models, teams with high perspective and low background diversity predictably manifest the highest knowledge integration (with 95% CI of [0.224, 0.259] for Gemma, nonoverlapping with the next highest CI [0.166, 0.187], and identical patterns of effect for Mistral-NeMo and QwQ-based experiments; see Fig. 4c, SI4.2.6 and Table S15), and teams with high background diversity demonstrate the highest speculation (with 95% CIs for those with low perspective diversity of [0.718, 0.743] and with high perspective diversity of [0.714, 0.737], nonoverlapping with other conditions for Gemma, and identical patterns of effect for Mistral-NeMo and QwQ-based experiments; see Fig. 4d, SI S4.2.6 and Table S15). We finally explored directly whether teams fix and focus or drift and speculate within simulated teams by measuring the distance between each paper and its assigned focal topic, something unmeasurable for observed teams. As graphed in Fig. 4e, teams with high perspective and low background diversity post the smallest distance from the assigned topic and the greatest focus on their collective expertise (with 95% CI of [0.224, 0.250] for Gemma with the next lowest of high perspective and background diversity barely overlapping [0.244, 0.275], and even stronger, nonoverlapping patterns for Mistral-NeMo and QwQ-based experiments, see Fig. 4e, SI4.2.6 and Table S15). By contrast, teams with low perspective diversity show the highest distance to their topic, and the greatest drift (with 95% CIs of [0.311, 0.392] for those with low background and [0.284, 0.336] for those with high in Gemma, nonoverlapping with other CIs; see Fig. 4e, SI4.2.6 and Table S15).

These observed and simulated findings suggest a team design principle: collaborators bring together plumes of diverse experience that spark inspiration through convergence and focused conflict to drive successful innovations. Nevertheless, innovators whose experiences are too far from one another fail to successfully communicate and collaborate.

**Discussion**

Our analysis demonstrates how machine-learned representations enable the precise measurement of a subjective human perspective, based on prior creative experience. These "objective subjectivities" are predictive of how individual scientists, inventors, entrepreneurs, writers, and Wikipedia editors adopt and abandon new ideas. More powerfully, measured subjectivities reveal novel and consistent patterns of success in creative collaborations. Perhaps most strikingly, they enable the instantiation and simulation of teams with these properties, which exhibit the same distribution of successful outcomes.

Our findings about creative teams reveal that innovation success fundamentally depends on the interaction between team members' diverse subjective perspectives and their shared background experiences. Across scientific discovery, technological invention, screenplay construction, venture creation, and Wikipedia article development, we consistently observe that teams with high perspective diversity combined with low background diversity achieve significantly higher performance outcomes. This pattern holds true in observational data, a natural experiment, and controlled LLM simulations, suggesting a robust relationship between these diversity dimensions and innovation success. These results clarify a longstanding tension in team assembly research by demonstrating that effective collaboration requires both cognitive variety in the form of diverse perspectives and also communicative common ground through shared backgrounds, enabling teams to simultaneously generate novel combinations while maintaining effective coordination.

These findings hold implications for science policy and team assembly processes. Funding agencies, research institutions, and innovation-driven organizations should consider measuring and convening teams that exhibit optimal configurations of diverse experience—high perspective diversity paired with low background diversity—rather than relying on traditional demographic or disciplinary diversity metrics alone. Our conceptual framework offers a quantitative method for identifying potential collaborators who occupy distinct vantage points within shared knowledge domains, enabling more strategic team formation. Policy interventions could focus on creating environments that foster connections between individuals with complementary perspectives rather than random, homogeneous or heterogeneous groupings through targeted networking opportunities, collaborative platforms, or funding mechanisms that explicitly foster teams demonstrating subjective diversity and shared experience. This approach could significantly enhance innovation capacity across domains while maximizing the effectiveness of limited research resources. Finally, our LLM-based measurements and simulations suggest a range of applications for generative social science[37,40].

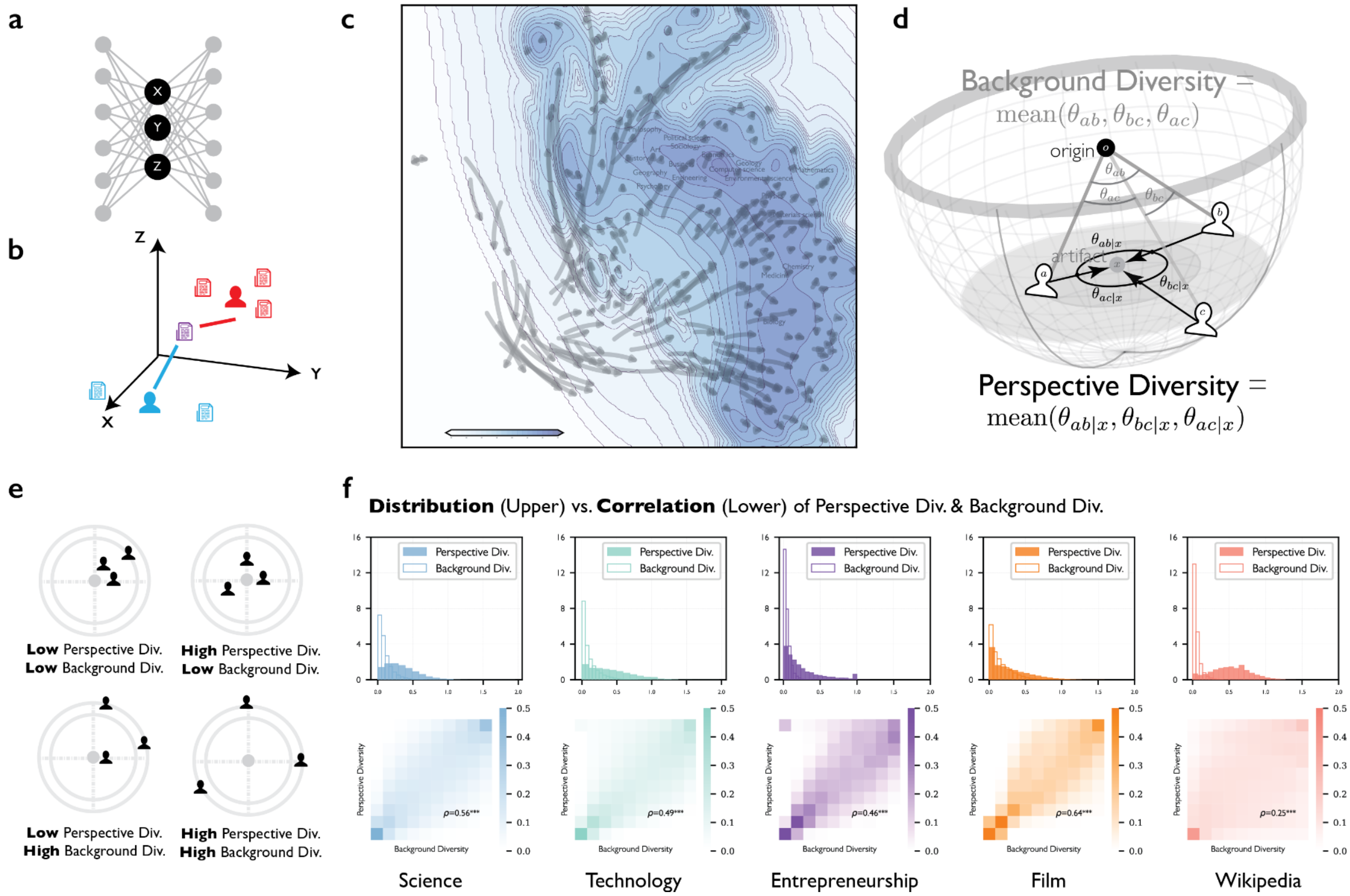

**Fig. 1. Conceptual space construction and dimensions of subjectivity. a-b**, Schematic diagram of the neural network construction of domain-specific conceptual spaces (**a**) and how they enable a geometrical mapping of creative works and their authors (**b**). **c**, Graph of idea movements across the conceptual space of science, which exhibits active movement toward and convergence upon the crowded contemporary scientific frontier. **d**, Definition of perspective diversity $\bar{\theta_p}$ versus background diversity $\bar{\theta_b}$ within an embedding model for hypothetical team members *a*, *b*, and *c*. **e**, Configurations of team-level perspective and background diversity. **f**, Distribution and statistical relationships between the distributions of perspective and background diversity by case, which exhibit mild correlation.

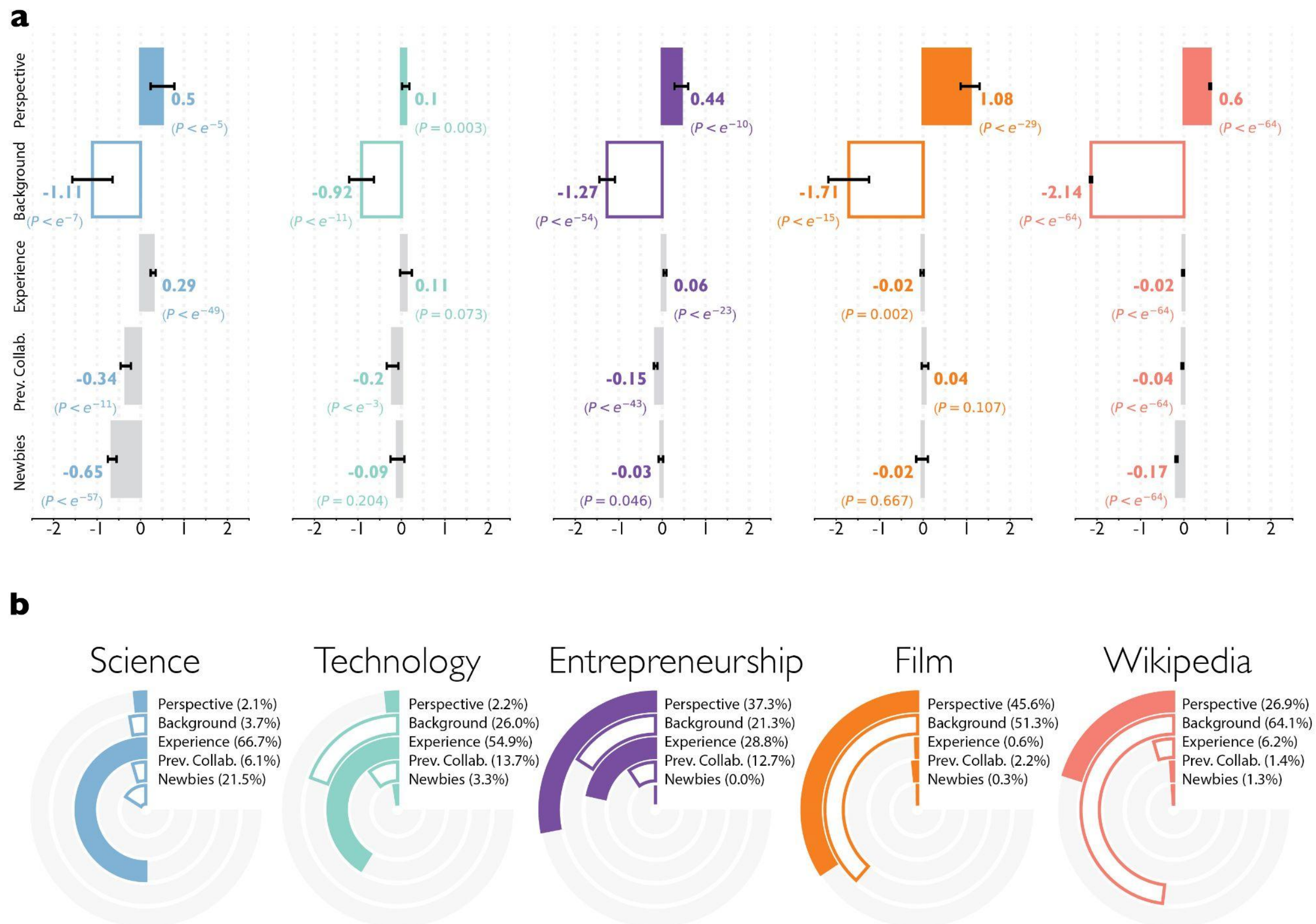

**Fig. 2. Comparison of team features in their contribution to the prediction of collaboration success.** **a**, Comparison of regression coefficients in predicting domain-specific success. Bars represent 95% CI. Models are reported in SI, Tables S3 and S4, with year, team size, and domain fixed effects controlled. **b**, Variable-level dominance analysis demonstrating the relative importance of each predictor by comparing the incremental contribution to the model's predictive power ($R^2$) for domain-specific success in science, patents, start-ups, movies, and Wikipedia pages.

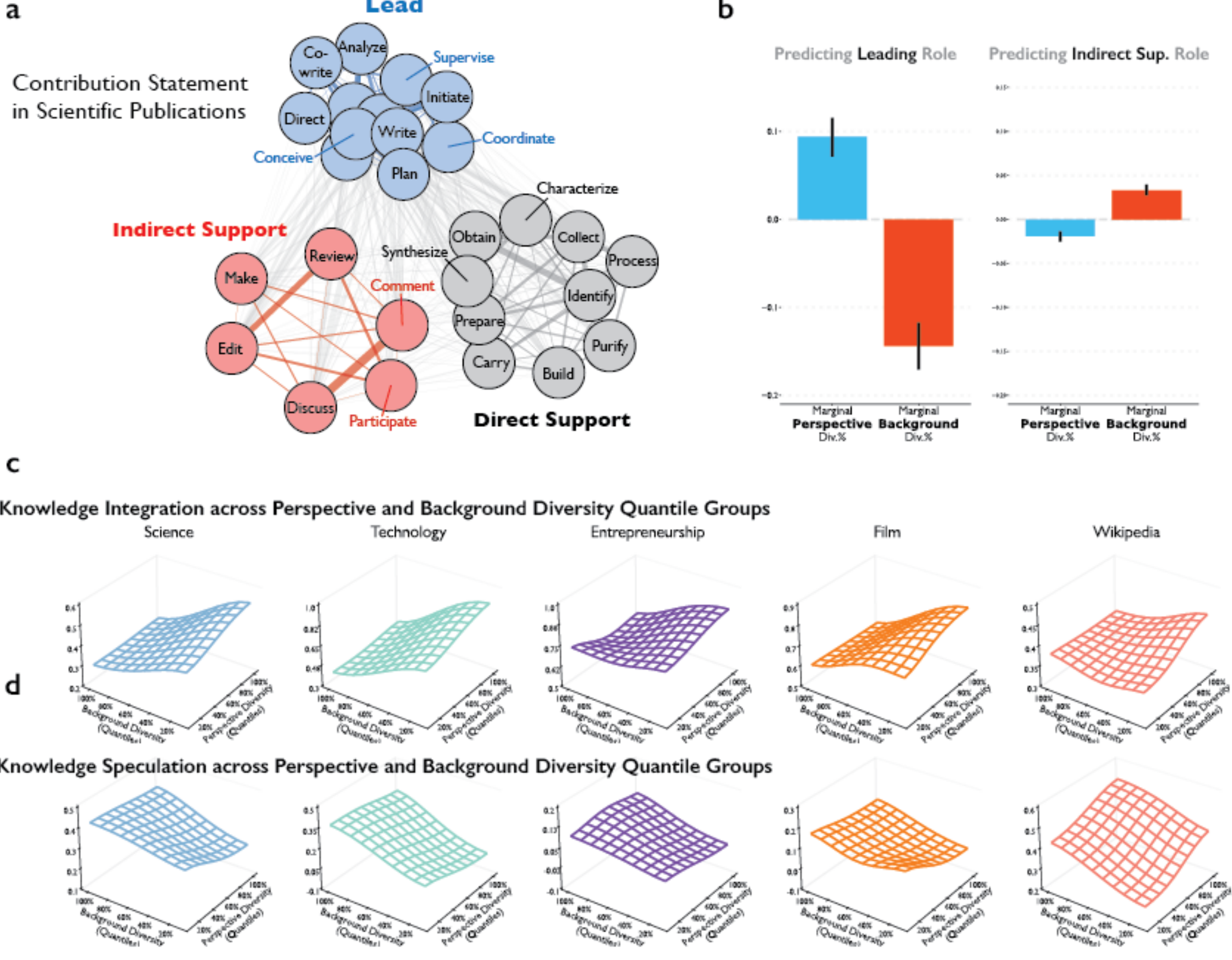

**Fig. 3. Mechanisms underlying the effects of diverse perspectives. a,** Clusters of research roles based on whether the same researcher co-contributes them in the same paper. The author's contribution statements come from four journals: *Science*, *Nature*, *PNAS*, and *PLOS ONE*. **b,** Model-based estimates of how contributing to perspective diversity is linked to a higher likelihood of assuming central roles, whereas contributing to background diversity is associated with more marginal roles. **c,** The relationship between perspective and background diversity and the level of knowledge integration in the focal project, reflecting whether team members effectively recognize and leverage the intellectual overlap in their prior experiences. **d,** Relationship between perspective and background diversity and the likelihood of speculation, defined as the proportion of knowledge components mobilized within the focal project that have never appeared in any team member's prior work.

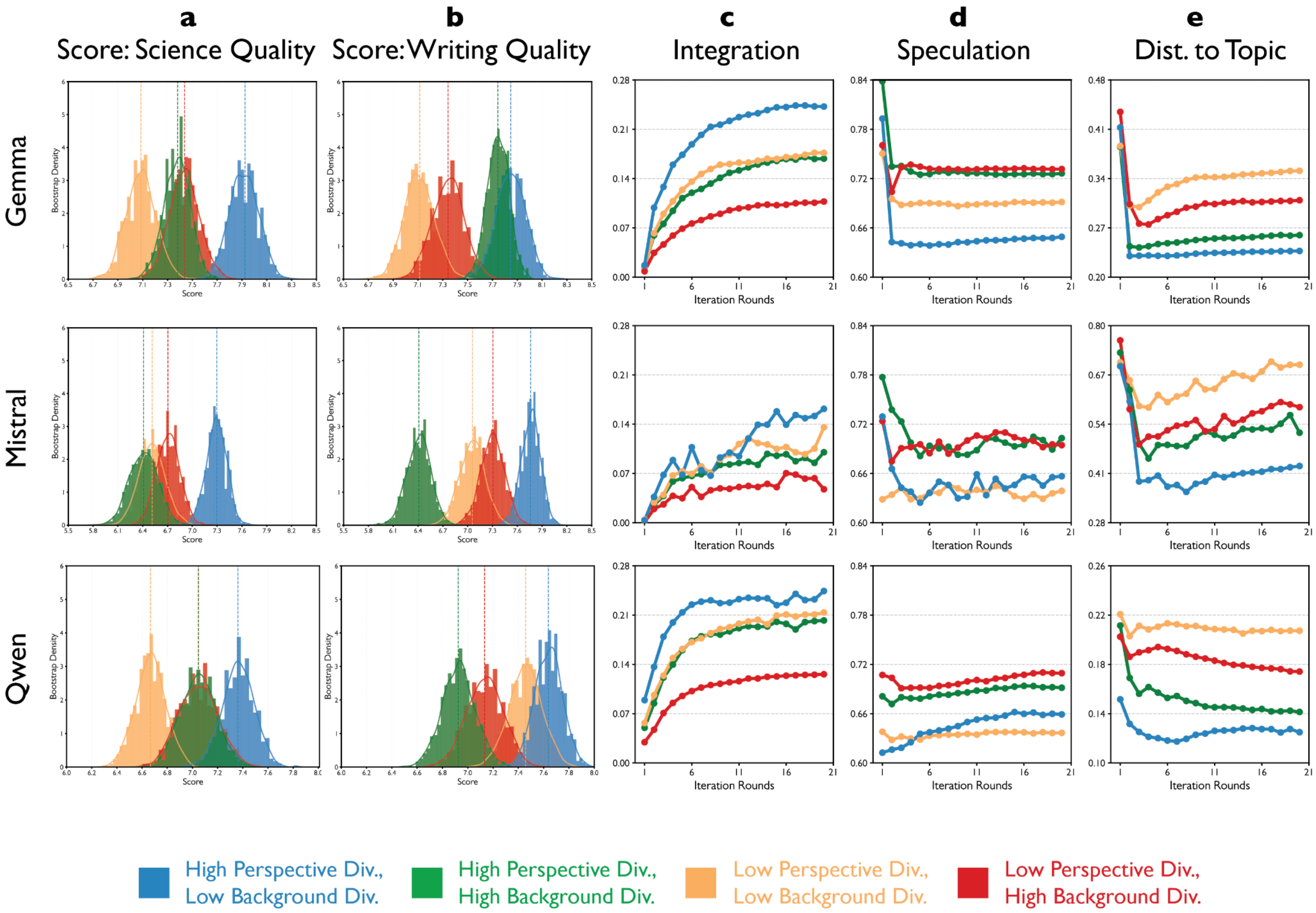

**Fig. 4. Findings from the LLM Experiment.** We provide randomly selected scientist profiles from MAG for LLMs to construct simulated teams across the perspective-based team typology. Each simulated team engages in a 20-round process of discussion and paper writing, during which we track intermediate versions of the papers. We replicate our simulated experiment using three open-source large language models—Gemma-3-27B-it, Mistral-Nemo-Instruct-2407, and QwQ-32B—which differ in parameter size and model architecture. In total, we run simulations for 240 teams (20 topics, 4 conditions, and 3 LLM settings). We design each team to have 3 agents. We assess the quality of the papers produced by our simulated teams using GPT-4o and drawing on two evaluation frameworks: **a**, the NeurIPS 2024 reviewer guide (highlighting originality, clarity, etc.), and **b**, the Association of American Colleges and Universities (AAC&U) *VALUE* Rubrics for written communication. Then, we construct token-level measures for **c**, knowledge integration and **d**, speculation just as we do in our statistical analysis of empirical data (see Fig. 3 and Fig. S8). **e**, We finally compute the distance between each paper and its assigned focal topic, defined as the cosine distance between the centroid of token embeddings in the team's generated paper and the embedding vector of the assigned topic.

# Data and Methods

## Datasets

**Scientific Publications.** For scientific publications, we utilize a snapshot of Microsoft Academic Graph (MAG), accessed in December 2021. The dataset comprises 269,813,751 entries, including scientific papers, repositories, patents, and datasets. Due to data sparsity before 1980 and after 2015, we focused on journal articles published between 1981 and 2015, and retained only those articles with valid abstracts and at least two authors who had publishing records covering the previous five years to ensure meaningful diversity based on their past experiences. The final compiled dataset includes 20,405,559 valid papers. The dataset initially contained 278,614,730 individual profiles. We used density-based clustering[39,41] to disambiguate multiple profiles linked to the same normalized author name, indexing these unique profile clusters with new author IDs. This additional disambiguation leaves 254,365,736 unique individuals in the whole dataset, among whom 16,878,569 are related to our paper sample. In mechanism analysis, we link MAG data to a dataset of increasingly mandatory author contribution statements from four journals, including 13 years of *Proceedings of the National Academy of Sciences* (*PNAS*; 18,354 statements from 2003–2015), 15 years of *Nature* (9,364 statements from 2006–2020), 3 years of *Science* (1,176 statements from 2018–2020), and 9 years of *PLOS ONE* (60,681 statements from 2006–2014). We classify the top 50 research activities within author contributions into categories including leadership (e.g., “conceived” of the study), direct support (e.g., “analyzed” data) and indirect support (e.g., “commented on” the manuscript) following the protocols established by Xu et al[18]. These activities account for 95.37% of all activities at the individual level (see SI, S1.1 and S1.6, for further details).

**Patented Inventions.** For technological inventions, we use the United States Patent and Trademark Office (USPTO) dataset. As of the data collection date, it contains a total of 8,517,464 patents from 1976 to the present. It provides content detail about the technology and its proposed use, such as patent abstracts, and metadata on inventors and assignees, such as name and gender of inventors, and patent ownership and transfers between organizations. We define a patent team as multiple inventors listed on the same granted patent, and retained only those patents with valid abstract and at least two inventors who had invention records in the previous five years, to ensure meaningful diversity measures and facilitate comparison with scientific publications. In the final dataset, we retained 2,841,066 valid patents with 1,844,890 related inventors (see SI, S1.2, for further details).

**Startups and Venture Capital Firms.** For entrepreneurial business designs, we use Crunchbase, a crowd-sourced platform that records extensive information on startup companies and their funding sources. Crunchbase data provides detailed information on both new ventures and investors, such as company name, address, industry, business descriptions, funding histories, and employee composition. We focused on startups based in the U.S. or with U.S. headquarters, established between January 1, 1950, and December 31, 2021. In total, this dataset contains 298,915 new ventures, 174,926 funding records, and 41,170 unique investors. Of these startups, 81,765 have received funding, and 198,226 have valid business descriptions. We retained only startups with (1) valid business descriptions and (2) at least two investors with investor records from the past five years to ensure meaningful diversity measures. The final dataset includes 10,181 valid startups and 11,572 related investors (see SI, S1.3, for further details).

**Movies.** For movie creation, we use the Internet Movie Database (IMDb, www.imdb.com), which provides detailed records for each film, including movie titles, languages, regions, genres, crew information, and ratings. The official bulk data provided by IMDb is supplemented with additional information from the IMDb API (https://imdb-api.com/), which offers more in-depth information, including extended plot summaries, awards, and box office figures. The dataset we used, downloaded in February 2023, includes 261,302 movies, featuring 107,743 directors, 175,952 screenwriters, and 1,218,612 actors and actresses. We retained only those movies with valid plot summaries and at least two screenplay writers who have produced movie plots in the past five years to ensure meaningful diversity measures. The final dataset includes 29,797 valid movies and 28,556 related screenplay writers (see SI, S1.4, for further details).

**Wikipedia Pages.** For Wikipedia data, we use the bulk download service provided by Wikimedia database dumps. We downloaded all English Wikipedia pages in December 2023, which included 17,485,337 pages, 9,280,838 registered users, and 197,562,914 substantive revisions (each containing at least 100 tokens). To assess editor team structure, minor revision records that involve fewer than 100 English characters are excluded. We use each yearly snapshot as the analytical level, and retain only those with at least two editors who had Wikipedia revision records in the previous two years to ensure meaningful diversity measures. The final dataset includes 5,058,522 first-year creation records of pages, 89,261,730 page-year records as longitudinal data, and 20,501,961 users (see SI, S1.5, for further details).

**Modeling Concept Spaces for Innovation Emergence**

**Estimation.** To construct conceptual spaces across multiple time periods, we use a dynamic embedding algorithm[42] specifically designed for longitudinal corpora. This approach assumes and enforces a smooth transition between word vectors and their associated meanings in adjacent time periods and trains all periods jointly. Specifically, it alters the objective for the joint word embedding to the following:

$$\min_{U(1),\ldots,U(T)} \frac{1}{2}\sum_{t=1}^{T} \|Y(t) = U(t)U(t)^{T}\|_F^2 + \frac{\lambda}{2}\sum_{t=1}^{T} \|U(t)\|_F^2 + \frac{\tau}{2}\sum_{t=2}^{T} \|U(t-1) = U(t)\|_F^2 \tag{1}$$

In addition to the standard word embedding objective to align the empirical word proximities $Y(t)$ and the modeled proximities $U(t)$, dynamic word embeddings add two additional terms that: 1) allow the algorithm to minimize unnecessary word vector loadings (the $L_2$ norm of $U(t)$), and 2) shrink the distance between time-adjacent embeddings (the $L_2$ norm of the difference between $U(t\text{-}1)$ and $U(t)$).[1] In this way, the jointly trained algorithm keeps adjacent semantic spaces close in structure. This approach allows us to construct comparable and historically accurate representations of conceptual structures for each time period.

For each case, we set a frequency threshold of 150 for word inclusion, learn 50-dimensional representations for each word using a window size of 5 and perform 10 iterations. For the science, technology, commerce, and culture cases, we use a 5-year time window, while for Wikipedia, we use a 2-year time window to capture the fast-paced changes and frequent updates typical of (hyper)active online platforms. To address instability in the IMDB and Crunchbase data due to their smaller volume, we use Google *n*-grams to construct a cultural space defined by fiction and nonfiction books, within which we project the IMDb screenplay summaries, and use 118 financial newspapers (e.g., *Wall Street Journal*, *Financial Times*, *Bloomberg Businessweek*) from ProQuest to construct a commercial space, within which we projected the Crunchbase venture descriptions (see SI, S2.1, for further details).

**Validation.** We validate our dynamic semantic space with a predictive task. We expect that within the space, regions in which concepts flow together and interact are precisely where innovations are most likely to emerge, consistent with theories of recombinant innovation[1–3]. For each case, we randomly select 5,000 uniform points in the semantic space and track the average "in-flow" of local concepts, defined by the increase in cosine similarity between selected points and nearby word vectors between two time periods. We then record the number of innovations (e.g., scientific papers, technology patents, screenplays, company descriptions, Wikipedia pages) that appear within a specific range in the second period. Correlations between local concept convergence and innovation bursts are consistently and significantly positive across contexts, as illustrated in Figures S2 to S6. We also conducted a Dumitrescu & Hurlin (2012) Granger Causality Test to examine the possibility of a directed correlation between concept "in-flow" and innovation emergence. This analysis was performed on a panel of 2,000 randomly selected subspaces. Results, presented in Table S1, indicate that spatial dynamics are a much stronger driver of innovation than innovation is of spatial dynamics.

[1] This dynamic word embedding approach relies on the formal equivalence between the standard negative sampling approach to word embedding optimization used by `word2vec` and the low-rank factorization of a pointwise mutual information (PMI) matrix of word vectors (Levy and Goldberg, n.d., 2014), and uses the PMI as input to the objective above.

We conducted a second validation test at the individual innovator level. We hypothesize that innovators are more likely to adopt a concept when (1) the concept moves closer to their position over time, and (2) this movement is more visible to them. To test these hypotheses, we run a series of fixed effects regressions. Within each dataset, we examine the geometric relationships between innovators and concepts by measuring: (1) the change in cosine similarity between the innovator and the concept over time, and (2) the "visual angle" of concept movement, defined by the cosine of the angle between its starting and ending points. These two variables, along with their interaction, are used to predict whether an innovator's subjective perspective influences their adoption of a new concept (not used in period t) in t+1. We control for author and period fixed effects. As shown in Extended Data Fig. 2, the results support our hypotheses: innovators are more likely to adopt concepts that move closer to them, especially when such movement is more visible from their unique perspective.

**Quantifying Diversity**

We calculate an innovator's experience vector $V_i$ as the average of word vectors from their previous engagements—publishing science, inventing technologies, financing businesses, writing movies, and drafting Wikipedia pages. Based on these experience vectors, we construct perspective vectors for member *i* within team *p* as $V_{p,i} = V_{task} - V_i$, which reflect the relationship between an innovator's previous experience and the focal artifact they are constructing together [43,44]. Given a set of experience vectors in a team $\{V1, V2, V3,...., Vn\}$, the team will have a corresponding set of perspective vectors $\{V_{p,1}, V_{p,2}, V_{p,3},...., V_{p,n}\}$. The team's background diversity can be measured as

$$BD = \frac{1}{n \times (n-1)} \sum_{i=1}^{n} \sum_{j=1, j \neq i}^{n} CosineDistance(V_i, V_j)$$

, while perspective diversity can be defined as

$$PD = \frac{1}{n \times (n-1)} \sum_{i=1}^{n} \sum_{j=1, j \neq i}^{n} CosineDistance(V_{p,i}, V_{p,j})$$

. The correlation between background diversity and perspective diversity across all five datasets is shown in Fig. 1 in the main text. The coefficients, all positive and highly significant, range from 0.25 to 0.64. Variance Inflation Factor (VIF) tests have been conducted after each regression to confirm no issues with collinearity have been identified in our models.

In our analysis of mechanisms, we also estimate whether each team member's diversity contribution increases their probability of taking on distinctive roles. To measure the marginal diversity contribution for each individual on each team, we construct marginal diversity contributions of each member,  as the proportion of the difference between the full team's diversities and those of the counterfactual team where the focal member is absent, capturing the proportion of diversity that a focal individual contributes to the original team:

Marginal Contribution to Background Diversity:

$$MBD_a = \frac{(BD_{full} - BD_{-a})}{BD_{full}}$$

Marginal Contribution to Perspective Diversity:

$$MPD_a = \frac{(PD_{full} - PD_{-a})}{PD_{full}}$$

See SI, S2.3, S2.4 and S2.6, for further details.

**Quantifying Team Performance**

We evaluate team performance within each case using domain-specific criteria, as extracted from previous studies. For science and technology teams, we use five-year forward citations as a proxy for impact[17,45]. For startups, we measure performance with the total number of funding rounds from venture capital firms[46,47]. Additionally, we assess long-term success using milestone events, such as initial public offerings (IPOs) or high-value acquisitions. For movies, we evaluate quality using average IMDb ratings, which reflect judgments from popular audiences, and complement this with a professional evaluation of whether the movie won an award at various film festivals (1=YES, 0=NO). For Wikipedia pages, we use the page-quality assessment algorithm available through the official Wikipedia API, trained on expert ratings regarding article content and is blind to the process of article creation. This score has been frequently used in scientific research on open source knowledge production[48]. We also calculate yearly page views as an alternative metric of page popularity, which manifests the same pattern of results. See SI, S2.5, for further details.

**Quantifying Knowledge Integration and Speculation**

To measure knowledge integration within each team project, we utilize knowledge taxonomies that represent the knowledge structure of each project, following a long tradition of innovation studies[1,3]. We measure whether all team members participate in decision-making and contribute effectively by observing whether they have selected modules that align with their common and collective experiences. We constructed two measures to capture this information. The first variable, *Knowledge Integration,* measures the extent to which knowledge modules overlap with team members' previous knowledge modules on average. The second variable, *Speculation,* calculates the proportion of knowledge modules that are new to all team members. We constructed the two measures with equations:

$$Integration = \frac{1}{N_c \times N_p} \sum_{c=1}^{N_c} \sum_{p=1}^{N_p} (I_{c,p} == 1)$$

$$Speculation = \frac{1}{N_c} \sum_{c=1}^{N_c} (\sum_{p=1}^{N_p} I_{c,p} == 0)$$

where $N_c$ represents the total number of relevant knowledge modules in the innovation product, $N_p$ represents the number of individuals in the team, and $I_{c,p}$ is a dummy variable indicating whether innovator $p$ has previously participated in category $c$, previous to this team project (1=Yes, 0=No). The operationalization details of these measures are illustrated in SI, S2.8, Fig. S8.

**Estimating the Relationship between Multi-Dimensional Diversity and Team Performance**

We perform fixed-effect regressions to control for confounders, including team size, year, and main subfield. Models with additional fixed effects, such as individual fixed effects, are included in the Appendix. The basic form of the regression equation is specified as:

$$TeamPerformance_i = \beta 0 + \beta 1 \times PD_i + \beta 2 \times BD_i + Controls_i + FE_{teamsize} + FE_{year} + FE_{subfield} + \epsilon,$$

where $Controls_i$ represents controls on the team level and is case-specific. For science publications, subfields are defined by disciplines (level 1—292 in total) within MAG. For patents, control variables include (1) Number of claims; (2) Number of other patents in the same family; (3) Number of citations to non-patents (i.e., scientific documents), and subfields are defined by Cooperative Patent Classification ("CPC") categories. For movies, control variables include (1) Dummy for language; (2) Dummy for country, and subfields are defined by genres, as labeled in the IMDb dataset. For startups, control variables include (1) Dummy for headquarter location; (2) Dummy for founder's gender (0=Male, 1=Female) and (3) Dummy for the founder's race (0=White, 1=Minority), and subfields are defined by industry codes in Crunchbase[2]. For Wikipedia, we do not control for subfield fixed effects in Wikipedia pages, as there is no mutually exclusive classification system. However, we do control for page ID fixed effects to account for variations in content across different versions of the same article, thereby addressing potential confounders related to the specific nature of knowledge.

Performance measures in our five cases vary in format. For zero-inflated count data (such as science citations, patent citations, and startup funding), we estimate negative binomial regressions. For binary outcomes (such as the ultimate success of startups and the award-winning status of movies), we use logistic regression. For continuous outcomes (such as average movie ratings and quality scores for Wikipedia pages), we apply OLS models.

Results for the main models and mechanism analyses are presented in Tables S3 to S7. Robustness checks are reported in Tables S8 to S13, along with Extended Data Figs. 3, 4, and 7.

[2] As listed in : https://support.crunchbase.com/hc/en-us/articles/360043146954-What-Industries-are-included-in-Crunchbase

See SI, S2.9, S2.10, S3.1 and S3.2, for further details.

**Wikipedia Natural Experiment**

Wikipedia's annual Central and Eastern Europe (CEE) Campaign serves as a natural experiment to help us understand how team diversity affects content quality, independent of selection through success bias. Running from March 21 to May 31 each year, the campaign encourages editors to create and improve articles about CEE history, culture, and notable figures, attracting participants from beyond the usual CEE editing communities. The researchers analyzed 495,262 editing records from 28,907 pages between 2017-2024, using coarsened exact matching to compare pages before and during the campaign period. We found that the campaign disrupted established editing circles and potentially improved Wikipedia content through increased perspective diversity (Extended Data Fig. 5). See SI, S4.1, for further details.

**LLM Simulation Experiment**

To test the causality of observed relationships between success, perspective and background diversity across five innovation settings, we design a virtual experiment based on a methodological framework of multi-agent interactions using large language models (LLMs)[49–51]. We begin by selecting 20 key topics from the MAG dataset and construct four virtual teams for each topic, following the design principles outlined in this paper. These teams are assembled from a pool of 20,000 real-world scientists, whose profiles we draw randomly from MAG. We then input both the topic information and team member backgrounds into LLMs to initiate simulated discussions, prompting the agents to collaboratively generate a shared output. We use three models—Gemma-3-27b-it, Mistral-Nemo-Instruct-2407, and QwQ-32b to replicate the simulation for each topic-condition. As of April 2025, when this analysis was conducted, these models were recognized as among the most top-performing models available. With all other potential confounders held constant and key predictors randomly assigned, the relationship between team design and the quality of collaborative outcomes can be interpreted as causal.

For each virtual team—defined by its assigned topic, selected members, and designated LLM—we simulate collaboration using the AutoGen framework[8]. Each team engages in 20 rounds of discussion and editing, during which all agents participate in conversation and jointly draft, revise, and polish a short paper. After collecting all papers produced through these simulations, we evaluate their quality using GPT-4o (accessed in April 2025)[52]. Two assessment criteria are employed: (1) the NeurIPS 2024 reviewer guidelines, to assess the strength of the scientific argument; and (2) the Association of American Colleges and Universities (AAC&U) VALUE Rubrics for written communication, to assess overall writing quality. We set the temperature of GPT-4o to 0 to ensure evaluation consistency. Finally, we statistically test whether the quality scores differ significantly across team compositions.

Details on the model settings—including prompts, examples of team interactions, and LLM-generated short papers—are provided in the SI, S4.2, with Extended Data Fig. 6 visually illustrating the methodological framework.

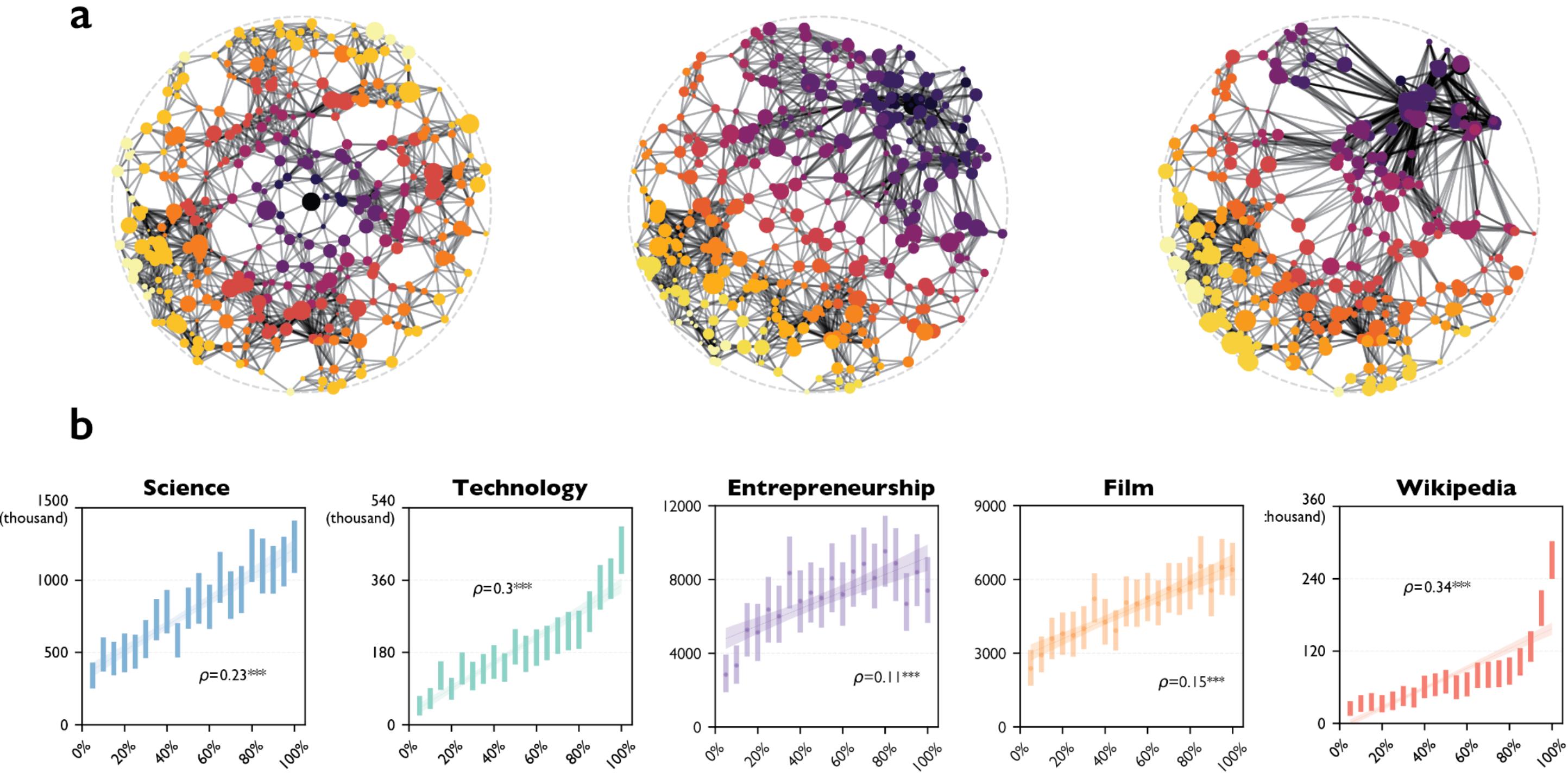

**Extended Data Figure 1. Predicting innovation emergence through the dynamics of knowledge space.** Innovation theories suggest that the compression of knowledge space brings idea elements closer together, thereby increasing local opportunities for novel ideas to emerge, as illustrated in the conceptual figure in Panel **(a)**. To test this hypothesis, we **(b)** randomly selected 5,000 points from the final slice of the high-dimensional space in each case. We then calculated the correlation between the conceptual flow in the surrounding area (defined by a distance threshold $t_1$, which by default includes the 30% of ideas nearest to each focal point) and the innovations that emerged near the focal point in the following period (using a second threshold $t_2$, which by default includes the closest 12% of new innovations). In other words, we measured the correlation between predicted and actual probability of innovation emergence. The results reveal a consistent and significant positive relationship, with Pearson correlation coefficients of 0.11 in the business space, 0.15 in the cultural space, 0.23 in the scientific space, 0.30 in the technological space, and 0.34 in the Wikipedia space. We also tested alternative thresholds: $t_1$ values corresponding to the 50%, 60%, 70%, 80%, and 90% percentiles of conceptual distance, and $t_2$ values at the 80%, 84%, 88%, 92%, and 96% percentiles of product distance to the focal point. These additional results, reported in the Supplement Materials (Figs. S2-S6), are consistent with the main findings presented here.

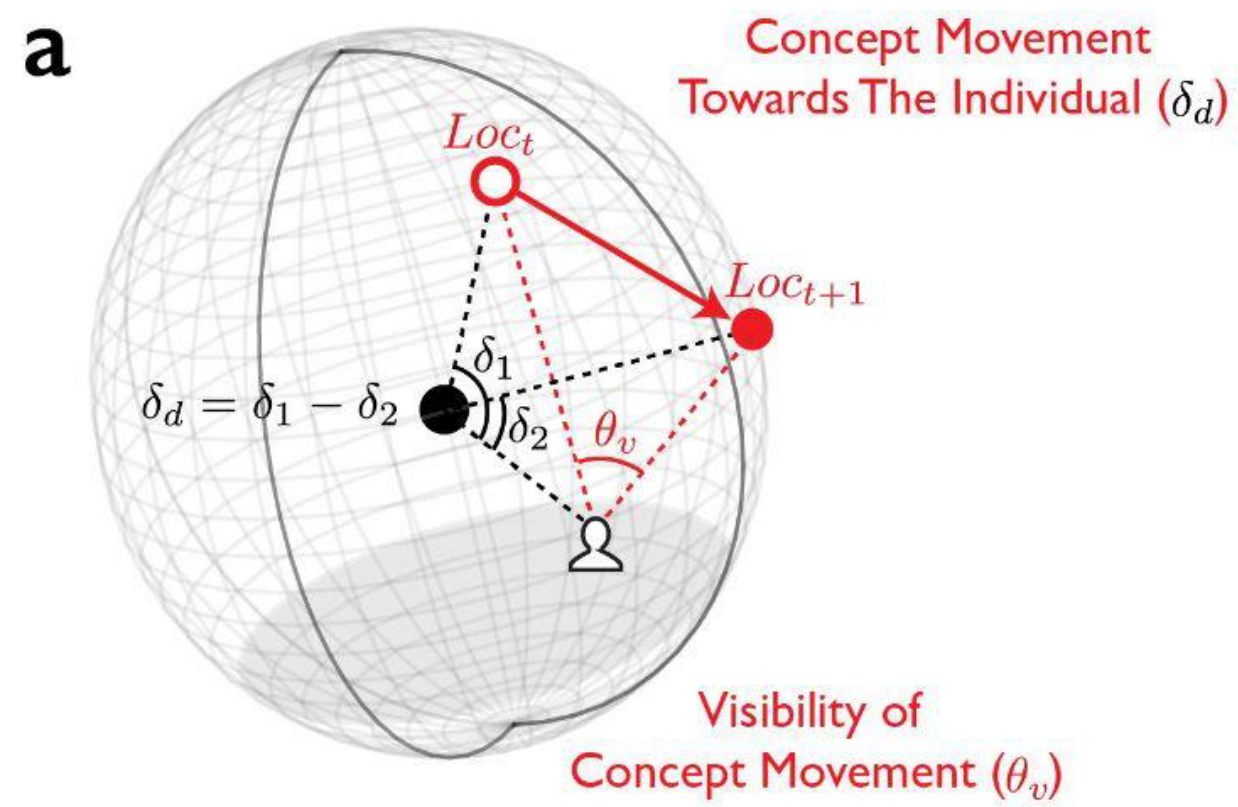

**We estimated the model**

$$y = \alpha + \beta_1 \delta_d + \beta_2 \theta_v + \beta_3 \delta_d \theta_v + Controls + FE_{period} + FE_{innovator} + \epsilon$$

$\beta_1$ captures the effect of **concept movement** with respect to individuals, $\beta_3$ reflects the **conditional impact of concept visibility** for those individuals and $y$ represents the dummy for whether the author will **adopt this new concept** in $t+1$

**b** **Concept Distance Shrinkage Leads to Higher Probability of Adoption**

Science | Technology | Entrepreneurship | Film | Wikipedia

y-axis: $\beta_1 + \beta_3 \times \theta_v$; x-axis: $\theta_v$

**Y:** Effect of Concept Moving Closer ( $\beta_1 + \beta_3 \times \theta_v$ )

**X:** Visibility ($\theta_v$)

**Extended Data Figure 2. Innovators' positions in the knowledge space predict their creative activities.** For each case and time period, we randomly sample 20,000 innovators, collect their innovation records, and examine whether they adopt new ideas based on shifts of ideas in the knowledge space. Regression analyses, controlling for period and individual fixed effects, show that innovators are more likely to adopt a new idea when it moves closer to their position in space. This effect is stronger when the "visual angle" of idea movement is larger—i.e., concept movement is more visible from the innovator's subjective view, given their own position in the space. Results of the models are reported in Supplementary Table S2.

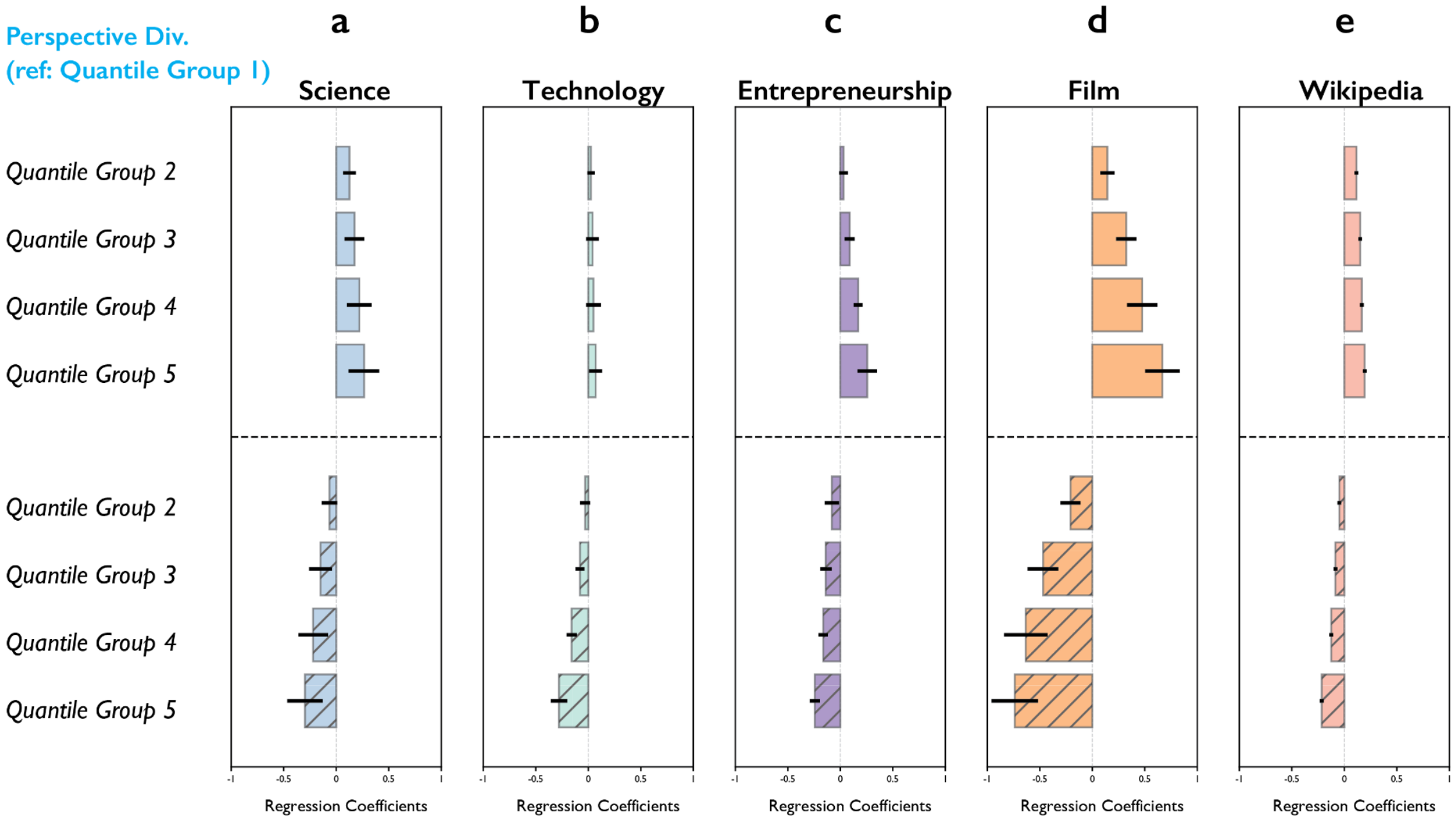

**Extended Data Figure 3. Robust linear relationship between two dimensions of diversity and innovation impact.** Based on our five datasets, including research papers published between 1981 and 2015 (N = 20,405,559), patent applications filed between 1976 and 2017 (N = 2,841,066), U.S.-based startups founded between 1950 and 2021 (N = 10,181), films produced globally between 1971 and 2023 (N = 29,797), and English Wikipedia pages initiated between 2003 and 2022 (N = 5,058,522), we estimate our regression model with major diversity measures discretized into categorical variables and the most stringent set of controls applied (model specification consistent with Tables S3 and S4). We divide both perspective diversity and background diversity into five levels, using the lowest level as the reference group in regressions. The results reveal a clear monotonic change in the effects of both perspective diversity and background diversity as the predictor values increase, providing strong evidence of a linear association with innovation impact. This linear pattern is consistently observed across all five domains.

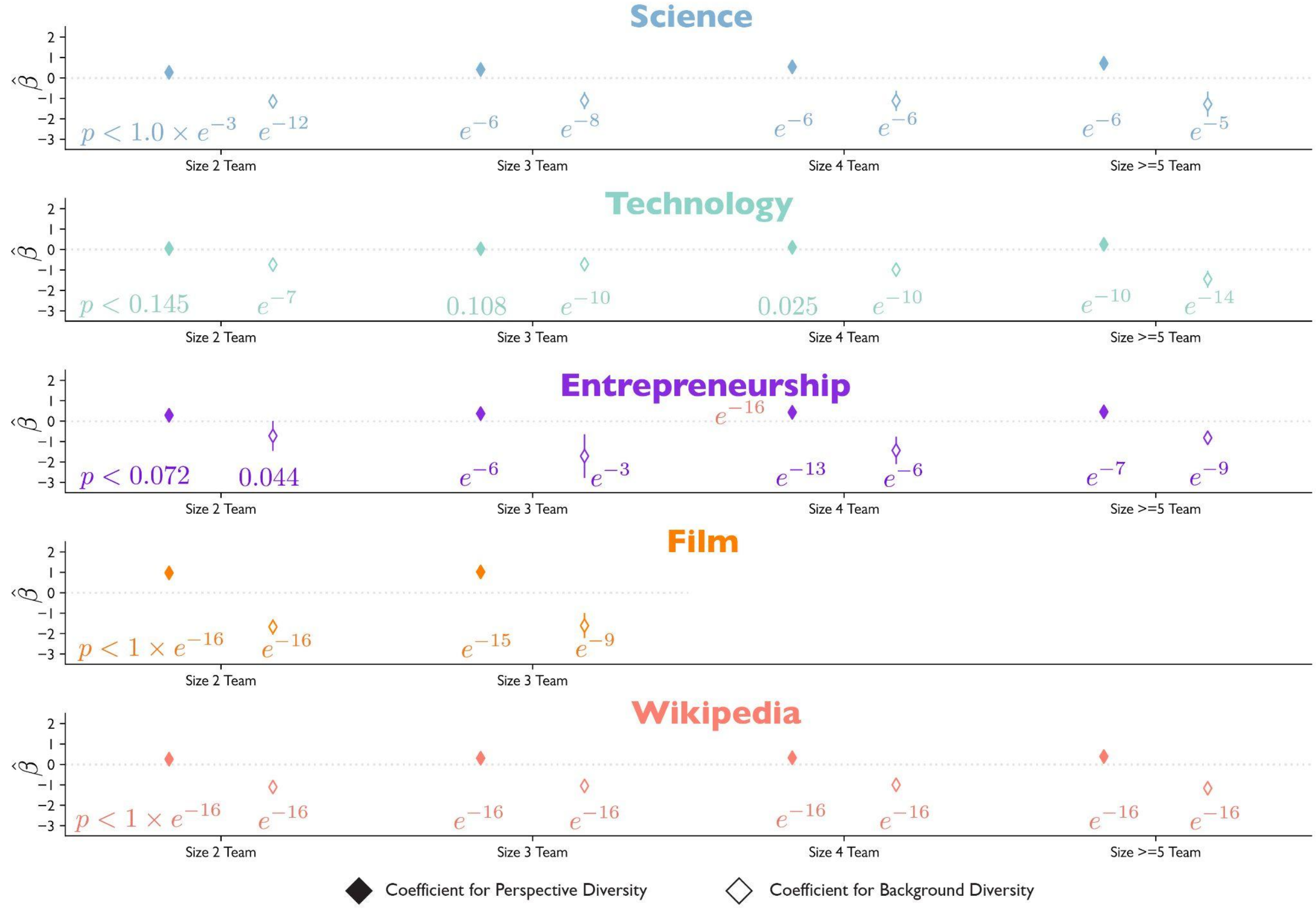

**Extended Data Figure 4. The robust effects of the relationship between two dimensions of diversity and innovation impact, against team sizes.** Based on our five datasets, including research papers published between 1981 and 2015 (N = 20,405,559), patent applications filed between 1976 and 2017 (N = 2,841,066), U.S.-based startups founded between 1950 and 2021 (N = 10,181), films produced globally between 1971 and 2023 (N = 29,797), and English Wikipedia pages initiated between 2003 and 2022 (N = 5,058,522), we estimate our regression model separately for each subsample defined by team size, with the most stringent set of controls applied (with model specification consistent with Tables S3 and S4). The results show that the positive estimated relationship ($\hat{\beta}$) between perspective diversity and innovation impact, as well as the negative relationship ($\hat{\beta}$) between background diversity and innovation impact, remain statistically significant and stable across different team sizes. These findings suggest that the observed effects are not driven by variation in number of team members.

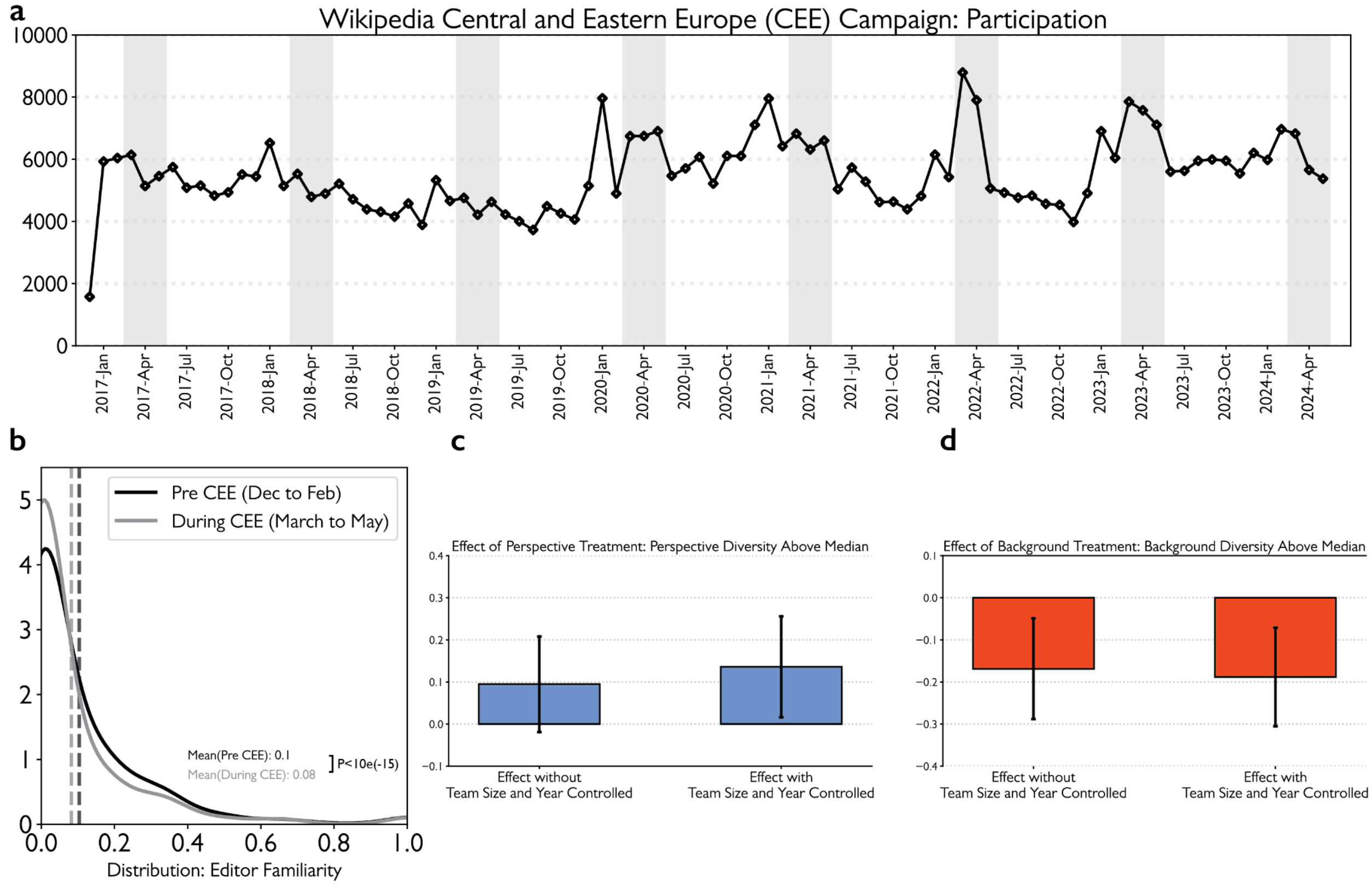

**Extended Data Figure 5. A natural experiment based on the Wikimedia Central and Eastern Europe (CEE) campaign.** We analyzed editing records of Wikipedia pages related to the CEE campaign from 2017 to 2024, resulting in 495,262 edits across 28,907 pages over 8 years. Both the average number of edits per month **(a)** and the density of prior collaboration networks among page editors **(b)** indicate that the campaign attracted new contributors who were previously unconnected to CEE pages and communities. This external shock, which partially disrupts existing collaboration structures within Wikipedia communities, offers an opportunity to examine the causal relationship between diversity and page quality, as it reduces the likelihood of intentional team design in advance. Our analysis focuses on the active period of the campaign, from March 21 to May 31 each year. We first apply a coarsened exact matching (CEM) design using editing records from the three months prior to the campaign. At the page-year level, pages are matched based on year, country, topic, page popularity, and editor team characteristics, including perspective diversity, background diversity, team size, and team scores (in groups). We then examine whether an increase in perspective and background diversity during the campaign period (treatment = 1 if diversity exceeds the median, 0 otherwise) predicts improved page quality. Results shown in Panels **(c)** and **(d)** demonstrate that, after accounting for these confounding factors, the effects of both diversity measures remain robust and align with those identified in our broader statistical analyses.

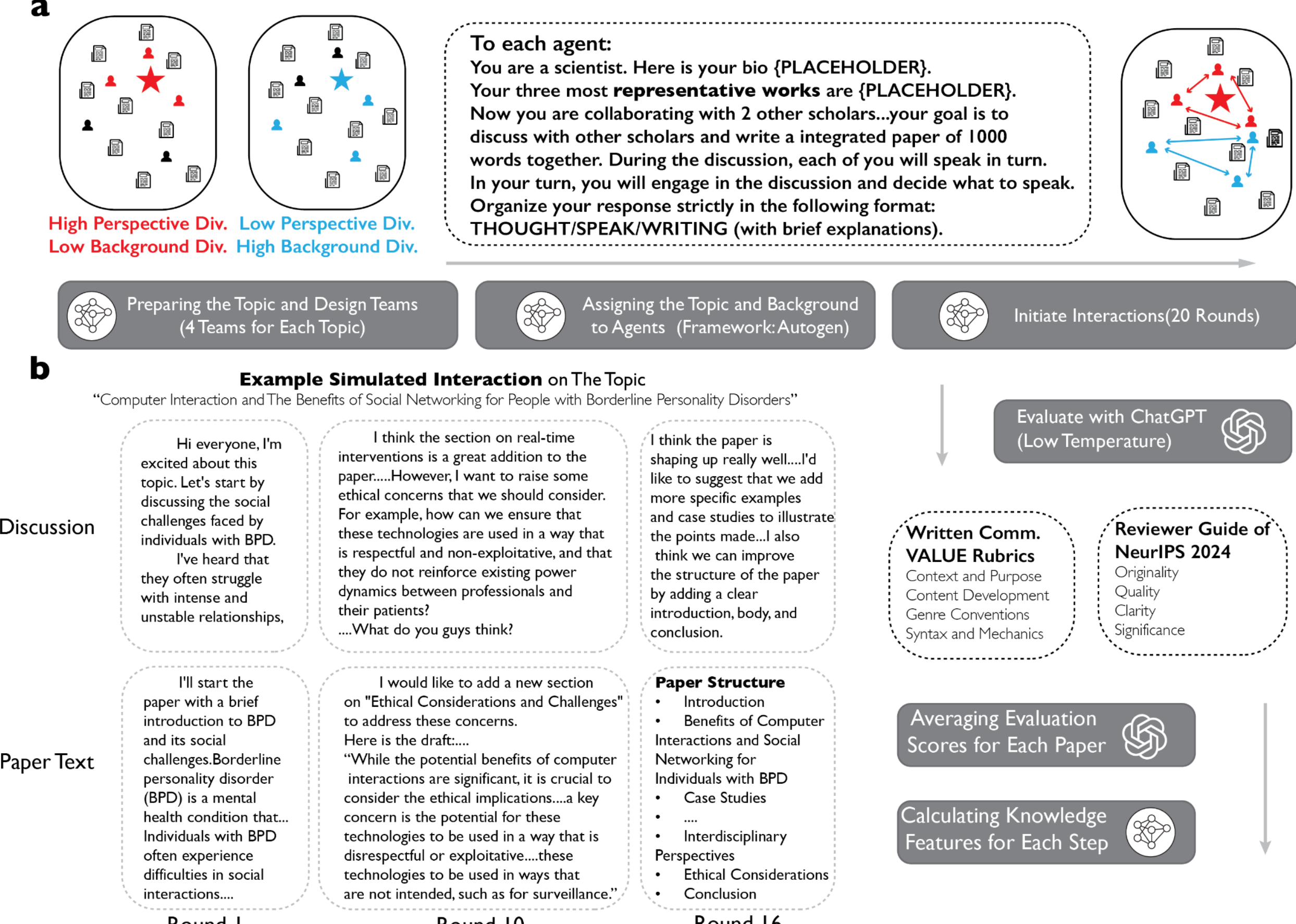

**Extended Data Figure 6. Procedure for Conducting LLM-Based Virtual Experiments Using the AutoGen Framework.** We replicate our simulated experiment using three open-source large language models: Mistral-Nemo-Instruct-2407, Gemma-3-27B-it, and QwQ-32B. We begin by selecting 20 scientific topics from the Microsoft Academic Graph (MAG) dataset, focusing on papers published after 2010 with at least 10 citations to ensure their relevance and significance. For each topic and condition (four in total, representing high and low levels of perspective and background diversity), we assemble a team of three members. Each simulated team (N = 20 topics × 4 conditions × 3 LLM settings = 240) are designed to have 3 members. For a given topic and condition, all LLM settings use the same set of team member profiles, ensuring that diversity levels align with the specified values across three embedding spaces. Each team engages in a 20-round process of discussion and paper writing, during which we record intermediate versions of the generated essay. Using the topic and member profiles provided to the LLMs, we construct token-level measures of knowledge integration and speculation, applying the same method used in our empirical analysis. We also calculate the distance between each generated paper—both intermediate and final versions—and its assigned topic, defined as the cosine distance between the centroid of token embeddings in the team's paper and the embedding vector of the topic. Finally, we use ChatGPT-4o to evaluate the simulated papers based on two rubrics: the NeurIPS 2024 Reviewer Guide for scientific quality and the AAC&U VALUE Rubrics for written communication. Results of this virtual experiment are reported in Figure 4 in the main text.

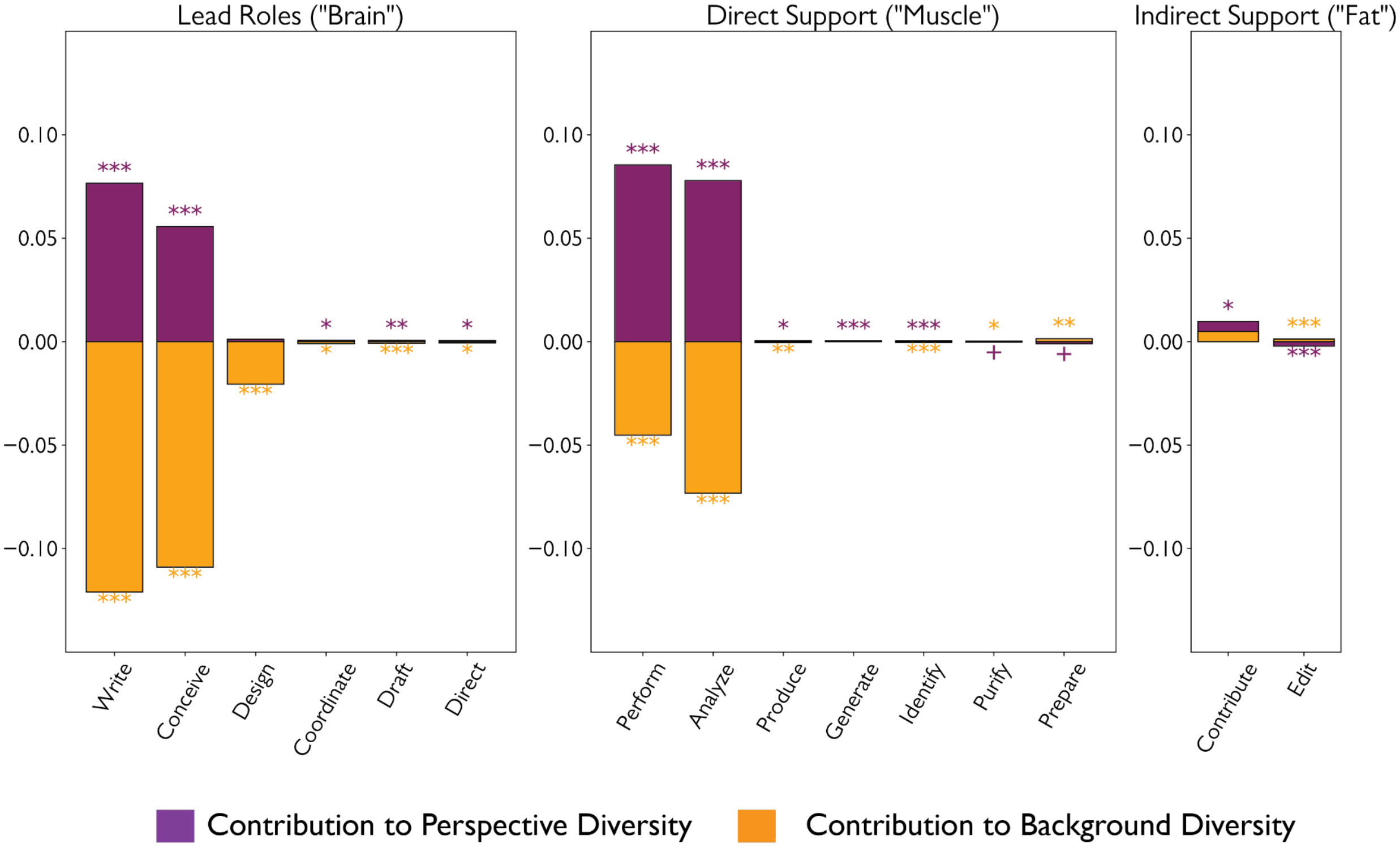

**Extended Data Figure 7. Verifying the effects of diversity contribution on collaborative roles in research teams through single-role examinations.** Drawing from 89,575 self-reported author contribution statements, collected from 13 years of *PNAS* (N = 18,354; 2003–2015), 15 years of *Nature* (N = 9,364; 2006–2020), 3 years of *Science* (N = 1,176; 2018–2020), and 9 years of *PLOS ONE* (N = 60,681; 2006–2014), we estimate whether the type of diversity an individual brings to their team influences their collaborative role. We construct data on paper–scientist level, and regress dummies for whether scientists performed each of the top 50 research activities on their diversity contribution to the team, as defined in Supplement Section 2.7. We report results only for roles where diversity contributions show statistically significant effects. Consistent with the findings in Extended Data Table 2 and Supplement Table S5, scientists who contribute a high level of perspective diversity are more likely to take on central roles, such as writing, generating ideas, and drafting the manuscript. In contrast, peripheral roles—such as preparation and editing—are more frequently taken by individuals who contribute background diversity without adding new perspectives.

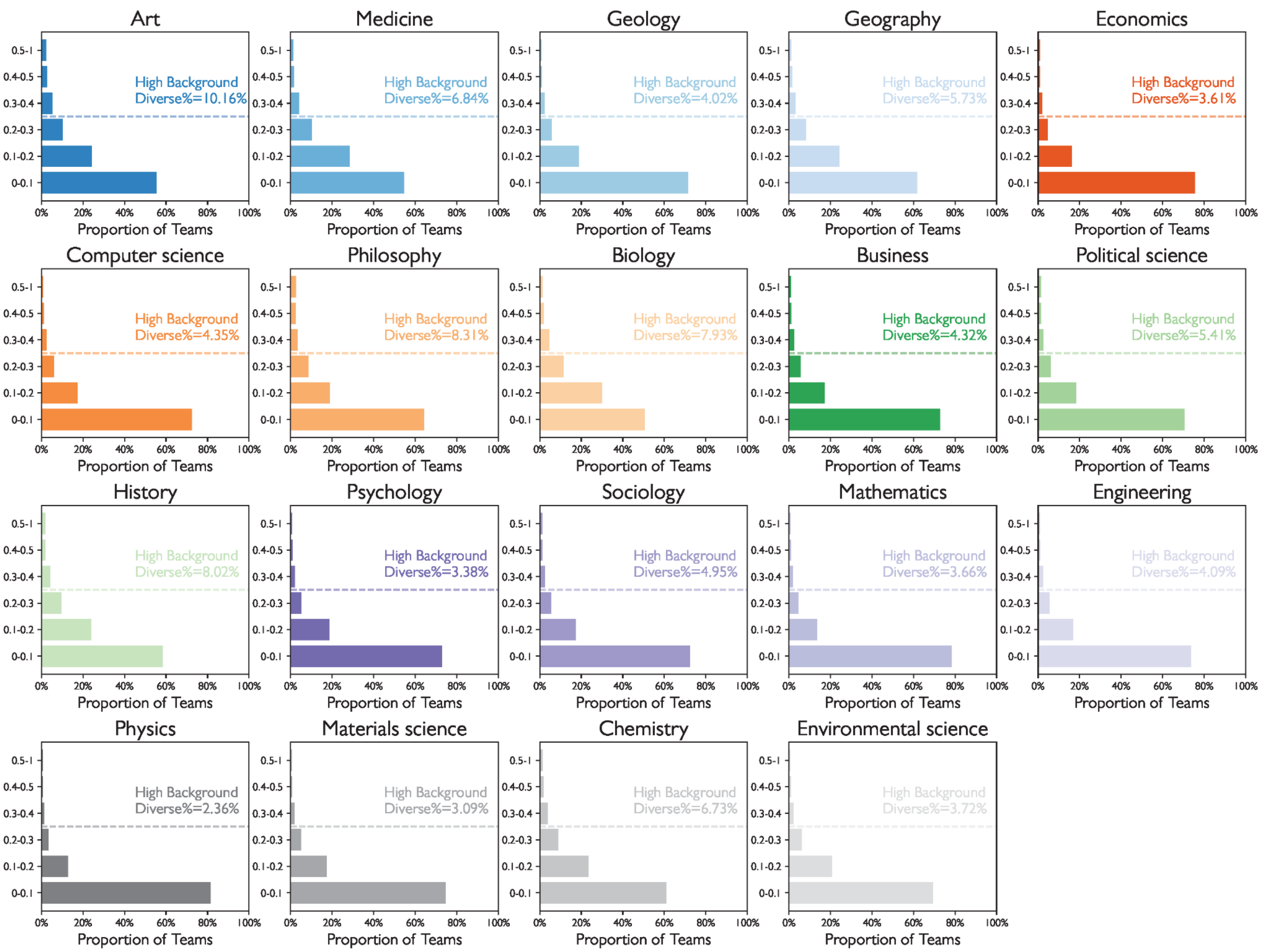

**Extended Data Figure 8. Distribution of background diversity in research teams across scientific disciplines.** Using author information from 20,405,559 research papers published between 1981 and 2015, we calculate the background diversity score for each publication and visualize its distribution across scientific disciplines. Disciplines are defined as level 0 fields in the Microsoft Academic Graph, covering a total of 19 fields. All diversity measures are calculated within the same space and therefore directly comparable. The results show that background diversity is highly right-skewed: in nearly all disciplines, fewer than 10% of research teams have a background diversity score above 0.3, with most teams concentrated in the 0–0.1 range. Despite this overall pattern, the humanities tend to have more generalist teams. High-diversity teams (with background diversity larger than 0.3) appear disproportionately in art (10.16%), philosophy (8.31%), and history (8.02%). In contrast, STEM disciplines show the lowest average background diversity, with the proportion of high-diversity teams at 4.35% in computer science, 3.66% in mathematics, 4.09% in engineering, 2.36% in physics, 3.09% in materials science, and 6.73% in chemistry.

**Extended Data Table 1. Assessing the robustness of the changing quality of innovation for the same innovator when switching from teams with different diversity compositions.** Utilizing five distinct datasets, we identified 8,704,150 scientists, 1,018,489 inventors, 5,446 investors, 12,121 screenplay writers, and 4,243,861 Wikipedia editors who have ever contributed to multiple projects. Based on their work histories, we constructed project–innovator level records, totaling 126,365,105 for scientific publications, 9,254,446 for patent applications, 62,953 for startup investments, 76,145 for movie screenplays, and for 74,783,662 Wikipedia pages. We applied fixed effects models to these records, incorporating all controls from Tables S3 and S4, and also accounting for innovator's individual fixed effects. The findings suggest that an innovator is more likely to produce high-impact work, when switching to a team where they form a team of diverse perspectives, but relatively homogeneous knowledge backgrounds.

| | Model 1 | Model 2 | Model 3 | Model 4 | Model 5 | Model 6 |
|---|---|---|---|---|---|---|
| | **Science (Citation)** | **Technology (Citation)** | **Entrepreneurship (Fundings)** | **Film (Rating)** | **Wikipedia (First-record; Page Quality Grading)** | **Wikipedia (Multi-records; Page Quality Grading)** |
| | Poisson Regression with High-Dimensional FE (pplmhdfe) | | | OLS with High-Dimensional FE (reghdfe) | | |
| Log(Perspective Diversity) | 0.254*** | -0.021 | 0.474*** | 0.672*** | 0.681*** | 0.369*** |
| | (0.071) | (0.035) | (0.066) | (0.040) | (0.001) | (0.002) |
| Log(Background Diversity) | -0.504** | -0.140** | -1.801*** | -0.913*** | -2.050*** | -1.247*** |
| | (0.159) | (0.045) | (0.168) | (0.063) | (0.002) | (0.007) |
| Log(Mean_Experience) | 0.189*** | 0.045 | 0.078*** | -0.045*** | -0.021*** | -0.005*** |
| | (0.029) | (0.048) | (0.008) | (0.010) | (0.000) | (0.000) |
| Previous Collaboration | -0.306*** | -0.288*** | -0.086*** | 0.083*** | 0.017*** | 0.009*** |
| | (0.037) | (0.010) | (0.014) | (0.008) | (0.001) | (0.001) |
| Proportion of New Inventors | -0.499*** | -0.033 | (omitted) | 0.011 | -0.092*** | -0.119*** |
| | (0.061) | (0.034) | (omitted) | (0.047) | (0.001) | (0.001) |
| Constant | 2.861*** | 1.988*** | 1.163*** | 6.036*** | 1.811*** | 2.280*** |
| | (0.138) | (0.113) | (0.033) | (0.024) | (0.001) | (0.001) |
| **Innovator-Individual FE** | YES | YES | YES | YES | YES | YES |
| Team Size FE | YES | YES | YES | YES | YES | YES |
| Year FE | YES | YES | YES | YES | YES | YES |

| Discipline/Field/Industry/Genre/Page ID FE | YES | YES | YES | YES | YES | YES |
| --- | --- | --- | --- | --- | --- | --- |
| Other Controls | | Claims, Family, References | Location, Demographics | Country, Language | | |
| Adjusted R2/Pseudo R2 | 0.503 | 0.451 | 0.114 | 0.467 | 0.404 | 0.780 |
| Observation | 1.170e+08 | 8340733 | 55195 | 59669 | 12537107 | 58998516 |

$^{*}$ $p < 0.05$, $^{**}$ $p < 0.01$, $^{***}$ $p < 0.001$. Standard errors clustered at industry/genre level.

**Extended Data Table 2. Marginal Diversity Contribution Influences Team Roles for Scientists.** We combine author contribution statements ($N$ = 18,354 for *PNAS*, $N$ = 9,364 for *Nature*, $N$ = 1,176 for *Science*, and $N$ = 60,681 for *PLOS ONE*) with the Microsoft Academic Graph to trace authors' prior experiences. Using this information, we calculate the proportion of new perspectives and different backgrounds each team member contributes to their team. We then employ author-fixed effects regressions to estimate the relationship between diverse contributions and scientists' roles in collaboration. Results show that the same scientist is more likely to take on a central leading role in a project, and less likely to occupy a marginal, indirect support one, when they contribute more unique perspectives and fewer background differences to the team, after accounting for discipline, year, and team size.

| | Model 1 | Model 2 | Model 3 | Model 4 | Model 5 | Model 6 | Model 7 | Model 8 |
|---|---|---|---|---|---|---|---|---|
| | **DV: Probability of Most Important Roles In Scientific Collaboration** | | | | | | | |
| | **Leading Role** | | | | **Indirect Sup Role** | | | |
| Marginal Perspective Diversity Contribution (Proportion) | 0.044$^{***}$ | 0.094$^{***}$ | 0.042$^{***}$ | 0.040$^{***}$ | -0.031$^{***}$ | -0.019$^{***}$ | -0.014$^{***}$ | -0.010$^{**}$ |
| | (0.007) | (0.010) | (0.005) | (0.005) | (0.004) | (0.002) | (0.003) | (0.003) |
| Marginal Background Diversity Contribution (Proportion) | -0.114$^{***}$ | -0.144$^{***}$ | -0.075$^{***}$ | -0.081$^{***}$ | 0.039$^{***}$ | 0.033$^{***}$ | 0.025$^{***}$ | 0.025$^{***}$ |
| | (0.008) | (0.012) | (0.006) | (0.006) | (0.003) | (0.002) | (0.003) | (0.003) |
| Log(Team Perspective Diversity) | 0.098$^{***}$ | 0.110$^{***}$ | 0.003 | 0.015 | -0.046$^{***}$ | -0.045$^{***}$ | -0.017$^{***}$ | -0.022$^{***}$ |
| | (0.009) | (0.007) | (0.009) | (0.008) | (0.002) | (0.002) | (0.002) | (0.002) |
| Log(Team Background Diversity) | -0.271$^{**}$ | -0.220$^{**}$ | 0.081$^{**}$ | 0.130$^{**}$ | 0.085$^{***}$ | 0.111$^{***}$ | 0.004 | -0.005 |
| | (0.075) | (0.063) | (0.025) | (0.036) | (0.007) | (0.014) | (0.019) | (0.029) |
| Log(Mean_Experience) | | 0.038$^{***}$ | | 0.019$^{***}$ | | 0.006$^{***}$ | | 0.001 |
| | | (0.002) | | (0.001) | | (0.001) | | (0.001) |
| Previous Collaboration | | -0.035$^{***}$ | | 0.040$^{***}$ | | -0.002 | | -0.016$^{*}$ |
| | | (0.006) | | (0.010) | | (0.004) | | (0.007) |
| Proportion of New Inventors | | 0.049$^{***}$ | | 0.110$^{***}$ | | -0.013$^{*}$ | | -0.048$^{***}$ |
| | | (0.005) | | (0.009) | | (0.005) | | (0.009) |
| Constant | 0.588$^{***}$ | 0.382$^{***}$ | 0.636$^{***}$ | 0.478$^{***}$ | 0.089$^{***}$ | 0.056$^{***}$ | 0.091$^{***}$ | 0.100$^{***}$ |
| | (0.005) | (0.013) | (0.001) | (0.005) | (0.001) | (0.003) | (0.002) | (0.004) |
| **Innovator-Individual FE** | | | YES | YES | | | YES | YES |
| Team Size FE | YES | YES | YES | YES | YES | YES | YES | YES |
| Year FE | YES | YES | YES | YES | YES | YES | YES | YES |
| Discipline FE | YES | YES | YES | YES | YES | YES | YES | YES |
| Adjusted R2 | 0.099 | 0.142 | 0.281 | 0.283 | 0.026 | 0.030 | 0.116 | 0.117 |
| Observation | 523633 | 523633 | 252972 | 252972 | 523633 | 523633 | 252972 | 252972 |

$^{*}$ $p < 0.05$, $^{**}$ $p < 0.01$, $^{***}$ $p < 0.001$. Standard errors clustered at discipline level.

# Supplementary Information for

# Subjective Perspectives within Learned Representations Predict High-Impact Innovation

Likun Cao, Rui Pan, James Evans

## Supplementary Methods

### S1 Data Description

In this study, we use five distinctive empirical datasets to study five types of innovative products: scientific publications, patented inventions, startup descriptions, screenplay summaries, and Wikipedia articles. These five artifacts can be arrayed along two dimensions (see Figure S1). On one dimension, we study the original innovative artifact itself (e.g., scientific papers, technology patents, Wikipedia articles), or we have a summary of the artifact (e.g., film and company descriptions). On the second dimension, the topic of the innovation is fixed or varies freely before the creatives engage with it. Scientists typically select their topic, but technology patents are often more constrained based on the demands of the company, and Wikipedia articles typically exist long before editors expand upon them. While films may be creatively imagined, their expense is often constrained (and sometimes dictated) by the production studio, and start-up companies typically exist before they receive investment and mentorship by venture capitalists.

Although both collaboration processes and products may be heterogeneous across domains, all five datasets allow us to trace (1) team composition, (2) creative history and perspective of team members, (3) original or summarized content of the focal innovation; and (4) team performance as evaluated by the primary audiences of the innovation.

### S1.1 Microsoft Academic Graph Data for Science

We utilize the Microsoft Academic Graph (MAG), one of the largest open-access bibliometric datasets available[1]. MAG includes records of tens of millions of scientific publications, including titles, abstracts, keywords, author affiliations, and journal sources. The snapshot we used, accessed in December 2021, comprises 269,813,751 entries, including scientific papers, repositories, patents, and datasets. Due to data sparsity before 1980 and after 2015, we focused on only journal articles published between 1981 and 2015. We retained only those articles with valid abstracts and at least two authors who had publishing records covering the previous five years to ensure meaningful diversity based on their past experiences. For each project, we collected information on the abstract, names, backgrounds and past experience of related scientists, as well as citations received.

The dataset initially contained 278,614,730 individual profiles. MAG's name disambiguation algorithm prioritizes the accuracy of an author's profile. Although publications in the profile are generally accurate in author attributions, the profile may not mention or link all that author's publications. In other words, publication records of the same author can be split into multiple profiles in MAG (or different authors can be merged into the same profile). To address this issue, we used density-based clustering[2,3] to identify multiple profiles linked to the same normalized author name and indexed these unique profile clusters with new author IDs. After this additional disambiguation, we identified 254,365,736 unique individuals in the dataset. The final compiled dataset includes 20,405,559 valid papers and 16,878,569 related scientists.

**S1.2 Granted Patents from the United States Patent and Trademark Office for Technology**

For patented technologies, we use the open-source dataset provided by the United States Patent and Trademark Office (USPTO). This dataset includes all historical information on U.S. patents, applications, and grants from 1976 to the present and is updated quarterly. It provides content detail about the technology and its proposed use, such as patent abstracts, and metadata on inventors and assignees, such as name and gender of inventors, and patent ownership and transfers between organizations. Following previous research[4], we define a patent team as multiple inventors listed on the same granted patent. As of the data collection date, the dataset contains a total of 8,517,464 patents, of which 6,957,544 contain valid abstracts. We retained only those patents where at least two inventors had invention records in the previous five years to ensure meaningful diversity measures and facilitate comparison with scientific publications. In the final dataset, we retained 2,841,066 valid patents with 1,844,890 related inventors.

### S1.3 Crunchbase Data for Entrepreneurship

For venture capitalists and startups, we use data from Crunchbase, a well-known crowd-sourced platform that records extensive information on startup companies and their funding sources. Crunchbase is widely used in entrepreneurship and business strategy research[5].

Crunchbase data provides detailed information on both new ventures and investors. In addition to basic details like company name, address, and industry, it includes comprehensive data on business descriptions, funding histories, and employee composition. The funding data encompasses all major investor types in the U.S., including angel investors, corporate venture capital funds, and independent venture capital funds[6].

We focused on startups based in the U.S. or with U.S. headquarters, established between January 1, 1950, and December 31, 2021. We collected funding histories and investor background information for these startups. The original dataset contains 298,915 new ventures, 174,926 funding records, and 41,170 unique investors. Of these startups, 81,765 have received funding, and 198,226 have valid business descriptions.

Building on previous research on venture capital syndication (i.e., collaborative investment in a single startup), we examine how the composition of venture capital teams affects startup survival and performance[7,8]. Venture capital teams provide not only funding but also resources, knowledge, and social networks, which can be complementary, redundant, or even conflict. Our focus is on early-stage syndicated investments, where startup performance is highly uncertain and maximally influenced by venture capital resources[9,10]. We retained only startups with (1) valid business descriptions and (2) at least two investors with investor records from the past five years to ensure meaningful diversity measures. The final dataset includes 10,181 valid startups and 11,572 related investors.

### S1.4 IMDb Film Dataset

For movie data, we use the Internet Movie Database (IMDb, www.imdb.com), a comprehensive digital database for films and television programs, which is widely used in studies on cultural products and creativity[11–15]. IMDb provides detailed records for each film, including cast and crew information, box office performance, ratings, and awards.

We downloaded the official bulk data provided by IMDb (https://developer.imdb.com/non-commercial-datasets/) and supplemented it with additional information from the IMDb API (https://imdb-api.com/). The official dataset includes basic details such as movie titles, languages, regions, genres, and crew members. The supplemental dataset offers more in-depth information, including extended plot summaries, awards, and box office figures. The dataset we used, downloaded in February 2023, includes 261,302 movies, featuring 107,743 directors, 175,952 screenwriters, and 1,218,612 actors and actresses.

Each director, producer, and actor/actress in IMDb is assigned a unique identifier. Our research focuses specifically on screenplay writers, given their crucial role in shaping plot quality and its impact on a movie's success or failure. In the dataset, each movie has 1 to 3 screenplay writers. We retained only those movies with (1) valid plot summaries and (2) at least two screenplay writers who have produced movie plots in the past five years to ensure meaningful diversity measures. The final dataset includes 29,797 valid movies and 28,556 related individuals.

### S1.5 Wikipedia Data

For Wikipedia articles, we use the bulk download service provided by Wikimedia database dumps, which are updated bi-monthly. We downloaded all English Wikipedia pages in December 2023. At the time of download, the dataset included 17,485,337 pages, 9,280,838 registered users and 5,030 bots. We keep human users in our analysis, registered or not. Based on previous research on Wikipedia open collaboration, we argue that the structure of editor teams, such as their diversity, significantly affects both project productivity and page quality[16–19]. To assess editor team structure, we excluded minor revision records that involved fewer than 100 English characters. In this way, we ensure that each edit in our sample represents a meaningful contribution to the project.

Given the continuous revisions of Wikipedia pages, we maintained a snapshot of each page on a yearly basis. We calculated user diversity and quality scores for each yearly snapshot and retained only those with at least two editors who had Wikipedia revision records in the previous two years to ensure meaningful diversity measures. The final dataset includes 5,058,522 first-year creation records of pages, 89,261,730 page-year records as longitudinal data, and 20,501,961 users.

### S1.6 Author Contribution Statements

In addition to the five main datasets, we include two supplementary datasets for the analysis of mechanisms associated with diversity and performance. First, we use a dataset of increasingly mandatory author contribution statements from four journals. This dataset includes 13 years of *Proceedings of the National Academy of Sciences* (*PNAS*; 18,354 statements from 2003–2015), 15 years of *Nature* (9,364 statements from 2006–2020), 3 years of *Science* (1,176 statements from 2018–2020), and 9 years of *PLOS ONE* (60,681 statements from 2006–2014). We classify the top 50 research activities within author contributions into categories including leadership (e.g., "conceived" of the study), direct support (e.g., "analyzed" data) and indirect support (e.g., "commented on" the manuscript) following the protocols established by Xu et al[20]. These activities account for 95.37% of all activities at the individual level.

After extracting contribution information, we linked it to the MAG data. Contribution statements often use name initials (e.g., "JE: conceptualize, write, design"), which facilitates author identification. Scientists not mentioned in the contribution statements or not identified successfully through our procedure are excluded from models in the mechanism analysis.

## S2 Model Details

### S2.1 Modeling the Concept Spaces for Invention Emergence

To construct conceptual spaces across multiple time periods, we use a dynamic embedding algorithm [21] specifically designed for longitudinal corpora. This approach assumes and enforces a smooth transition between word vectors and their associated meanings in adjacent time periods and trains all periods jointly. Specifically, it alters the objective for the joint word embedding to the following:

$$\min_{U(1),\ldots,U(T)} \frac{1}{2}\sum_{t=1}^{T} \|Y(t) = U(t)U(t)^T\|_F^2 + \frac{\lambda}{2}\sum_{t=1}^{T} \|U(t)\|_F^2 + \frac{\tau}{2}\sum_{t=2}^{T} \|U(t-1) = U(t)\|_F^2 \quad (1)$$

The first term in this objective, the $L_2$ norm of the difference between the empirical word proximities $Y(t)$ and the modeled proximities $U(t)$, represents the standard word embedding objective. This objective guides optimization of the algorithm to accurately model word distances within time periods, and the $L_2$ norm regularizer is used, as in ridge regression, to improve out-of-sample predictive performance of the model. Dynamic word embeddings add two additional terms that: 1) allow the algorithm to minimize unnecessary word vector loadings (the $L_2$ norm of $U(t)$), and 2) shrink the distance between time-adjacent embeddings (the $L_2$ norm of the difference between $U(t\text{-}1)$ and $U(t)$).[1] In minimizing this cost function, adjacent word embedding spaces are grown in alignment, with a $t \times n \times k$ matrix, where *t* is the number of time slices, *n* is the number of words, and *k* is the dimensions of word vectors. The word embeddings of period *t* is stored in one $n \times k$ slice, which is optimized to remain similar to the *t*-

---

[1] This dynamic word embedding approach relies on the formal equivalence between the standard negative sampling approach to word embedding optimization used by `word2vec` and the low-rank factorization of a pointwise mutual information (PMI) matrix of word vectors (Levy and Goldberg, n.d., 2014), and uses the PMI as input to the objective above.

1 and $t$+1 matrices. In this way, the jointly trained algorithm keeps adjacent semantic spaces close in structure. This approach allows us to construct comparable and historically accurate representations of conceptual structures for each time period. We learn 50-dimensional representations for each word using a window size of 5 and perform 10 iterations, which is more than the number suggested in the method paper to ensure convergence.

Given that low-frequency words can lead to unstable embeddings, we set a frequency threshold of 150 for word inclusion. To address instability in the IMDb and Crunchbase data due to their smaller volume, we assessed screenplay summaries and venture descriptions within the context of other, comparable semantic spaces. For IMDb, we use Google *n*-grams to construct a cultural space defined by fiction and nonfiction, within which we project the screenplay summaries. For Crunchbase, we use 118 financial newspapers (e.g., *Wall Street Journal*, *Financial Times*, *Bloomberg Businessweek*) from ProQuest to construct a commercial space, within which we projected the new venture descriptions. The science, technology, and Wikipedia spaces were trained on their respective original datasets. For the science, technology, entrepreneurship, and film cases, we use a 5-year time window, while for Wikipedia, we use a 2-year time window to capture the fast-paced changes and frequent updates typical of (hyper)active online platforms.

**S2.2 Validating Semantic Space for Innovation Emergence**

We first validate our dynamic semantic space with a predictive task. If the embedding spaces accurately represent how conceptual elements evolve over time, we would expect that areas where elements flow in and interact are precisely where innovations are most likely to emerge, consistent with theories of recombinant innovation[22–24].

For each case, we randomly select 5,000 points in the space and track the average "in-flow" of local concepts, defined by the increase in cosine similarity between points and nearby concepts between two time periods. To focus on relevant, nearby concepts, we only consider those close enough to the points (with cosine similarity greater than threshold $t1$). We cluster the concepts within that range using the density peak algorithm [2,25], then establish whether the centroids of those clusters are, on average, moving closer or further from the focal point. Next, we then record the number of innovations (e.g., scientific papers, technology patents, film screenplays, company descriptions, Wikipedia pages) that appear within a specific range (with cosine similarity greater than threshold $t2$, where $t2$ may or may not equal $t1$) in the second period. If our hypothesis holds, we expect a positive correlation between the increasing proximity of concepts between periods and the number of innovations in the second period. Correlations between local concept convergence and innovation bursts are always positively and significantly correlated across all contexts and robust to variation in the selection of $t1$ and $t2$, as illustrated in Extended Data Figure 1. Robustness tests with different t1 and t2 parameters are presented in Figures S2 to S6, and the results remain consistent with those of the main model.

Given that space dynamics can both drive and be driven by innovation, we conducted a Dumitrescu & Hurlin [26] Granger Causality Test to examine the possibility of a causal relationship between the "in-flow" effect in semantic space and innovation emergence. This analysis was performed on a panel of 2,000 randomly selected subspaces. Due to the test's requirement of at least eight time periods, it was only applied to the science, entrepreneurship, film, and Wikipedia cases, excluding the technology case. Results, presented in Table S1, indicate that, overall, spatial dynamics are a much stronger driver of innovation than the reverse, showing the dominance of structural forces over agency.

The second validation test is conducted at the level of individual innovators. If the geometric relationships between individuals and concepts are meaningful, we would expect innovators to be more likely to adopt a concept when (1) the concept moves closer to their own position over time, and (2) this movement is more visible to them—measured by the visual angle between the concept's starting and ending points.

We test these hypotheses using a series of fixed effects regressions. Within each dataset, we examine concept adoption behaviors across periods at the innovator–concept–period level. From period t to period t+1, we measure: (1) the change in cosine similarity between the author and the concept, where a positive value indicates increased proximity; (2) the "visual angle" of concept movement, defined by the angle between its starting and ending points with respect to individual observers; and (3) whether the author adopts a new concept (not used in t) in t+1. The adoption outcomes are coded as binary variables. We estimate the models with author and period fixed effects, using the PyFixest package in Python 3.

Details on measurement, model specification, and results are presented in Extended Data Figure 2 and support our hypotheses.

### S2.3 Quantifying Diversity

It is common practice to use the centroid or high-dimensional average value of word vectors to represent the position of one or a body of creative works within the semantic space of the broader culture[15,27]. We follow this approach by calculating an innovator's experience vector as the average of word vectors from their previous engagements—publishing science, inventing technologies, financing businesses, writing movies, and drafting Wikipedia pages. For the first four cases, we use records from the past 5 years for this calculation to account for the decay in expertise and the slow movement of interests over time[28]. For Wikipedia, we use a 2-year window due to the rapid evolution of online platforms. We also conduct robustness checks with other time windows, and the results remain consistent.

Given a set of experience vectors in a team $\{V1, V2, V3, ...., Vn\}$, we measure the team's background diversity as:

$$BD = \frac{1}{n \times (n-1)} \sum_{i=1}^{n} \sum_{j=1, j \neq i}^{n} CosineDistance(V_i, V_j)$$

,

This captures the average pairwise distance between members' experience vectors.

Meanwhile, we measure the perspective diversity of the team using perspective vectors, which reflect the relationship between an innovator's previous experience and the focal work they are constructing together. In word embedding space, vectors often capture logical relationships between entities. For example, the vector from "Washington" to "US" represents the relationship "is the capital of", and this vector is typically parallel to the vector from "Beijing" to "China". Similarly, the vector from "man" to "king" is often parallel to the vector from "woman" to "queen"[29,30], reflecting the parallelogram model of semantics [31,32].

Therefore, when collaborators' past experiences point to the focal task from different directions, we assume they bring distinct prior experiences to the work and can offer diverse perspectives. Conversely, if their experience vectors converge on the focal task from similar directions, their viewpoints may be more consistent, even if their backgrounds can be diverse (e.g., a painter, a singer, and a historian collaborating on a computer science project).

Thus, we define the perspective vector for team member $i$ as:

$$V_{p,i} = V_{task} - V_i.$$

A team with experience vectors $\{V1, V2, V3, ...., Vn\}$ will have a corresponding set of perspective vectors $\{V_{p,1}, V_{p,2}, V_{p,3}, ...., V_{p,n}\}$, based on this equation.

The perspective diversity of a team can therefore be defined as:

$$PD = \frac{1}{n \times (n-1)} \sum_{i=1}^{n} \sum_{j=1, j \neq i}^{n} CosineDistance(V_{p,i}, V_{p,j})$$

Note that these two variables are interrelated. A higher background diversity creates a broader range of potential directions for task vectors, leading to higher perspective diversity. This aligns with our intuition and reflects the real-world scenario where teams with diverse backgrounds often synthesize complementary insights from their members.

The correlation between background diversity and perspective diversity across all five datasets is shown in the main text. The coefficients, all positive and significant, range from 0.25 to 0.66. To address this correlation, we perform a Variance Inflation Factor (VIF) test after each regression and find no issues with collinearity in our models.

### S2.4 Validating Diversity Measures

We used two vector-based variables to capture background and perspective differences between team members. These measurements utilize the vector-based structure of embeddings and are designed to reflect the underlying theoretical constructs. To validate the operationalization, we conducted additional validation tests, as reported in Figure S7.

First, we validate our measure of background diversity. Our theoretical framework suggests that when innovators share similar past experiences, background diversity should yield a low value. To examine this, we randomly sample a pair of scientists from each scientific paper, calculate the Jaccard distance (1 – Jaccard similarity) between all tokens associated with their prior work. The Jaccard similarity reflects the number of tokens their background work shares over their total number background tokens. We then correlate this ground-truth against our background diversity measure. As shown in the upper panel of Figure S7, the two measures exhibit a Spearman (rank order) correlation of approximately 0.7. Overall, background diversity reflects differences in scientists' past creative trajectories: Individuals with many overlapping tokens (low Jaccard distance) consistently exhibit lower background diversity than those with fewer overlaps. This suggests that our measure effectively captures variation in innovators' expertise backgrounds.

Second, we validate perspective diversity. Our theoretical framework suggests that individuals with similar perspectives on a task are likely to contribute similar ideas. To test this, we conducted two analyses in the science domain. First, we identified scientists who contributed to multiple papers and examined the words they contributed to each. We defined a scientist's word contribution as the set of words in the final paper for which they had a substantially higher tf-idf score (at least a 0.05 difference) than their collaborators, suggesting that they use those words much more frequently than their collaborators and the scientific population as a whole[33]. As shown in the middle panel of Figure S7, the same author tends to contribute similar words to two projects when they hold similar perspectives (≤30 degrees), even if the co-authors differ. In contrast, when their perspectives on the two projects differ substantially (≥90 degrees, which signifies semantic unrelatedness)[34], their word contributions vary significantly. This pattern holds even when the distance between projects is controlled.

We extend this analysis with a more stringent test, examining whether *different* authors will contribute similar words when they share similar perspectives. We extract paper-author pairs with either similar or divergent perspective vectors. The results, shown in the lower panel of Figure S7, support our argument: When two authors share similar perspectives (≤30 degrees), their word contributions are significantly more similar than when their perspectives differ substantially (≥90 degrees). This result holds even when the distance between authors is controlled.

**S2.5 Quantifying Team Performance**

Given that teams in our five cases have different goals, we evaluate their performance using domain-specific criteria, as extracted from previous studies.

For science and technology teams, we use five-year forward citations as a proxy for impact, following widely used methods from the science of science and informetrics[4,35].

For startups, we adopt criteria from the entrepreneurship literature[36,37] and measure performance by the total number of funding rounds received from investors. Additionally, we evaluate long-term success based on milestone events, including initial public offerings (IPOs) and high-value acquisitions.

For films, we evaluate their quality using average IMDb ratings, reflecting judgments from popular audiences. To complement this with professional evaluation, we also include a dummy variable for whether the movie won an award, across the wide range of film festivals (e.g., Oscars, Golden Globes, and Tribeca Film Festival awards). 42% percent of IMDb receive some form of award.

For Wikipedia pages, we use the page-quality assessment algorithm available through the official Wikipedia API. This algorithm was initially trained on widespread assessment of expert ratings regarding article content, and is blind to the process of article creation, found in the Wikipedia talk page. This score is widely utilized by Wikipedia editors for evaluating article quality, and is also frequently used in scientific research on open source knowledge production[38].

To complement this score, we also include annual page views as an indicator of page popularity.

**S2.6 Quantifying Marginal Diversity in Scientific Teams**

In our analysis of mechanisms driving the benefits of close, diverse perspectives across domains, we estimate whether each team member's diversity contribution increases their probability of taking on different roles. To measure the marginal diversity contribution for each individual, we construct a counterfactual scenario where the team continues to function without that specific member.

For example, consider a team of four members: *a*, *b*, *c*, and *d*. We first calculate the background diversity and perspective diversity based on the full team, denoted as $BD_{full}$ and $PD_{full}$.

Next, we focus on one specific team member, say member *a*. We recalculate the diversities excluding this member, using only the data from the remaining three members (e.g., *b*, *c,* and *d*). These recalculated diversities are denoted as $BD_{-a}$ and $PD_{-a}$. The equations we use to compute these values are the same as those applied to the full team, but the vectors for member *a* are excluded in the counterfactual team.

The marginal diversity contributions of member *a* can be calculated as the proportion of the difference between the full team's diversities and those of the counterfactual team:

Marginal Contribution to Background Diversity

$$MBD_a = \frac{(BD_{full} - BD_{-a})}{BD_{full}}$$

Marginal Contribution to Perspective Diversity

$$MPD_a = \frac{(PD_{full} - PD_{-a})}{PD_{full}}$$

These measures capture the proportion of diversity that a focal individual contributes to the original team, compared to a counterfactual team without that individual.

### S2.7 Analyzing the Probability of Roles with Marginal Diversities in Scientific Teams

We argue that a team member's marginal diversity significantly contributes to their role within the team. To test this hypothesis, we analyze empirical data from author contribution statements across four journals: *Nature* (9,364 entries from 2006–2020), *Science* (1,176 entries from 2018–2020), *PNAS* (18,354 entries from 2003–2015), and *PLOS ONE* (60,681 entries from 2006–2014). Team roles were identified by clustering the role co-appearance network using the Louvain community detection algorithm, following methods in Xu et al[20] (see Fig. 3). Based on this clustering, all team roles were categorized into three categories, with a modularity score of 0.111: *leaders* who "conceive", "plan", "initiate", "direct", "supervise", "coordinate", or "write" their articles; *direct supporters* who "obtain", "collect", "synthesize", "prepare", "process", "purify", "build", or "carry" for their articles; and *indirect supporters* who "participate", "discuss", "comment", "review", and "edit" on their articles. By definition, the "leader" role signifies the highest level of contribution, followed by "direct supporter" and then "indirect supporter". We define a team member's role based on the highest level of contribution they make. For example, if a scientist is involved in both conceiving (a leader task) and analyzing (a direct supporter task), we classify her as a leader because she has participated in the core leadership of the project.

### S2.8 Quantifying Knowledge Integration and Knowledge Speculation

One of our core arguments is that when background diversity is low and perspective diversity is high, more team members will be actively engaged, and the team's output will better integrate their past experiences, resulting in a more original and refined final product. Section 2.5 details our approach to quantifying team members' engagement levels. In this section, we focus on finding a proxy to measure knowledge integration within each team project by using knowledge modules.

Most innovative products involve multiple knowledge "modules". The composition of these modules, according to the findings of innovation studies[23,24], represents the project's knowledge structure. This structure is shaped through team collaboration, where members exchange ideas, discuss, and negotiate potential solutions, ultimately deciding which modules to include or exclude.

We argue that the final choice of modules reflects the team's level of knowledge integration. When all team members participate in decision-making and contribute effectively, the team is more likely to select modules that align with their common and collective experiences. Conversely, if the team does not collaborate effectively, they may end up using modules that are familiar to only one or two members, or those entirely new to the team. This hypothesis reflects prominent findings about the collective intelligence of teams that finds a strong, positive correlation between team solutions and the evenness of the distribution of contributions to the collaboration. Conversations that include more equal contributions from more members perform better at arbitrary problem-solving tasks [39].

Following this intuition, we constructed two measures: Knowledge *Integration* and *Speculation*. The concepts and operationalization of these measures are illustrated in Figure S8. *Integration* captures to what extent the knowledge modules overlap with team members' previous knowledge modules on average. *Speculation* captures the proportion of knowledge modules fresh to all team members.

To formalize, we operationalize the two variables as below:

$$Integration = \frac{1}{N_c \times N_p} \sum_{c=1}^{N_c} \sum_{p=1}^{N_p} (I_{c,p} == 1),$$

$$Speculation = \frac{1}{N_c}\sum_{c=1}^{N_c}(\sum_{p=1}^{N_p} I_{c,p} == 0)$$

where $N_c$ represents the total number of relevant knowledge modules in the innovation product, $N_p$ represents the number of individuals in the team, and $I_{c,p}$ is a dummy variable indicating whether innovator $p$ has previously participated in category $c$, previous to this team project (1=Yes, 0=No). We use these measures to analyze how diversity impacts the mechanism of knowledge integration. The operationalization is visually illustrated in Figure S8.

**S2.9 Estimating the Relationship between Multi-Dimensional Diversity and Team Performance**

Based on the measurements described, we calculate two types of diversity and team performance for each team. We then regress team performance on these diversity measures. The basic form of the regression equation is as follows:

$$TeamPerformance_i = \beta 0 + \beta 1 \times PD_i + \beta 2 \times BD_i + Controls_i + FE_{teamsize} + FE_{year} + FE_{subfield} + \epsilon,$$

where $TeamPerformance_i$ represents the performance measure for team, $PD_i$ and $BD_i$ are perspective and background diversity measures for team $i$, and $Controls_i$ represents controls on the team level. Control variables include the average experience of team members (in log scale)[40], the proportion of pairwise members who have ever collaborated in the past [41,42], and the proportion of new members without a prior record of knowledge-production. We also account for team size fixed effects, year fixed effects and subfield fixed effects for all the cases. Additionally, each case includes specific controls as listed below:

**Science**. For science publications, subfields are defined by disciplines (level 1—292 in total) within MAG. Papers covering multiple disciplines are labeled with the first discipline in order to preserve the discipline fixed effects estimation.

**Technology**. For patents, control variables include: (1) Number of claims; (2) Number of other patents in the same family; (3) Number of citations to non-patents (i.e., scientific documents). For patents, subfields are defined by Cooperative Patent Classification ("CPC") categories, and patents covering multiple categories are labeled with the first category in order to preserve the category fixed effects estimation.

**Film**. For movies, control variables include: (1) Dummy for language; (2) Dummy for country. For movies, subfields are defined by genres, as labeled in the IMDb dataset. When a movie spans multiple genres (which is rare), it is labeled with the first in order to preserve the genre fixed effects estimation.

**Entrepreneurship**. For startups, control variables include: (1) Dummy for headquarter location; (2) Dummy for founder's gender (0=Male, 1=Female) and (3) Dummy for the founder's race (0=White, 1=Minority). For startups, subfields are defined by industry codes in Crunchbase[2], and startups involved in multiple industries are labeled with the first in order to preserve the industry fixed effects estimation.

**Wikipedia**: We did not control for subfield fixed effects in Wikipedia pages, because Wikipedia lacks a mutually exclusive classification system and each page can have multiple overlapping labels that are interconnected through inclusion relationships[3]. Instead, for a subset of models, we control for page ID fixed effects to account for variations in content across different versions of the same article. This approach provides stricter control by addressing potential confounders related to the specific nature of knowledge.

---

[2] As listed in : https://support.crunchbase.com/hc/en-us/articles/360043146954-What-Industries-are-included-in-Crunchbase

[3] For details, see: https://en.wikipedia.org/wiki/Help:Category

Performance measures in our five cases vary in format. Some are zero-inflated count data (such as science citations, patent citations, and total rounds of investments for startups), for which we estimate negative binomial regressions. Others are binary outcomes (such as the final success of startups and award-winning status among movies), for which we use logistic regressions. Additionally, some measures are continuous (such as average movie ratings and quality scores for Wikipedia pages), for which we apply OLS models. The results of the basic models are presented in Table S3 and Table S4.

We also conduct robustness tests by estimating our equations with additional fixed effects or across different sample groups, as detailed in the following section (S3).

**S2.10 Mechanism Analysis**

After estimating the central relationships linking background and perspective diversity with performance, we analyze the mechanisms involved. The first mechanism analyses focus on science contribution statements and the role of individual scientists. We regress a dummy variable indicating whether a scientist holds a leading role in the team (versus a direct or indirect support role), or an indirect support role in the team (versus a leading or indirect support role) on their marginal diversity contribution (see S2.7). We specified the model as follows:

$$LeadingRole_{i,p} = \beta 0 + \beta 1 \times MPD_{i,p} + \beta 2 \times MBD_{i,p} + Controls_{i,p} + FE_{teamsize} + FE_{year} + FE_{subfield} + FE_{scientist} + \epsilon,$$

$$IndirectSup_{i,p} = \beta 0 + \beta 1 \times MPD_{i,p} + \beta 2 \times MBD_{i,p} + Controls_{i,p} + FE_{teamsize} + FE_{year} + FE_{subfield} + FE_{scientist} + \epsilon.$$

where *MPD* is the marginal perspective diversity contributed by the innovator to the team and *MBD* the marginal background diversity contrived to the team. Results from these models are detailed in Extended Data Table 2. In Table S5, we report an alternative specification of the model in which leading roles are compared with direct support roles, and indirect roles are compared with direct support roles, manifesting the same pattern of results. To further investigate the effect of marginal diversity contribution on team roles, we regressed dummy variables for each role on marginal diversity contribution. Findings are illustrated in Extended Data Figure 7.

Next, we regress the level of knowledge integration within a team on the diversity measures using OLS.

$$Integration_i = \beta 0 + \beta 1 \times PD_i + \beta 2 \times BD_i + Controls_i + FE_{teamsize} + FE_{year} + FE_{subfield} + \epsilon,$$

$$Speculation_i = \beta 0 + \beta 1 \times PD_i + \beta 2 \times BD_i + Controls_i + FE_{teamsize} + FE_{year} + FE_{subfield} + \epsilon.$$

Results of these models are presented in Tables S6 and S7. Overall, perspective diversity within teams substantially and consistently increases knowledge integration, while background diversity reduces it. This pattern holds across cases.

## S3 Analysis Validation and Robustness

### S3.1 Robustness Test of Main Effect on Subsamples or with Additional Controls

To ensure the robustness of our main conclusions, we estimate several additional models. First, we replace the continuous measures of background diversity and perspective diversity with two categorical variables based on equal quantile cuts. This adjustment allows us to test whether the effects observed in the original model are linear. Results reported in Extended Data Figure 3 show that the effects are indeed linear, with no marked nonlinear (e.g., U-shaped) effects.

Second, we estimate models for subgroups of teams, categorized by size. This approach enables us to rule out the possibility that team size is driving both diversity and performance, as suggested by Wu et al[4]. The results across different team sizes remain consistent, reinforcing the robustness of our conclusions. Results from this additional analysis are reported in Extended Data Figure 4.

Third, we test alternative performance measures for entrepreneurship, film, and Wikipedia cases. For startups, we use milestone events such as IPOs or high-value acquisitions. For movies, we use award-winning status. For Wikipedia, we use total annual page views, with a logarithmic transformation to mitigate overdispersion (this variable is limited to the 2012–2023 period due to data availability). We re-run our models using these alternative dependent variables, and the results, shown in Table S8, remain consistent with our main findings.

Fourth, we construct measures of team composition for patents situated in the scientific space. Science has long been recognized as a driving force and pioneer of technological advancement [43]. This robustness check provides additional evidence that helps rule out an alternative explanation—namely, that innovators shape the structure of space, rather than being influenced by it. The results, reported in Table S9, are overall robust.

Fifth, we include the distance between the team experience centroid and the focal task as a control variable in our models. As reported in Table S10, while the effect sizes of the key predictors decrease slightly, a substantial portion of the effects remain significant and consistent, reinforcing our main findings.

Sixth, we include innovator fixed effects in our models to account for potential confounding from individual differences in personality and experience. The results, reported in Extended Data Table 1, align with our main findings, confirming the robustness of our conclusions.

Seventh, we control for the cross-period convergence of team members' experiences, defined as the average decrease in cosine distance between each member's position and the central topic from time t to time t+1. By accounting for this factor, we estimate the "clean" effects of perspective diversity and background diversity, isolating them from the influence of experience breadth in prior work trajectories. As shown in Table S11, the results support our argument. Although the effects of experience convergence vary across cases, the main effects in our analysis remain robust after this adjustment.

Eighth, when constructing our diversity measures, we originally did not include new innovators without prior work records, as their knowledge trajectories could not be tracked. This may overlook the contributions of junior innovators. To address this limitation, we re-estimate our models using only teams composed entirely of experienced innovators. After excluding all teams that include any new innovator, we replicate our analyses and report the results in Table S12. The findings remain robust and consistent with those of our main models.

Finally, we examine the scenario in which team dynamics may influence the positioning of the focal product. In such cases, one important potential confounder is the variation in team members' prior experiences, as more experienced individuals may steer the topic toward their areas of expertise. To account for this, we replicate our models while controlling for the standard deviation of team experience. As shown in Table S13, the results remain consistent with our main findings, reinforcing the robustness of our conclusions.

### S3.2 Further Supporting Evidence from Science Teams

We conducted additional statistical analyses to further support our arguments. Our findings indicate that as scientists gain experience, they are more likely to form research teams with optimal structure—characterized by high perspective and low background diversity, as demonstrated in this study (see Figure S9). This trend may arise either consciously, through deliberate team design, or unconsciously, as scientists develop a more accurate sense of "who can be valuable collaborators" across their careers. In either case, our results suggest that scientists tend to become increasingly adept at team formation.

Although individual scientists' experience has led to more effective teams, the overall experience of the scientific community has not yielded similar benefits. While average perspective diversity in scientific teams increased before 1990, we observe a steady decline in perspective diversity over the past three decades, accompanied by a decrease in background diversity (see Figure S10). This trend underscores the double-edged nature of scientific specialization: while it enables individuals to focus on relevant topics and make significant contributions, it also limits access to complementary insights and reduces the potential for creative conflict, thereby diminishing opportunities for outsized innovation.

The difference in team composition also varies by discipline. As expected, teams in the humanities are more likely to exhibit greater background diversity, while STEM teams tend to be more specialized (see Extended Data Figure 8).

## S4 Experiments

Our diversity measures—particularly perspective diversity—may be shaped by team interaction processes. Knowledge workers, such as scientists, inventors, and screenwriters, often engage in intensive discussions during their collaborative efforts. These dynamics can simultaneously influence both team diversity and the quality of outcomes. In addition, some teams are systematically missing from our statistical samples, including those that underperform and leave no traceable outputs. To address the potential bias this could have on our inferences, we employ two separate experimental designs to mitigate endogeneity, and to begin to establish causal relationships.

### S4.1 Natural Experiment with Wikimedia Central and Eastern Europe (CEE) Campaign

We begin by examining one of the most prominent and influential initiatives on Wikipedia—the Central and Eastern Europe (CEE) Campaign—as a natural experiment. Initiated by Wikimedia communities across the Central and Eastern European region, this annual campaign encourages the creation and improvement of articles related to the history, culture, and notable figures of CEE countries. Held each year from March 21st to May 31st, the campaign spans 70 days and brings together contributors from diverse linguistic and national backgrounds.

This campaign, widely advertised across Wikipedia during the spring period, attracts editors from outside the CEE communities to participate in editing efforts. By disrupting established social circles and interest groups, it significantly alters both the background and perspective diversity of editing teams. As such, it serves as an external shock that changes team composition without directly affecting the quality of editor contributions.

We identified the target pages directly from open CEE records available online[4]. Between 2017 and 2024, we gathered 495,262 editing records from 28,907 pages associated with the CEE campaign. Our analysis centers on the campaign's active period, spanning March 21 to May 31 each year. Initially, we employed a coarsened exact matching (CEM) approach, utilizing editing records from the three months *preceding* the campaign. At the page-year level, we matched pages based on year, country topic, page popularity, and characteristics of the editing team in the preceding period, such as perspective diversity, background diversity, team size, and team scores, all categorized into groups.

Subsequently, we investigated whether an increase in perspective and background diversity during the campaign period (defined as treatment = 1 if diversity exceeds the median, 0 otherwise) predicts page quality at the end of the campaign period (i.e., quality score of the pages on May 31 of each year).

Extended Data Figure 5 illustrates the temporal changes in the number of CEE-related edits, the distribution of editor familiarity, and the regression outcomes based on the Coarsened Exact Matching.

### S4.2 Simulated Randomized Experiment with Large Language Models

Although natural experiments partially address endogeneity—especially from factors such as topic drift during team collaboration—they do not offer the complete randomization typically found in laboratory settings. To address this limitation, we develop another experiment—a controlled virtual experiment using a multi-agent large language model (LLM) simulation, to model team collaboration under lab-like conditions.

---

[4] An example can be found at: https://meta.wikimedia.org/wiki/Wikimedia_CEE_Spring_2024/Article_Lists

Large language models have been shown to possess the ability to predict and simulate fundamental psychological traits, political preferences, and social behaviors of humans [44–47]. In multi-agent settings, they also exhibit the capacity to communicate, collaborate, and pursue shared goals—such as navigating unfamiliar environments, retrieving information, and self-organizing to establish regulation and minimize anti-social or toxic conduct [48–50].

Building on these state-of-the-art findings, we design a virtual experiment based on interactions among multi-agent LLMs. We first construct teams according to the design principles outlined in this paper and supported by our statistical results. The attributes of these virtual teams are then fed into the LLM to initiate a simulated discussion, prompting the agents to collaboratively produce a shared output. With all other potential confounders held constant and key predictors randomly assigned, the relationship between team design and the quality of the collaborative outcome can be interpreted as causal.

### S4.2.1 Model Overview

In our simulation, we utilize three open-source large language models: Mistral-Nemo-Instruct-2407, Gemma-3-27B-it, and QwQ-32B. We needed to use open source models so that we could identify the embedded position of the scientists that we assembled into three-person teams that were high and low on perspective and background diversity. Top performing closed source generative models (e.g., GPT, Claude, and Gemini) could not enable the design of teams composed of specific perspectives.

**Mistral-Nemo-Instruct-2407** is a state-of-the-art 12-billion-parameter language model developed collaboratively by Mistral AI and NVIDIA. It comprises 40 transformer layers, each with 32 attention heads and a hidden dimension size of 14,336. The model offers a substantial context window of up to 128,000 tokens, enabling the processing of extensive and complex information. It excels in various tasks, including reasoning, world knowledge, coding accuracy, and multilingual applications.

**Gemma-3-27B-it** is a powerful language model developed by Google, featuring 27 billion parameters. Optimized to handle complex tasks with high accuracy, it incorporates advanced technologies and supports a context window of up to 128,000 tokens. It has 40 layers, with 5,120 units, 14,336 hidden dimensions and 32 attention heads per layer. As of April 2025, when this analysis was conducted, this model was recognized as one of the top-performing models available.

**QwQ-32B** is an advanced reasoning model developed by Alibaba's AI research team. Built upon the Qwen2.5-32B foundation model, it incorporates reinforcement learning techniques to enhance its reasoning and problem-solving capabilities. The model comprises 64 transformer layers, each with 40 attention heads for queries and 8 key-value heads, utilizing Grouped Query Attention (GQA). Each layer has a hidden size of 5,120 units, and the model supports a context length of up to 131,072 tokens.

### S4.2.2 Designing Virtual Teams

We select 20 scientific topics from the Microsoft Academic Graph (MAG) dataset, focusing on papers published after 2010 with at least 10 citations to ensure the relevance and significance of each topic. For every selected topic, we assign four experimental conditions based on our theoretical framework: teams with (1) high perspective and high background diversity, (2) high perspective and low background diversity, (3) low perspective and high background diversity, and (4) low perspective and low background diversity.

These teams are assembled from a pool of 20,000 real-world scientists, whose profiles are also drawn from the MAG dataset. We begin by extracting the abstracts of their most representative works and feed this information into large language models, to generate embeddings through forward propagation. These embeddings represent each scientist's position in a semantic space, reflecting their intellectual orientation within a knowledge landscape as understood and encoded by artificial intelligence. From the multi-layered embeddings, we select the final layer, as it captures the most semantically and conceptually rich representation of the input text. Between the three LLMs that we use, the correlations of cosine distances between pairs of scientists all exceed 0.8, indicating a generally consistent semantic structure.

We identify candidate team compositions through a stochastic search process. First, we calculate background differences—based on cosine distance—between all candidate scientists. For each topic, we also compute all pairwise perspective differences. We then randomly select three candidates from the scientist pool to form a team. A team is classified as having high perspective diversity if all pairwise PD values exceed the 70th percentile of the PD distribution, and as having low perspective diversity if all pairwise values fall below the 30th percentile. Similarly, a team is considered to have high background diversity if all pairwise BD values are above the 70th percentile of the BD distribution, and low background diversity if they are below the 30th percentile. These thresholds are based on two principles: (1) to create a buffer zone between high and low diversity groups, enhancing the contrast between conditions; and (2) to ensure that most topics can have team samples across all four conditions. For each LLM, BD and PD are calculated within its own embedding space. A team is classified as high diversity only if it exceeds the high diversity threshold across all three LLMs, making this a conservative experimental setting. In total, we conduct simulations for 240 teams, based on 20 topics, 4 conditions, and 3 LLMs, with the LLMs sharing topic and team profiles. This stochastic search for team members continues and automatically stops once one virtual team is identified for each condition within the same topic.

### S4.2.3 Simulated Team Interaction

For each virtual team—defined by its assigned topic, selected members, and designated LLM—we simulate team collaboration using the AutoGen framework. AutoGen is an open-source platform that enables researchers and developers to build multi-agent LLM applications in which agents can interact with one another to complete tasks [51]. This framework has been widely used in previous studies on multi-agent LLM interaction and collaboration.

We provide the simulation with information on both the research topic and the backgrounds of the team members using the following prompt:

You are a scientist. Here is your bio: {profile-PLACEHOLDER}.

*Now you are collaborating with {num_scholars-PLACEHOLDER} other scholars with different academic backgrounds. After being given a topic, your goal is to discuss with other scholars and write an integrated paper of {num_words-PLACEHOLDER} words together. During the discussion, each of you will speak in turn. In your turn, you need to engage in the discussion and decide what to speak. Besides, you can also contribute to the paper by adding, editing, or deleting content based on the current version. Make sure to not only add content from your own field, but also build on others' points by integrating your expertise into editing ideas of what others have written. Be critical and do not only agree on others' points.*

*Organize your response strictly in the following format:*

*THOUGHT: ...(Your thoughts on the topic, others' opinions and the paper. Please think step by step to analyze the problem.)*

-----

*SPEAK: ... (What you want to say at this turn. You can ask questions, express your opinions, or respond to others' opinions.)*

-----

*WRITING: ... (What you want to add, edit, or delete. Make sure there is a substantive contribution. Make sure there are clear logic structure in each paragraph, and between paragraphs in paper. You may use subtitles to clarify the structure. Only write the incremental changes in the paper. Do not copy paste the existing content.)*

*Notes:*

*1. THOUGHT, SPEAK and WRITING is optional in each turn. But you are encouraged to provide your THOUGHT and SPEAK to contribute to the discussions.*

*2. You should make substantive contributions in both the discussion and paper writing according to your expertise. Do not only make incremental changes.*

*3. You should read the discussions and paper updates from other scholars carefully before making your own contributions and build on their points. Try to integrate your expertise into editing ideas of what others have written.*

*4. Please control the final length of the paper to be around {num_words-PLACEHOLDER}words. You can add, edit, or delete content to make the paper more coherent and complete.*

*5. In SPEAK, try to behave like a real human being and react to other people's ideas. You must speak. You can also summarize your own THOUGHT and speak to others. You may be excited by good ideas, bored by bad ones, pleased with polite discussion and offended by impolite ones.*

*6. No need to include references at the end of the paper.*

In our experiment, each simulation of team collaboration runs for 20 rounds. In every round, all agents engage in the discussion, and jointly drafting, revising, and polishing a short paper. A simplified example of the team conversation—excluding detailed notes on paper revisions—is provided below.

*Round 1 [Speaker: margaret_ayre_0]: Hey guys, today's topic is: computer interaction and the benefits of social networking for people with borderline personality disorder enlightening mental health professionals. Let's start the discussion and write the paper together!*
*Round 2 [Speaker: ricardo_corderosoto_0]: I'm excited to explore this topic, but I must admit, I'm not very familiar with mental health. Can someone provide a brief overview of BPD and how social networking might help? [Edit: I'll start by adding a section on the complexity and dynamics of quantum systems to provide a foundation for our discussion...]*
*Round 3 [Speaker: 3149225960]: I appreciate the attempt to connect quantum complexity to BPD, but I think we should prioritize a clear understanding of BPD and its relation to technology. Could we start by defining BPD and its symptoms? Then we can discuss how social networking and computer interaction might help manage these symptoms. [Edit: - Borderline Personality Disorder (BPD) and Its Symptoms Borderline Personality Disorder (BPD) is a mental health condition characterized by pervasive instability in moods, interpersonal relationships, and self-image.....]*
*Round 4 [Speaker: margaret_ayre_0]: Thanks for the great start! I think it's crucial to acknowledge the potential risks and challenges of using technology for managing BPD. Could we add a section on that? Also, I'd like to bring in my expertise in collaborative processes and indigenous knowledge integration. How can we use these technologies to empower individuals with BPD and their communities, rather than just providing a service? [Edit: Potential Risks and Challenges of Technology Use for Individuals with BPD; Empowering Individuals with BPD and Their Communities through Technology]*
*Round 5 [Speaker: ricardo_corderosoto_0]: I appreciate the comprehensive overview of BPD and the potential benefits and risks of technology use. I'd like to suggest that we delve deeper into the empowerment aspect and explore how we can use principles from my field to illustrate the complex dynamics of BPD and how technology can help manage them. What do you think?*
*......*
*Round 20 [Speaker: 3149225960]: I want to build on the previous scholar's points about the importance of involving mental health professionals in technology-based interventions for*

*individuals with BPD, especially when AI and machine learning are involved. I think it's crucial to ensure that these interventions are evidence-based, safe, and ethical. I also want to suggest that we consider the potential of digital storytelling as a means of promoting cultural competency and reducing stigma associated with BPD. Digital storytelling can provide a platform for individuals with BPD to share their stories and experiences, which can help humanize the condition and challenge stereotypes. However, we should also consider the potential risks and challenges, such as the need for mental health professional involvement to prevent re-traumatization and ensure the accuracy of the stories being shared. Additionally, we should ensure that the digital storytelling platforms are accessible and inclusive, so that individuals from diverse backgrounds and communities can participate and benefit from them. What are your thoughts on this? [Edit: I would like to add a new subtitle "Digital Storytelling for Cultural Competency and Stigma Reduction" and include the following points under it.....]*

In each round, we also get an intermediate version of the essay, which evolves through the collaborative process.

The methodological framework of our LLM experiment is illustrated in Extended Data Figure 6. According to our design, each team produces a short 1,000/2,000 word paper following the simulated discussion, representing the outcome of their collaborative effort. An example of such a paper is provided below. While the LLM generates multiple layers of subtitles, we numbered them chronologically to improve the clarity and organization of the paper's presentation.

*Title: Computer Interaction and The Benefits of Social Networking for People with Borderline Personality Disorder Enlightening Mental Health Professionals*

*1. Borderline Personality Disorder (BPD) and Its Symptoms*

*Borderline Personality Disorder (BPD) is a mental health condition characterized by pervasive instability in moods, interpersonal relationships, and self-image. Individuals with BPD often experience intense emotions, distorted self-images, and impulsive, self-destructive behaviors. They may also have difficulty regulating their emotions and maintaining stable relationships. BPD can be usefully understood through the lens of dynamical systems theory, which is a branch of mathematics that studies the behavior of systems that change over time. In this context, the symptoms of BPD can be seen as the result of unstable dynamics, where small perturbations can lead to large changes in mood, behavior, and interpersonal relationships.*

## *2. Potential Impacts of Technology Use for Individuals with BPD*

### *2.1 Empowerment through Understanding and Control*

*While more research is needed to fully understand the impact of technology on mental health, there are several potential benefits and risks of social networking and computer interaction for individuals with BPD. These include:*

*- Emotional Support and Validation: Online communities can provide a safe space for individuals with BPD to express their feelings and receive validation from others who understand their experiences.*

*- Stability and Coping Strategies: Online platforms can provide a stable environment for individuals with BPD to practice and develop coping strategies, much like a stable physical environment can help stabilize a physical system.*

*- Social Buffering: Online communities can provide a form of social buffering, where the presence of others can help stabilize emotional dynamics, similar to how a stable social environment can help stabilize a social system.*

*- Skill Building and Coping Strategies: Online resources and platforms can offer evidence-based strategies and tools for managing BPD symptoms, such as mindfulness exercises and cognitive-behavioral techniques.*

*- Reduced Social Anxiety: For individuals with BPD who experience social anxiety, online interactions can provide a less intimidating environment to practice social skills and build relationships.*

*- Access to Professional Help: Online therapy and support groups can provide access to mental health professionals and peer support, which may be particularly beneficial for individuals in remote or underserved areas.*

*- Isolation: While online communities can provide support, they can also contribute to social isolation if they replace offline interactions.*

*- Misinformation: The internet can be a source of misinformation about BPD, which can be harmful if individuals with BPD rely on it for guidance.*

*- Cyberbullying and Harassment: Online environments can expose individuals with BPD to cyberbullying and harassment, which can exacerbate their symptoms.*

*- Privacy Concerns: Sharing personal information online can raise privacy concerns and potentially lead to stigma or discrimination.*

*2.2 Promoting Cultural Competency and Reducing Stigma through Technology*

*Technology can also play a role in promoting cultural competency and reducing stigma associated with BPD. This can be achieved by:*

*- Culturally Sensitive Resources: Developing online resources that are culturally sensitive and relevant to diverse communities.*

*- Storytelling and Sharing: Providing platforms for individuals with BPD to share their stories and experiences, which can help humanize the condition and challenge stereotypes.*

*- Education and Awareness: Using technology to educate the public about BPD, its causes, and its impacts, which can help reduce stigma and promote understanding.*

*- Cultural Brokerage: Using technology to facilitate cultural brokerage, where individuals from different cultural backgrounds can connect, learn from each other, and build bridges of understanding.*

*From a dynamical systems perspective, this integration can be seen as a way of enriching the system, providing additional resources and dynamics that can help stabilize the system and promote resilience.*

### *2.3 Digital Storytelling for Cultural Competency and Stigma Reduction*

*Digital storytelling can provide a platform for individuals with BPD to share their stories and experiences, which can help humanize the condition and challenge stereotypes. To promote cultural competency through digital storytelling, it is important to ensure that the stories being shared are culturally sensitive and relevant to diverse communities. Mental health professionals should be involved in the development and implementation of digital storytelling interventions to prevent re-traumatization and ensure the accuracy of the stories being shared. Digital storytelling platforms should be accessible and inclusive, so that individuals from diverse backgrounds and communities can participate and benefit from them. Digital storytelling can also be used to educate the public about BPD, its causes, and its impacts, helping to raise awareness and promote understanding.*

### *2.4 Promoting Self-Regulation and Self-Awareness through Technology*

*Technology can be a valuable tool in promoting self-regulation and self-awareness for individuals with BPD. By incorporating various tools and applications, individuals can track their moods and behaviors, which can lead to increased awareness and understanding of their emotional patterns and triggers. This self-monitoring can serve as a foundation for developing better self-regulation skills. Furthermore, technology can offer interactive exercises and real-time feedback that help users practice mindfulness and emotional regulation techniques. Through these digital means, individuals with BPD can work towards greater emotional stability and self-awareness.*

### *2.5 Leveraging Peer Support through Technology*

*Peer support can be a crucial component of managing BPD, as it provides individuals with a sense of connection, validation, and shared understanding. Technology can facilitate and enhance peer support in several ways:*

*- Online Communities: Online communities and forums can provide a safe and accessible space for individuals with BPD to connect with others who understand their experiences.*

*- Peer-to-Peer Support: Technology can enable peer-to-peer support, where individuals with BPD can connect directly with each other to share coping strategies, provide emotional support, and challenge distorted thoughts.*

*- Mentoring: Technology can facilitate mentoring relationships, where individuals with BPD can connect with mentors who have successfully managed their condition. These mentors can provide guidance, encouragement, and practical advice.*

*- Gamification: Technology can be used to gamify peer support, making it more engaging and motivating. For example, individuals with BPD could earn points or badges for completing certain tasks or achieving specific goals, with the support of their peers.*

*By leveraging peer support through technology, individuals with BPD can build a strong support network that can help them manage their condition more effectively.*

*3. The Role of Artificial Intelligence and Machine Learning in Personalizing Technology-Based Interventions*

*Artificial intelligence (AI) and machine learning have the potential to significantly enhance technology-based interventions for individuals with BPD by personalizing these interventions to better meet their unique needs. AI algorithms can analyze data from users to identify patterns and predict future behaviors, allowing for more tailored interventions that address specific symptoms and challenges experienced by individuals with BPD. Machine learning can also facilitate the development of adaptive interventions that evolve based on user responses, increasing their effectiveness over time.*

*AI and machine learning can improve the accessibility of interventions by automating aspects of mental health care, such as screening and self-assessment, which can help reach individuals in remote or underserved areas. Additionally, AI-driven applications can offer real-time feedback and support, assisting individuals in developing effective coping strategies and emotional regulation techniques.*

*4. Navigating Ethical Challenges: Ensuring Responsible Technology Use*

*While technology offers numerous potential benefits for managing BPD, it is crucial to consider the ethical implications of its use. Some key ethical challenges include:*

*- Informed Consent: Ensuring that individuals with BPD are fully informed about the potential benefits and risks of using technology for managing their condition, and that they provide their informed consent before participating in any technology-based interventions. This is particularly important given the emotional volatility and impulsivity that can characterize BPD.*

*- Privacy and Confidentiality: Protecting the privacy and confidentiality of individuals with BPD, particularly when they are sharing personal information online. This is essential to prevent re-traumatization and to build trust in technology-based interventions.*

*- Cultural Sensitivity: Ensuring that technology-based interventions are culturally sensitive and relevant to diverse communities. This involves acknowledging and respecting the unique cultural contexts and needs of individuals with BPD, and avoiding approaches that perpetuate stereotypes or stigmatize individuals with BPD.*

*- Power Dynamics: Being mindful of power dynamics in the development and implementation of technology-based interventions. This involves ensuring that individuals with BPD and their communities have a say in the process, and that their voices are heard and respected.*

*- Digital Divide: Being aware of the digital divide, and ensuring that technology-based interventions are accessible to individuals with BPD from diverse backgrounds and locations.*

*This involves considering factors such as internet access, digital literacy, and the availability of appropriate devices.*

*- Mental Health Professional Involvement: Ensuring that mental health professionals are involved in the development and implementation of technology-based interventions. This can help ensure that these interventions are evidence-based, safe, and effective.*

### *4.1 The Potential for Technology to Exacerbate Inequalities*

*While technology offers numerous potential benefits for managing BPD, it is also important to consider the potential for it to exacerbate existing inequalities and create new ones. For instance, the use of technology for managing BPD may be more accessible to individuals with higher income and education levels, who may have better access to the internet and appropriate devices. Similarly, the use of technology for managing BPD may be more appealing to individuals who are already comfortable with technology, potentially creating a digital divide between those who can access and use technology-based interventions and those who cannot. To mitigate these risks, it is essential to ensure that technology-based interventions are designed with equity in mind, and that they are accessible to individuals from diverse backgrounds and locations. This involves considering factors such as income, education, digital literacy, and cultural relevance, and ensuring that technology-based interventions are tailored to the unique needs and preferences of diverse communities.*

*By considering these ethical challenges, we can ensure that technology is used in a way that is beneficial, respectful, and empowering for individuals with BPD and their communities. This involves a commitment to ongoing dialogue, reflection, and adaptation, as well as a willingness to learn from and collaborate with individuals with BPD and their communities.*

## *5. The Role of Mental Health Professionals in Technology-Based Interventions*

*The involvement of mental health professionals is crucial in the development and implementation of technology-based interventions for individuals with BPD, particularly those involving artificial intelligence (AI) and machine learning. Their expertise is vital to ensuring the*

*safety, effectiveness, and ethical use of these interventions. Mental health professionals can provide guidance on evidence-based practices and ensure that interventions are tailored to the unique needs of individuals with BPD. Their participation helps to establish trust and rapport with individuals, making it more likely that AI and machine learning-based interventions will be accepted and used effectively.*

*6. Empowerment through Collaborative Processes and Indigenous Knowledge Integration*

*To fully empower individuals with BPD and their communities, it is essential to involve them in the development and implementation of technology-based interventions. This can be achieved through collaborative processes that engage individuals with BPD, their families, and community members in decision-making and problem-solving. By working together, these stakeholders can ensure that technology-based interventions are tailored to their needs and preferences, and that they are culturally appropriate and relevant. Indigenous knowledge integration is also crucial in this process. Incorporating indigenous knowledge and cultural practices into technology-based interventions can help ensure that they are relevant and meaningful to diverse communities. This integration can also provide additional resources and dynamics that can help stabilize the system and promote resilience.*

*7. Bridging the Digital Divide: Ensuring Accessibility and Digital Literacy*

*Technology use for managing BPD has the potential to exacerbate existing inequalities and create new ones. One key concern is the digital divide, which refers to the gap between individuals, households, businesses, and geographic areas in their access to information and communication technologies (ICTs) and their use of the internet. This divide can be due to various factors, including income, education, age, gender, race, and location. For individuals with BPD, the digital divide can present significant barriers to accessing technology-based interventions. For instance, individuals with BPD who live in remote or underserved areas may not have access to the internet or appropriate devices. Similarly, individuals with BPD who have low digital literacy may struggle to use technology-based interventions effectively.*

*To promote digital equity, it is crucial to invest in infrastructure, education, and support services that target underserved communities. This can involve partnerships with local organizations, governments, and non-profits to provide access to technology, digital literacy training, and mental health resources. By addressing the digital divide, we can ensure that technology-based interventions for managing BPD are accessible and beneficial to all individuals, regardless of their background or location.*

*8. The Importance of Human Connection and Active Involvement*

*While technology offers numerous potential benefits for managing BPD, it is also important to consider the potential for it to disempower individuals if not used carefully. For instance, the use of technology for managing BPD could potentially replace human interaction with digital tools, which could lead to a loss of social support and a sense of isolation. Additionally, the use of technology for managing BPD could potentially lead to a reliance on digital tools, which could limit individuals' ability to develop and use their own coping strategies and skills.*

*To mitigate these risks, it is essential to ensure that technology-based interventions are used in a way that complements and enhances human interaction, rather than replacing it. It is also important to ensure that individuals with BPD are actively involved in the development and implementation of technology-based interventions, and that they have a say in how these interventions are used. This can help ensure that technology-based interventions are used in a way that is empowering and beneficial for individuals with BPD and their communities.*

### S4.2.4 Prompts to Rate Quality of the Simulated Paper

We use GPT-4o (accessed in February 2025) to evaluate the quality of the papers generated through the LLM simulation. This model has been validated for rating and evaluation tasks [45]. We further discuss the validity and robustness of AI-based evaluation in the next section.

Given that our simulation is situated within a scientific context, we adopt two criteria for assessment. First, we evaluate the strength of the scientific argument, using the reviewer guide of NeurIPS 2024 for this purpose. Specifically, we ask the model to evaluate:

1. *Originality: Are the tasks or methods new? Is the work a novel combination of well-known techniques? Is it clear how this work differs from previous contributions? Is related work adequately cited?*
2. *Quality: Is the submission technically sound? Are claims well supported (e.g., by theoretical analysis or experimental results)? Are the methods used appropriate? Is this a complete piece of work or work in progress? Are the authors careful and honest about evaluating both the strengths and weaknesses of their work?*
3. *Clarity: Is the submission clearly written? Is it well organized? Does it adequately inform the reader?*
4. *Significance: Are the results important? Are others (researchers or practitioners) likely to use the ideas or build on them? Does the submission address a difficult task in a better way than previous work? Does it advance the state of the art in a demonstrable way? Does it provide unique data, unique conclusions about existing data, or a unique theoretical or experimental approach?*

Second, we assess the overall quality of writing, for which we use the Association of American Colleges and Universities (AAC&U) Valid Assessment of Learning in Undergraduate Education (VALUE) Rubrics for writing communication. We specifically focus on four aspects of essay:

1. *Context of and Purpose for Writing: Whether the paper demonstrates a thorough understanding of context, audience, and purpose that is responsive to the assigned task(s) and focuses all elements of the work.*

2. *Content Development: Whether the paper uses appropriate, relevant, and compelling content to illustrate mastery of the subject, conveying the writer's understanding, and shaping the whole work.*
3. *Genre and Disciplinary Conventions: Whether the paper demonstrates detailed attention to and successful execution of a wide range of conventions particular to a specific discipline and/or writing task (s) including organization, content, presentation, formatting, and stylistic choices*
4. *Control of Syntax and Mechanics: Whether the paper uses graceful language that skillfully communicates meaning to readers with clarity and fluency, and is virtually error free.*

These criteria are explicitly incorporated into our prompt. Specifically, we instruct the grading LLM using the following prompt, with a temperature setting of 0 to ensure consistency while allowing for some variation in responses. We ask the rating model to give a rating score from 1 to 10 on each criteria. The Prompt is written as below:

*You are a scholar with expertise in the topic of {topic-PLACEHOLDER}. Now you are reviewing four article submissions based on the four criteria as below. You need to read the articles, compare the four articles based on the four criteria, and provide evaluation scores for each article.*

*Here are the articles:*

*{article1-PLACEHOLDER}*

*{article2-PLACEHOLDER}*

*{article3-PLACEHOLDER}*

*{article4-PLACEHOLDER}*

*Here are the four criteria for evaluating the articles:*

*{criteria-PLACEHOLDER}*

*For each article, please give a rating score from 1 to 10 on each criteria. Then, give an overall rating score based on the overall quality of the article.*

*Importantly, please be critical when giving the score to each article and differentiate the quality of the four articles, because the four articles can be very different in quality. Your scores should reflect the different quality of the articles.*

*-Do not default to a middle/neutral score (such as 5 or 6). Only assign a middle score if the paper is truly mediocre in every aspect.*

*-If the paper shows notable strengths, even if imperfect, lean toward a higher score (7-10) based on its impact or originality.*

*-If the paper shows clear weaknesses, even if it has some merits, lean toward a lower score (1-4) based on its overall deficiencies.*

*-Be decisive and bold: Emphasize clear strengths or weaknesses rather than averaging them out.*

*Your response format should strictly follow the format below:*

*Article 1: [score1, score2, score3, score4, overall_score]*

*Article 2: [score1, score2, score3, score4, overall_score]*

*Article 3: [score1, score2, score3, score4, overall_score]*

*Article 4: [score1, score2, score3, score4, overall_score]*

An example response from the grading LLM is provided below:

*Article 1: [7, 8, 8, 7, 8]*

*Article 2: [6, 7, 7, 6, 7]*

*Article 3: [8, 8, 9, 9, 9]*

*Article 4: [7, 8, 8, 8, 8]*

The response format is standardized to facilitate parsing and score collection.

### S4.2.5 AI-Based Evaluation of Simulated Paper Quality: Literature Support and Validation

Large Language Models (LLMs) show considerable potential in evaluating academic papers, providing a reliable and effective proxy for human peer review. Their ability to identify higher-quality work is well supported in the literature[52–55] . Research has shown that LLMs can distinguish between higher- and lower-impact work[53], offer constructive scientific feedback to authors, and generate comments not raised by human reviewers[55]. Given the large number of simulated papers from various disciplines and the risk of bias when involving human evaluators with different mindsets and baseline scores, we use LLMs to assess simulated papers as a proxy for team collaboration effectiveness in our AI experiment.

To complete this task, we follow the optimal AI evaluation settings identified in prior work. First, we use a comparison-based design for our evaluation. Instead of assessing each paper individually, we compare and rank papers across conditions. Previous studies show that comparison-based evaluation aligns more closely with human judgments and yields greater consistency[52,56–58]. This approach reflects the observation that humans—and likely LLMs—find it easier to compare two options directly than to assign absolute scores independently [59]. Second, following common practice in this literature, we use an updated version of GPT for evaluation tasks. While, in theory, any LLM could serve this role, GPT has been the most frequently applied in the scientific evaluation literature, and its effectiveness has been repeatedly demonstrated[53–55,60]. Accordingly, we use GPT-4o in the main text and GPT-4o-mini in this appendix for additional robustness tests.

We first validate LLM evaluators using a dataset with ground truth, conducting a pairwise comparison task with ICLR 2025 submissions and their peer reviews scores. Papers are categorized into three quality tiers based on the ICLR rating rubric and average human review scores: Low (average score < 5), Middle (scores in the [5.0, 6.0) range), and High (score ≥ 6). Ratings of 5 ("marginally below the acceptance threshold") and 6 ("marginally above the acceptance threshold") represent the critical borderline for acceptance decisions. To assess the ability of LLMs to distinguish papers with higher human review ratings, we formed pairs by randomly selecting two papers from different tiers. GPT-4o and GPT-4o-mini were then prompted to evaluate the pairs, assign ratings, and select the higher-quality paper. For each comparison group (Low vs. Middle, Middle vs. High, and Low vs. High), we randomly sampled 200 paper pairs. To focus on papers with clear and consensual reviews, we only included submissions with high reviewer agreement (score standard deviation < 1.2) and confidence (average confidence ≥ 4.3).

The results, shown in Table S14, indicate that GPT-4o's performance closely mirrors human evaluation and captures the decisional difficulty of the peer-review process. First, our experiment replicates prior evidence on LLM effectiveness in identifying papers for acceptance and aligning with human preferences (Low vs. High), achieving high accuracy: 69.8% for GPT-4o and 58.3% for GPT-4o-mini. Second, for the Low vs. Middle and Middle vs. High comparison groups, where human judgments are more ambiguous, performance was lower but still exceeded random chance. GPT-4o correctly identified the better paper with 60.7% and 58.1% accuracy for these groups, with lower accuracy for GPT-4o-mini. These findings underscore the model's sensitivity to the inherent ambiguity in academic evaluation, reflecting the well-known challenge of distinguishing a “marginally below” paper from a “marginally above” one.

Second, we replicate the evaluation from Figure 4 in the main text using GPT-4o-mini (temperature = 0). As shown in Figure S11, the results are generally consistent with our main findings, demonstrating the replicability of LLM-based evaluations.

### S4.2.6 Statistical Results of the LLM Experiment

For each essay, we collect four evaluations and use the average score as its quality metric. We then examine the relationship between team composition and the quality of the collaborative work to assess whether the patterns identified in our statistical analysis reflect a causal relationship. We reported the average team performance under each condition in Figure 4 in the main text. In addition to Gemma, which we discussed in the main text, the “optimal” teams in Mistral-NeMo space have a 95% bootstrap CI of scientific scores of [7.048, 7,533], and a 95% bootstrap CI of writing scores of [7.543,7.990]; and “optimal teams” in QwQ space have a 95% bootstrap CI of scientific scores of [7.124, 7.619] and a 95% bootstrap CI of writing scores of [7.419, 7.857].

For knowledge integration, teams with high perspective and low background diversity have the highest integration score, with “optimal teams” having a 95% bootstrap CI of [0.127, 0.193] in Mistral space, and [0.230, 0.261] in QwQ space.

For knowledge speculation, teams with low perspective and high background diversity demonstrate the highest speculation score, with “worst teams” having a 95% bootstrap CI of [0.664, 0.722] in Mistral space, and [0.700, 0.719] in QwQ space,

For distance between assigned topic and final generated draft, teams with high perspective and low background diversity post the smallest distance, with a 95% bootstrap CI of [0.395, 0.467] in Mistral space, and [0.110, 0.141] in QwQ space. By contrast, teams with low perspective and background diversity show the highest distance to their topic, and the greatest drift, with 95% CI of [0.629, 0.762] for Mistral, and [0.192, 0.222] for QwQ.

We report the 95% bootstrap confidence intervals for all five scores—scientific rubric score, writing score, knowledge integration, knowledge speculation, and distance to topic—across three LLM spaces (Gemma, Mistral, QwQ) and four conditions, in Table S15.

**S4.2.7 Measuring Knowledge Integration, Speculation, and Topic Distance in the Intermediate Version of Simulated Papers**

For each round of interaction, AutoGen generates an intermediate version of the essay, which can be analyzed as a list of tokens. A successful essay should effectively incorporate the prior experiences of team members. To evaluate this, we construct measures of knowledge integration and speculation. Our operationalization follows the approach in Section S 2.7, with the only difference that tokens are used in place of knowledge modules, as no predefined modules exist for the simulated data.

Additionally, we calculate DistanceToTopic as the cosine distance between the average token embedding of the generated essay and the embedding of the assigned topic within each LLM's embedding space.

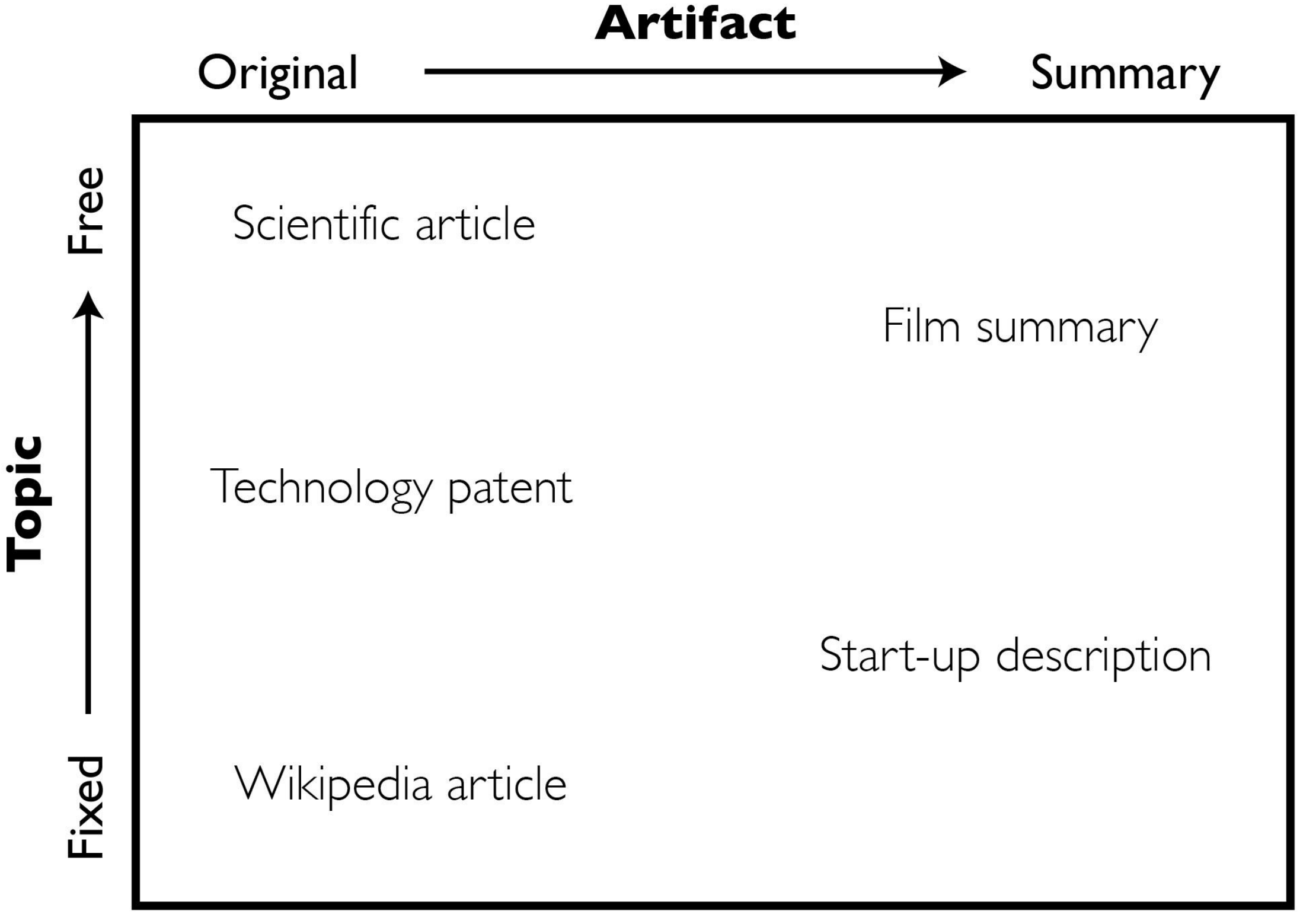

Figure S1. Space of Innovations.

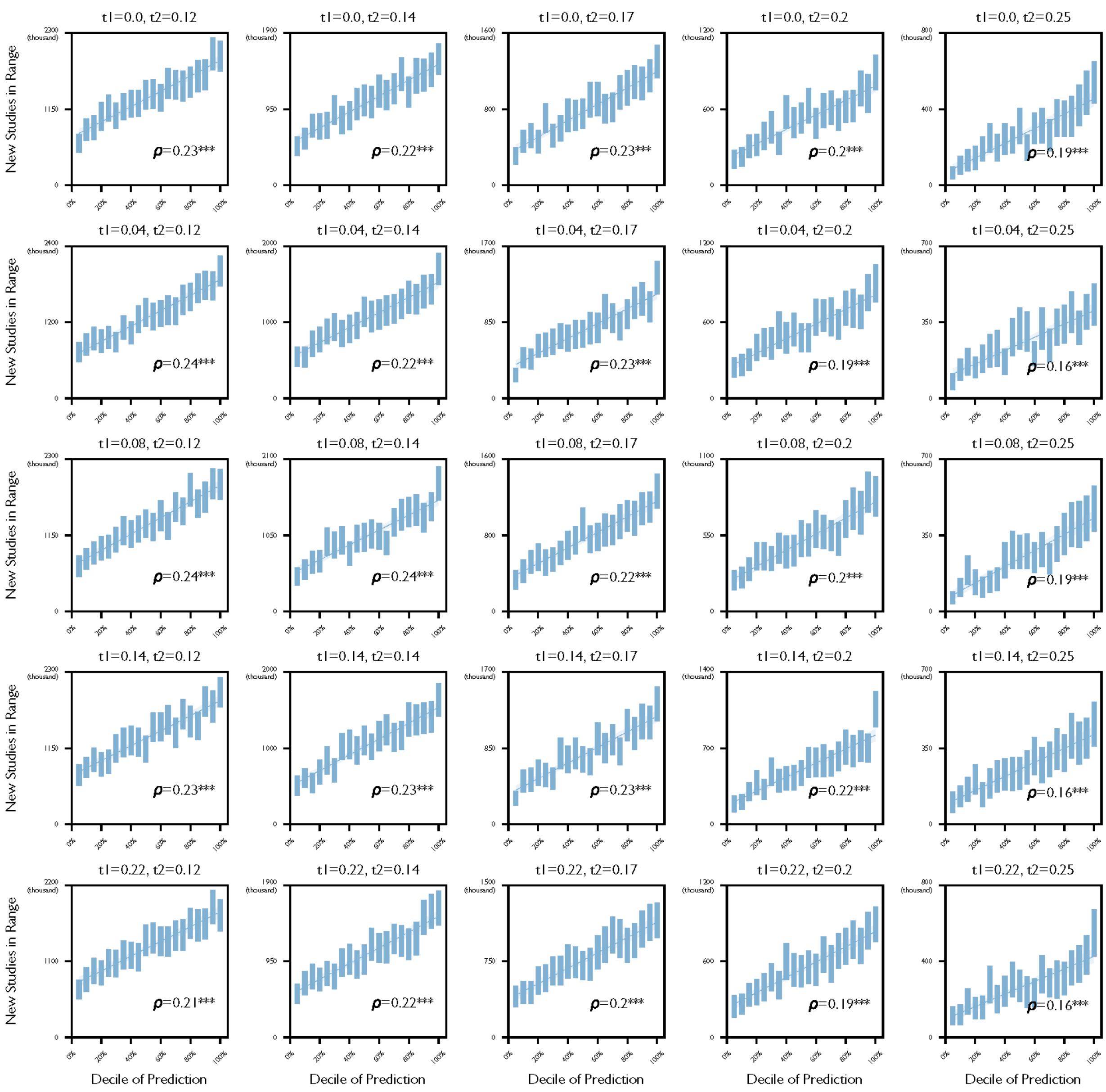

Figure S2. Science. We examine the correlation between conceptual flow in the surrounding area, defined by a distance threshold $t_1$, and innovations that emerge near the focal point in the subsequent period, using a second threshold $t_2$. Robustness tests confirm that our findings generally hold across different values of $t_1$ and $t_2$.

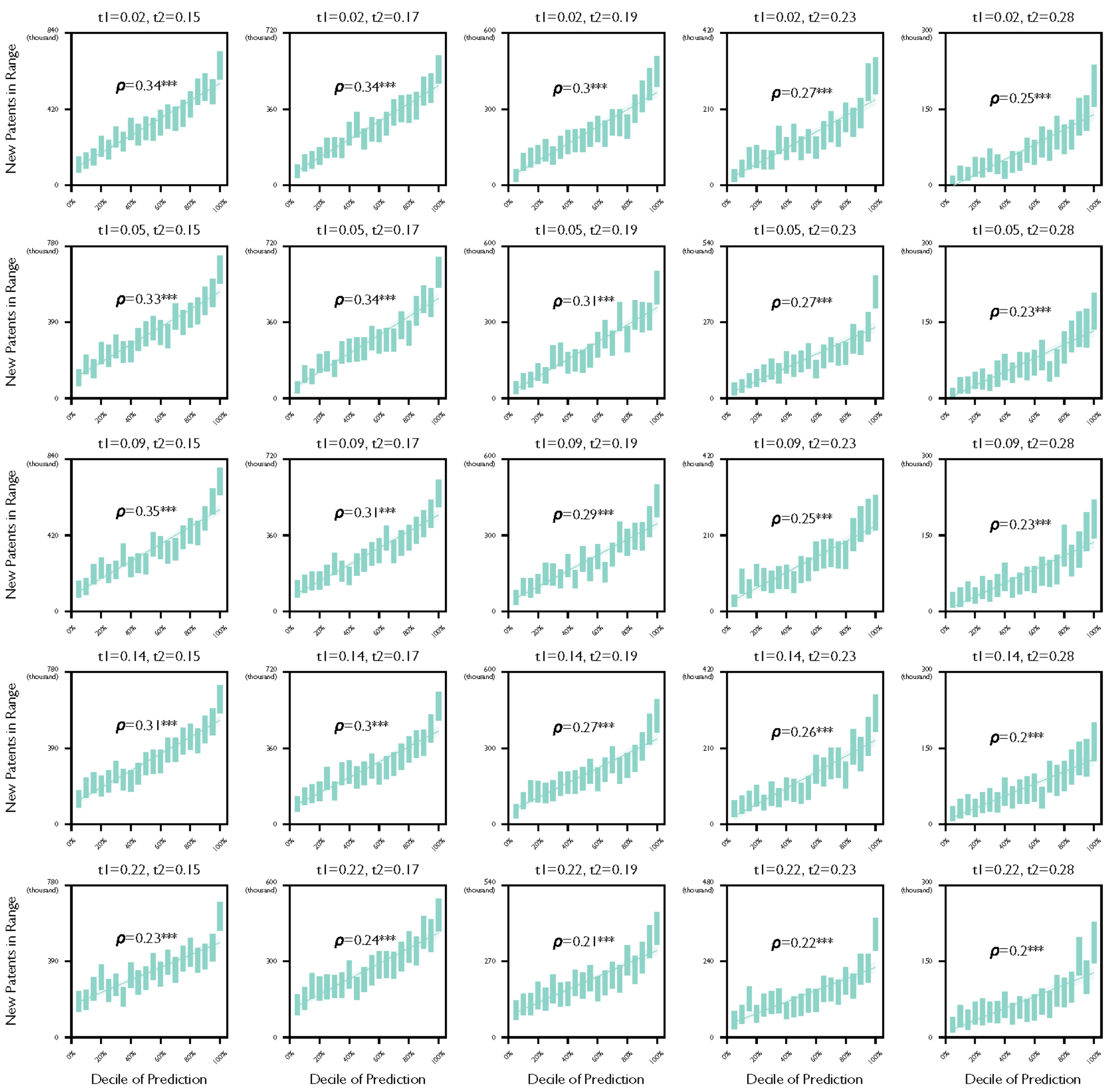

Figure S3. Technology. We examine the correlation between conceptual flow in the surrounding area, defined by a distance threshold $t_1$, and innovations that emerge near the focal point in the subsequent period, using a second threshold $t_2$. Robustness tests confirm that our findings generally hold across different values of $t_1$ and $t_2$.

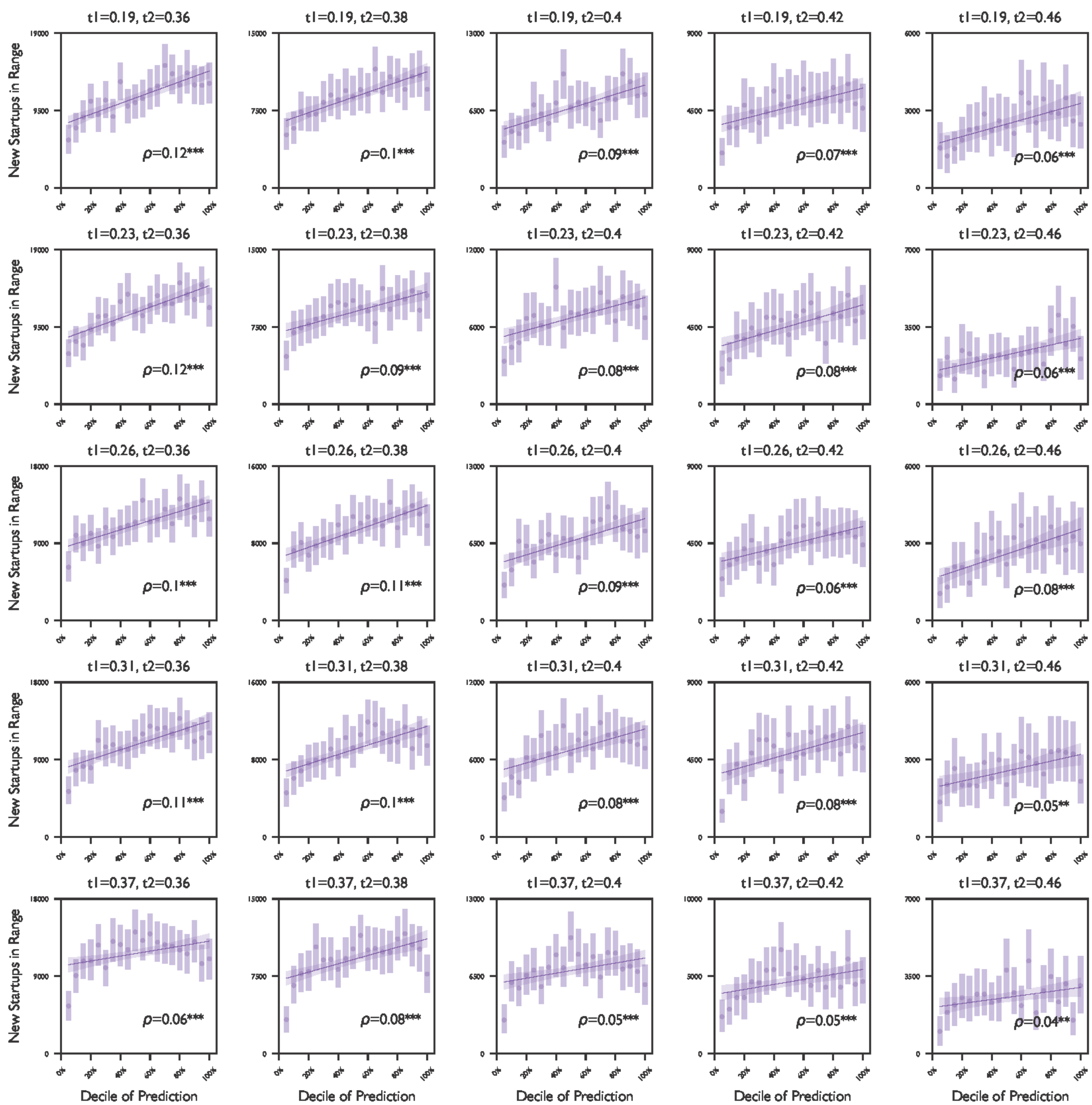

Figure S4. Entrepreneurship. We examine the correlation between conceptual flow in the surrounding area, defined by a distance threshold $t_1$, and innovations that emerge near the focal point in the subsequent period, using a second threshold $t_2$. Robustness tests confirm that our findings generally hold across different values of $t_1$ and $t_2$.

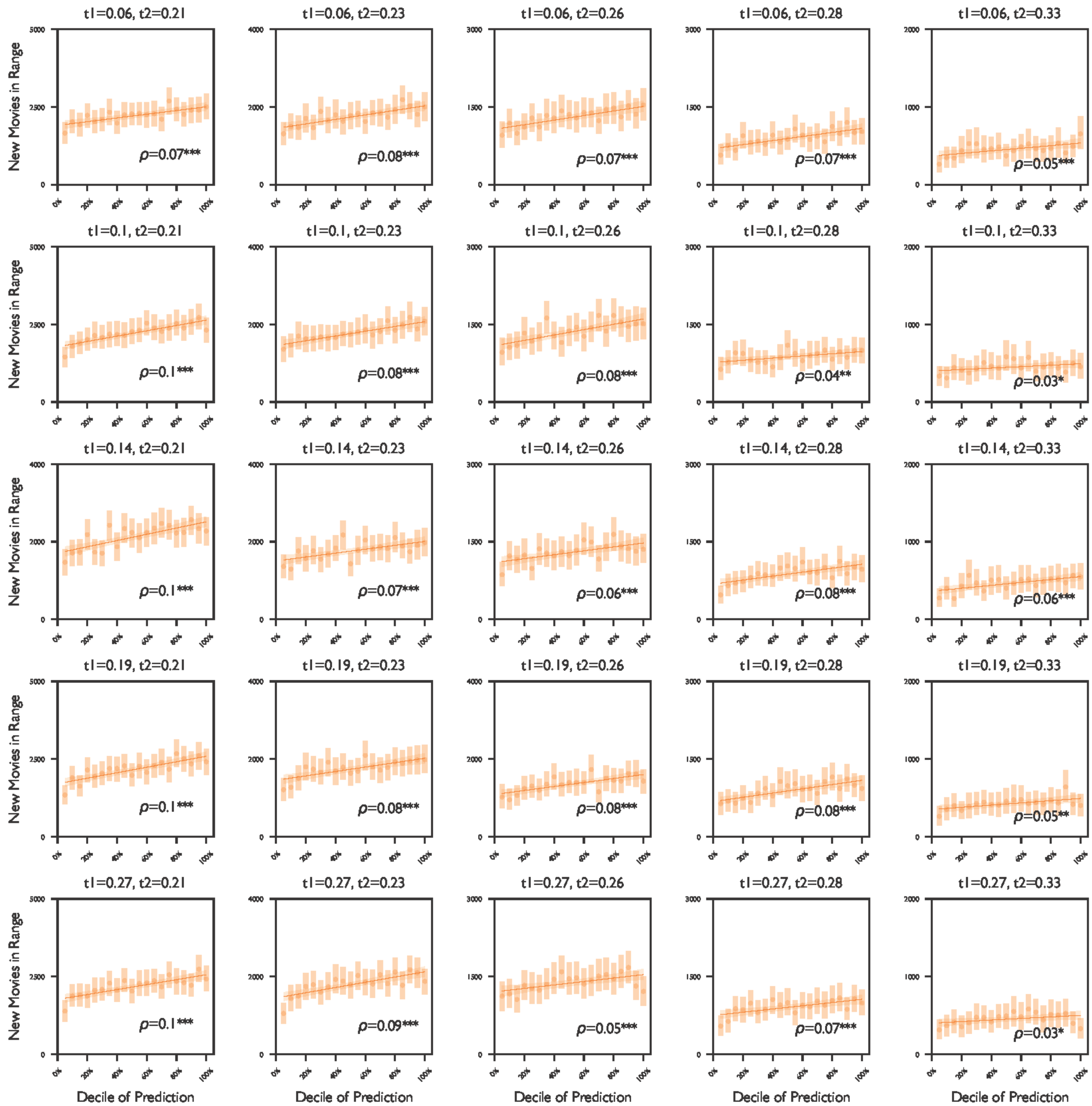

Figure S5. Film. We examine the correlation between conceptual flow in the surrounding area, defined by a distance threshold $t_1$, and innovations that emerge near the focal point in the subsequent period, using a second threshold $t_2$. Robustness tests confirm that our findings generally hold across different values of $t_1$ and $t_2$.

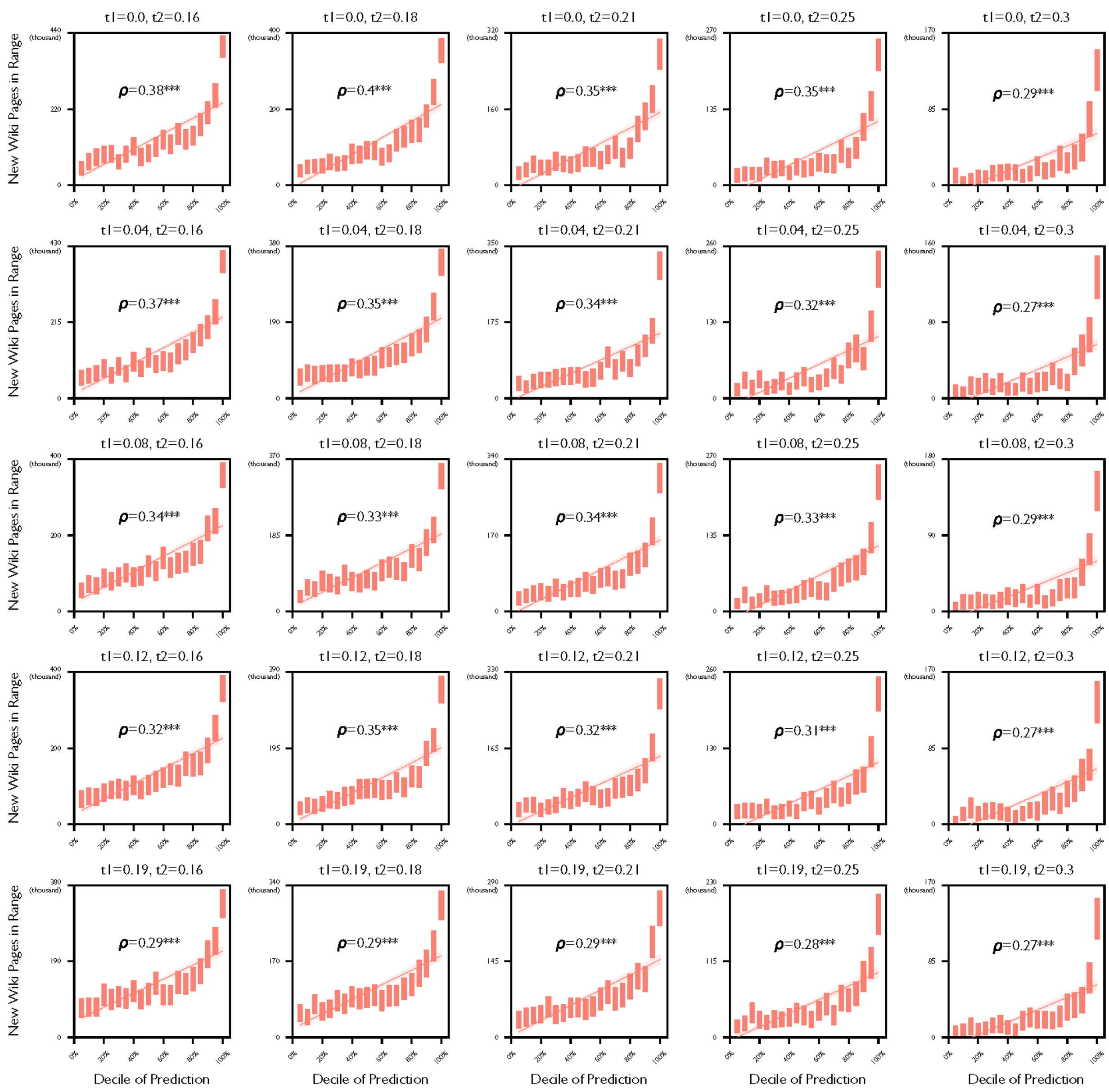

Figure S6. Wikipedia. We examine the correlation between conceptual flow in the surrounding area, defined by a distance threshold $t_1$, and innovations that emerge near the focal point in the subsequent period, using a second threshold $t_2$. Robustness tests confirm that our findings generally hold across different values of $t_1$ and $t_2$.

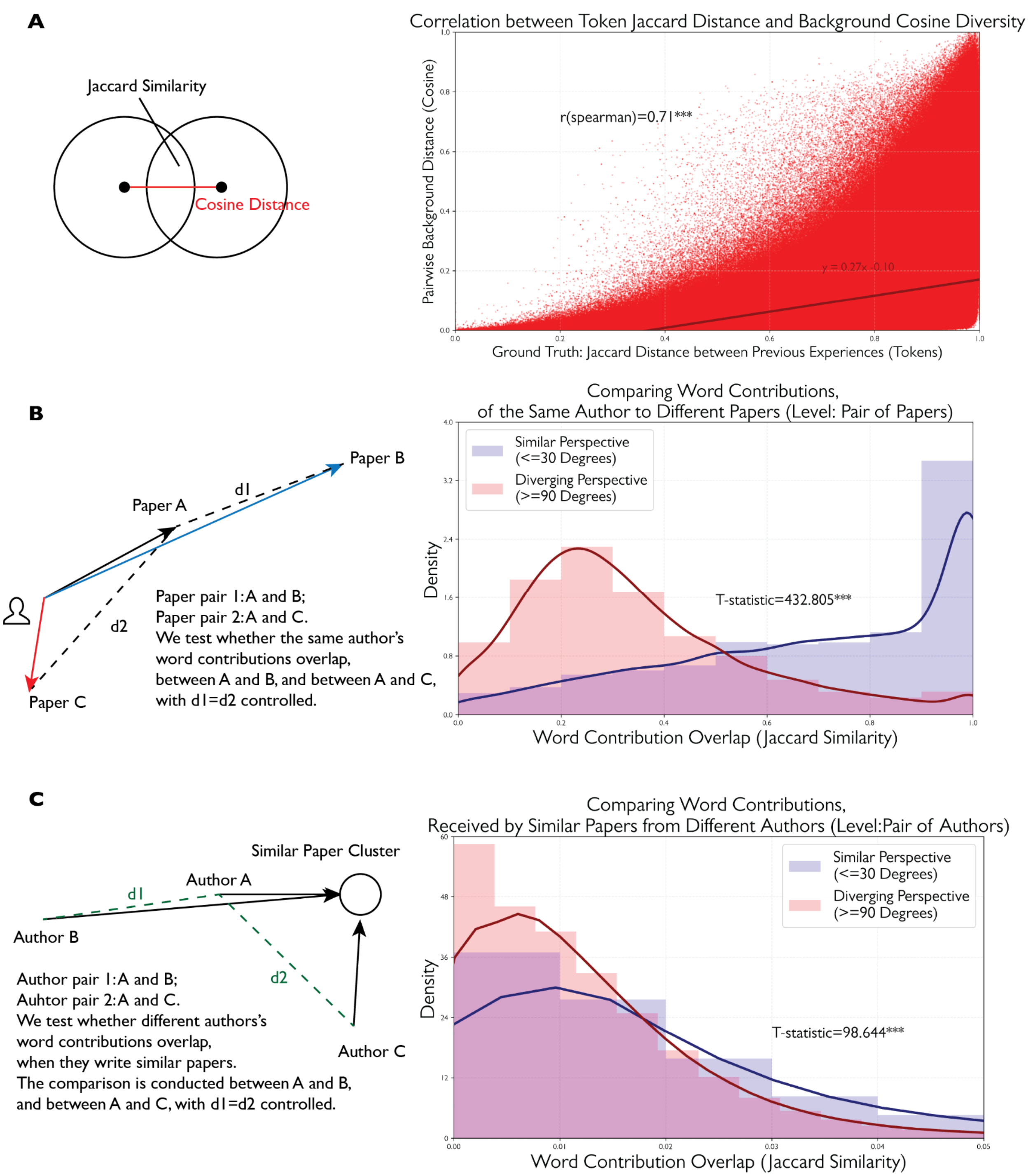

Figure S7. Validating Background Diversity and Perspective Diversity Measures.

Experience in Category 1-8

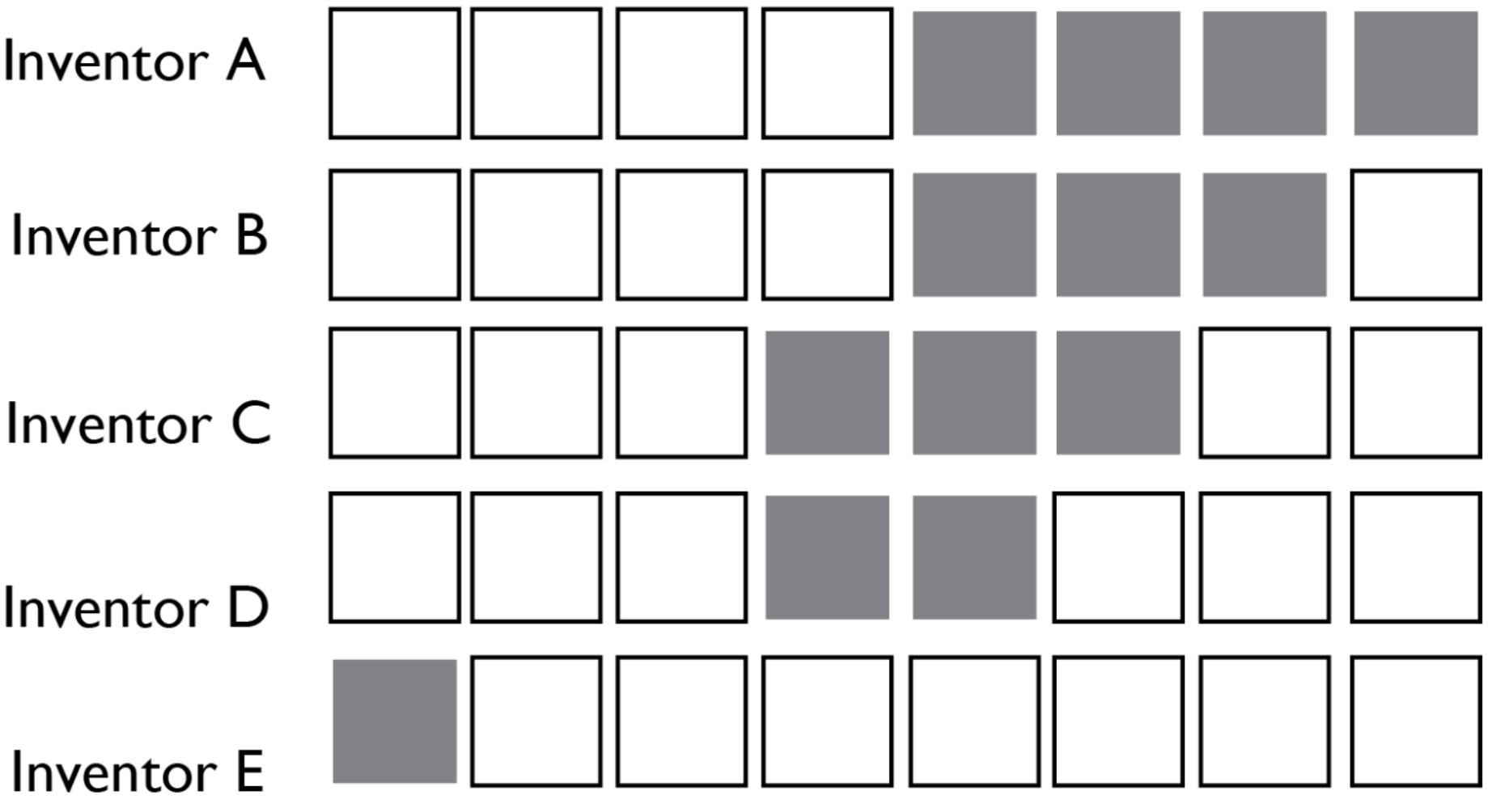

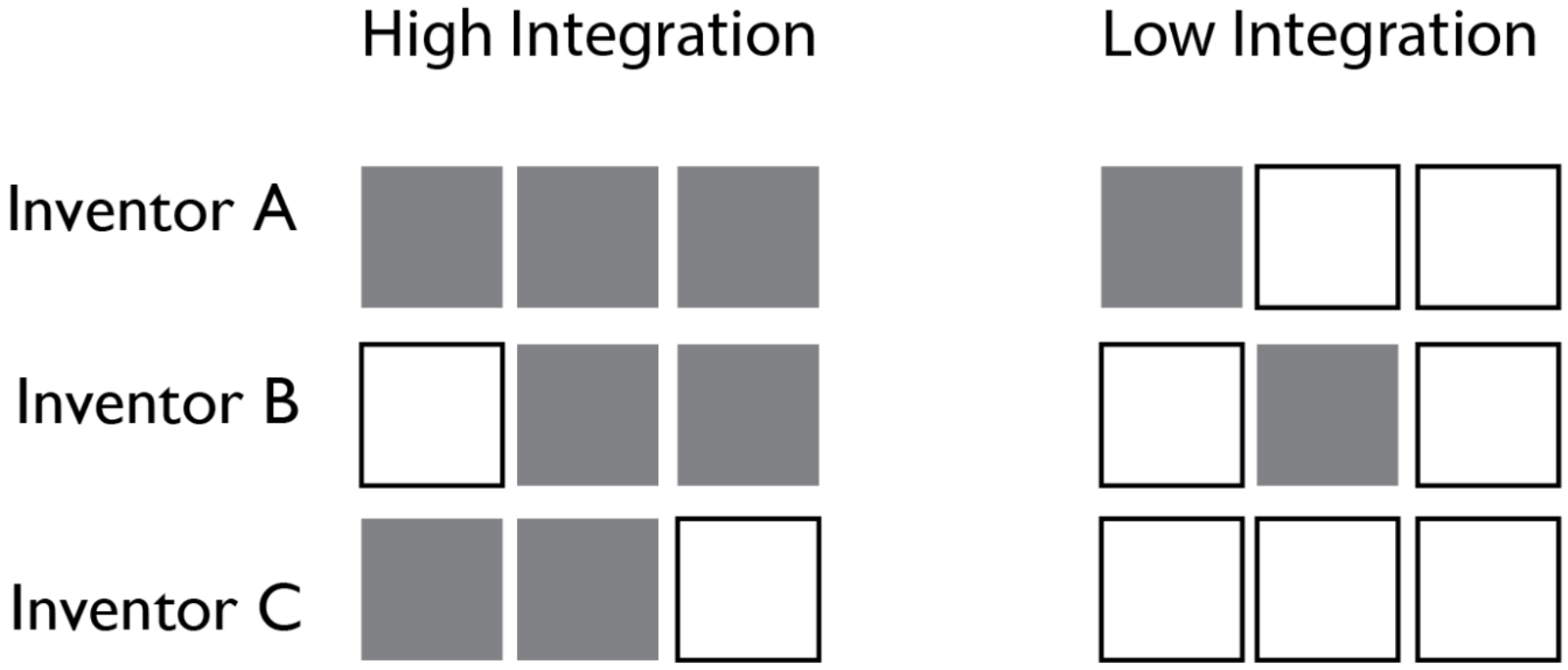

Average Member Experience in Related Categories (KNOWLEDGE_INTEGRATION)

$(2/3 + 3/3 + 2/3)/3 = 7/9$ $(1/3 + 1/3 + 0/3)/3 = 2/9$

Average Proportion of New Categories (KNOWLEDGE_SPECULATION)

$(0 + 0 + 0)/3 = 0/3$ $(0 + 0 + 1)/3 = 1/3$

Figure S8. Conceptual Figure for Measure Construction: Knowledge Integration and Speculation

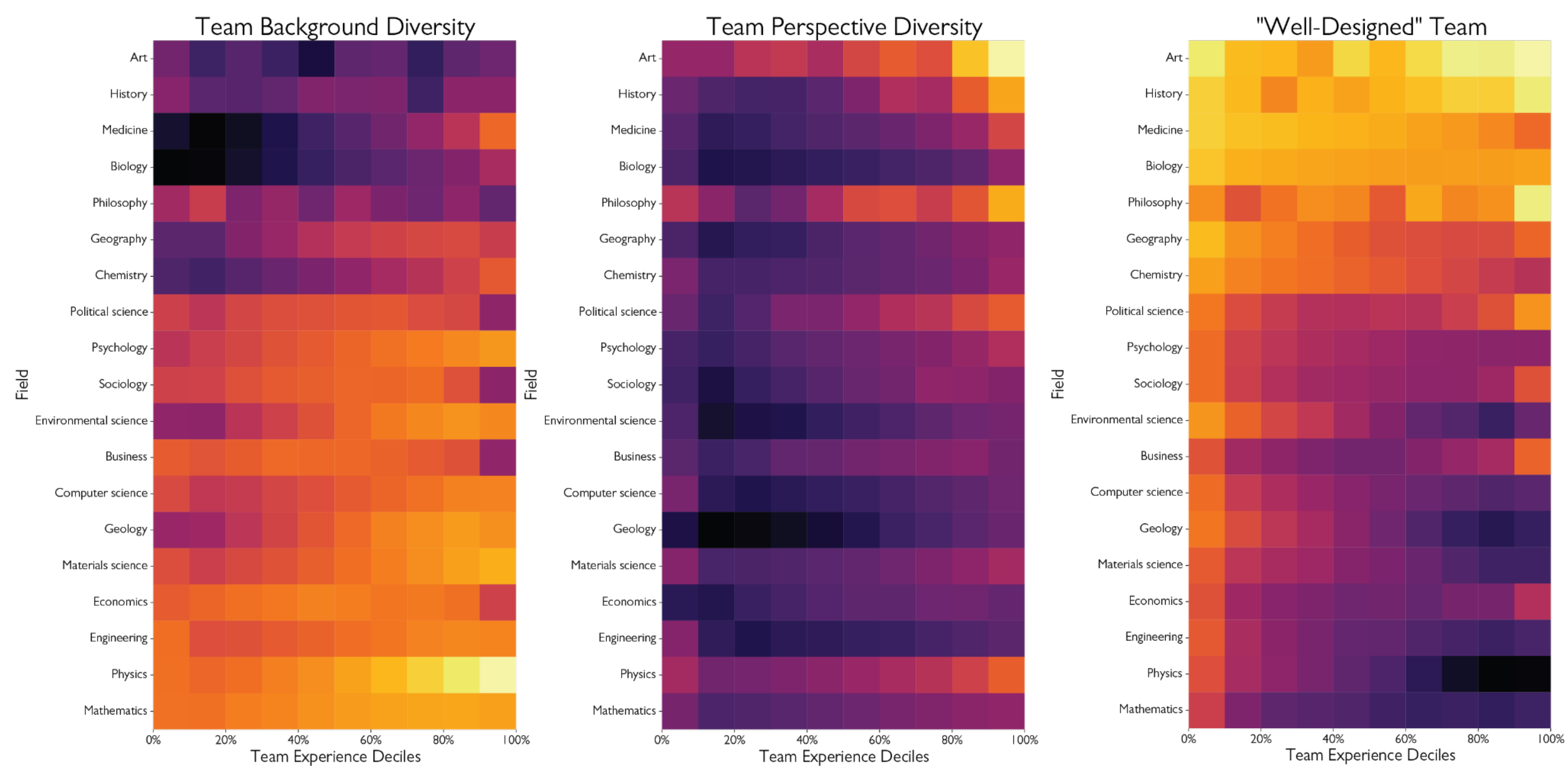

Figure S9. More Experienced Teams are More Likely to Have “Optimal” Compositions in Science.

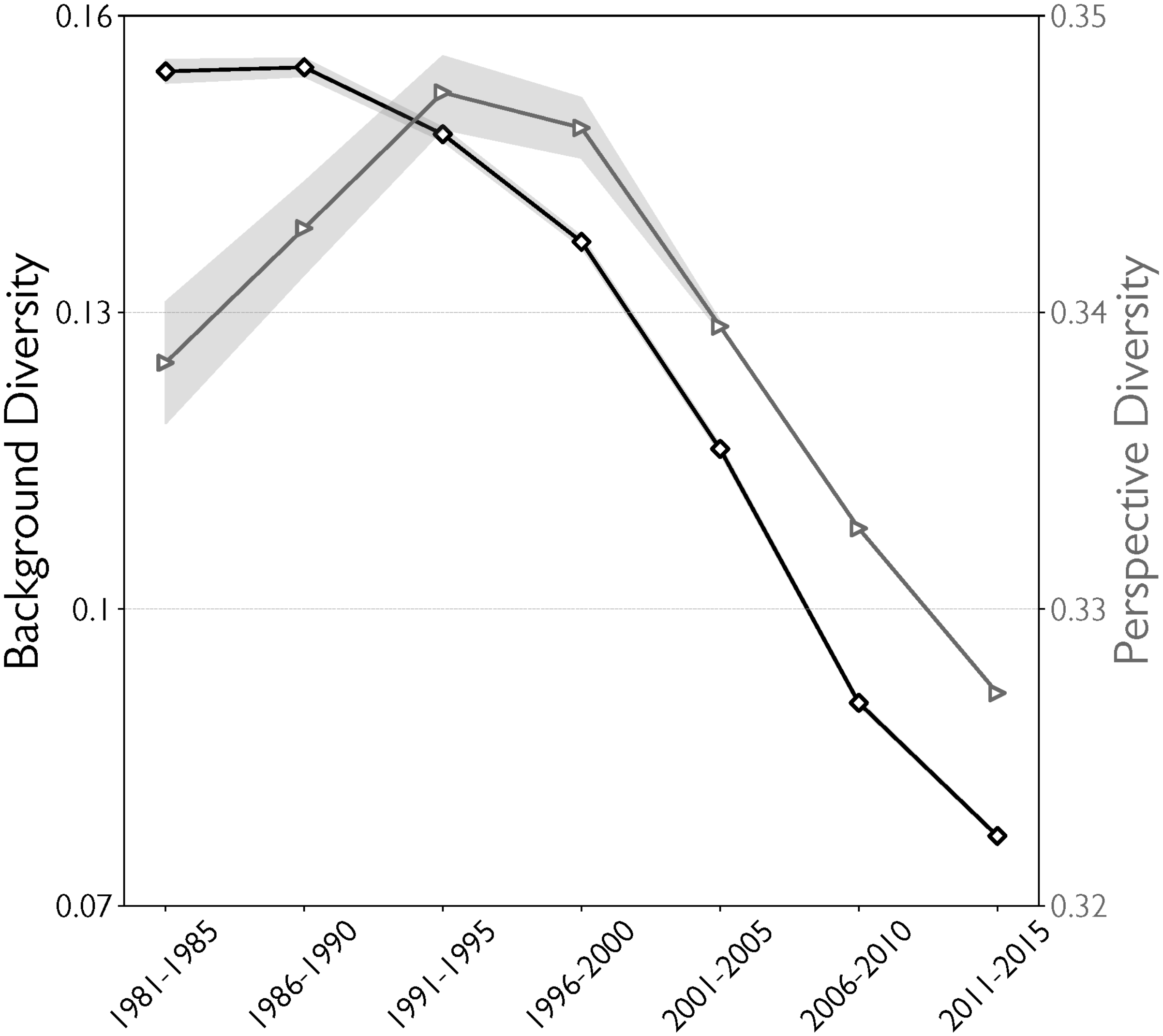

Figure S10. Temporal Change of Perspective and Background Diversities of Science Teams, 1981-2015

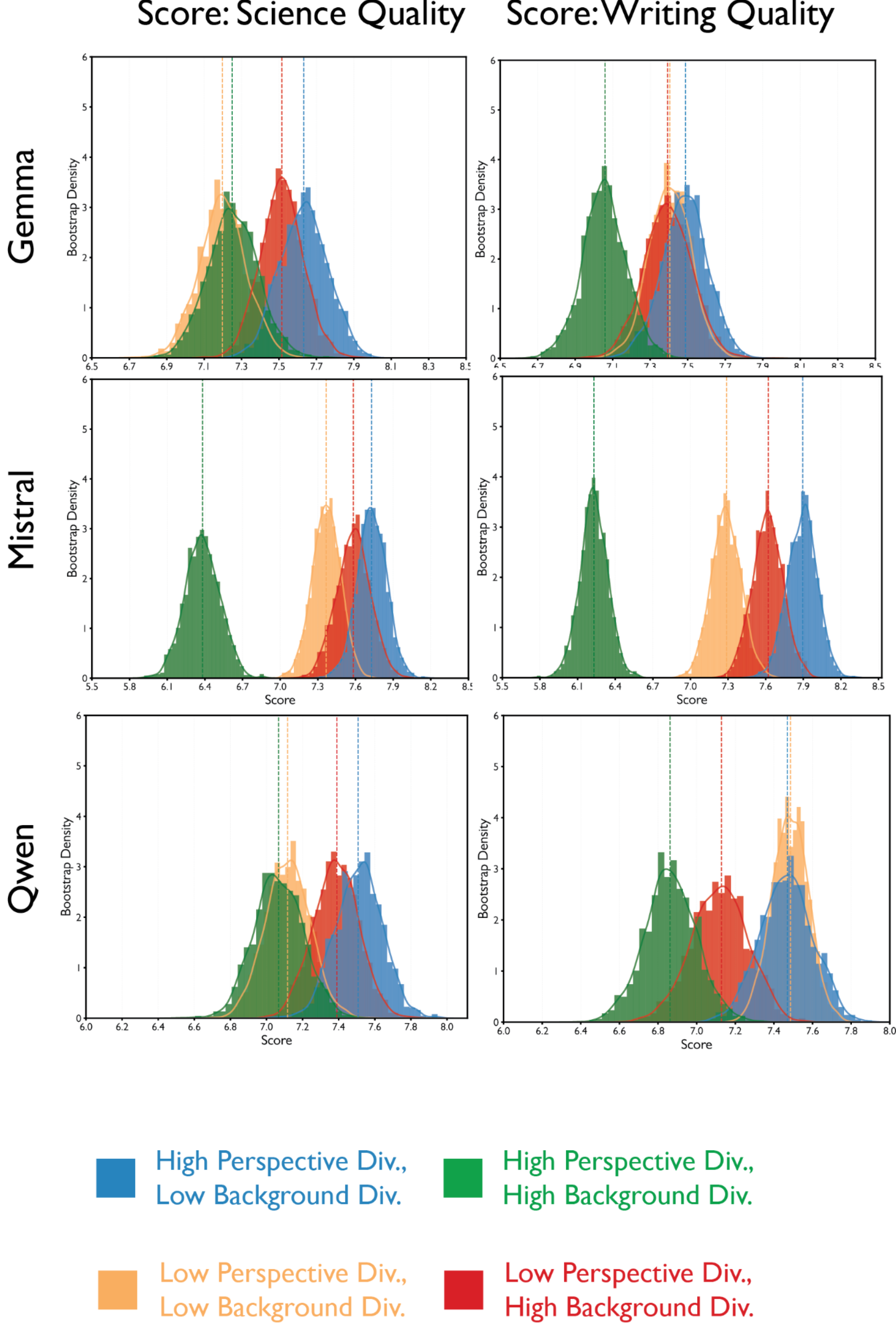

Figure S11. LLM Evaluation Results Using GPT-4o-mini

Table S1. Panel Granger Causality Test on Novelty Emergence and Space Dynamics for Science, Entrepreneurship, Film and Wikipedia Cases.

| | Dumitrescu & Hurlin (2012) Granger Non-Causality Test On Panel Dataset (based on 2000 random location in the dynamic space) | |
|---|---|---|
| | Space Dynamics on Novelty Emergence | Novelty Emergence on Space Dynamics |
| Science | W-bar = 2.872<br>Z-bar = 59.194 (p = 0.000)<br>Z-bar tilde = 11.434 (p = 0.000) | W-bar = 2.393<br>Z-bar = 44.033 (p = 0.000)<br>Z-bar tilde = 6.886 (p = 0.000) |
| Entrepreneurship | W-bar = 7.934<br>Z-bar = 196.436 (p=0.000)<br>Z-bar tilde = 150.568 (p=0.000) | W-bar = 1.239<br>Z-bar = 6.781 (p=0.000)<br>Z-bar tilde = 2.140 (p=0.032) |
| Film | W-bar = 2.035<br>Z-bar = 32.716 (p=0.000)<br>Z-bar tilde = 14.561 (p=0.000) | W-bar = 1.859<br>Z-bar = 27.153 (p=0.000)<br>Z-bar tilde = 11.140 (p=0.000) |
| Wikipedia | W-bar = 2.098<br>Z-bar = 34.732 (p = 0.000)<br>Z-bar tilde = 7.978 (p = 0.000) | W-bar = 1.755<br>Z-bar = 23.885 (p=0.000)<br>Z-bar tilde = 3.404 (p=0.001) |

Note: The Dumitrescu & Hurlin (2012) Granger Non-Causality Test requires a minimum of 8 periods, so the patent case is excluded due to its limited time span.

Table S2. Predicting Creative Activities (New Concept Adoption) of Innovators with Spatial Dynamics and Subjective Perspective.

| | Model 1 **Science** | Model 2 **Technology** | Model 3 **Entrepreneurship** | Model 4 **Film** | Model 5 **Wikipedia** |
|---|---|---|---|---|---|
| Concept Movement | 0.106 | $0.178^{***}$ | -0.069 | $0.566^{***}$ | -0.055 |
| | (0.120) | (0.080) | (0.074) | (0.078) | (0.059) |
| Concept Visibility | $0.006^{***}$ | $0.005^{***}$ | $0.006^{***}$ | $0.002^{**}$ | 0.000 |
| | (0.000) | (0.001) | (0.000) | (0.000) | (0.000) |
| Movement × Visibility | $0.024^{**}$ | $0.013^{*}$ | $0.019^{**}$ | 0.001 | 0.002 |
| | (0.005) | (0.004) | (0.006) | (0.002) | (0.002) |
| Control: Concept Original Distance | $1.260^{***}$ | $1.284^{***}$ | $0.183^{***}$ | $-0.679^{**}$ | $0.041^{**}$ |
| | (0.151) | (0.044) | (0.015) | (0.028) | (0.012) |
| Author Fixed Effects | YES | YES | YES | YES | YES |
| Period Fixed Effects | YES | YES | YES | YES | YES |
| Adjusted $R^2$ | 0.125 | 0.122 | 0.021 | 0.092 | 0.32 |
| Observation *(N)* | 67,022,618 | 115,173,030 | 10,044,500 | 2,611,918 | 18,040,106 |

$^{*}$ $p < 0.05$, $^{**}$ $p < 0.01$, $^{***}$ $p < 0.001$. Standard errors clustered at discipline/tech field/industry level.

Table S3. Regression.

| | Model 1 | Model 2 | Model 3 | Model 4 | Model 5 | Model 6 |
|---|---|---|---|---|---|---|
| | **Science** | | **Technology** | | **Entrepreneurship** | |
| | Negative Binomial Regression DV: 5Y Forward Citations | | Negative Binomial Regression DV: 5Y Forward Citations | | Negative Binomial Regression Total Rounds of Funding from VCs | |
| Log(Perspective Diversity) | 0.600$^{***}$ | 0.504$^{***}$ | 0.131$^{***}$ | 0.099$^{**}$ | 0.371$^{***}$ | 0.438$^{***}$ |
| | (0.132) | (0.129) | (0.037) | (0.033) | (0.079) | (0.071) |
| Log(Background Diversity) | -0.916$^{***}$ | -1.107$^{***}$ | -0.800$^{***}$ | -0.916$^{***}$ | -0.998$^{***}$ | -1.268$^{***}$ |
| | (0.239) | (0.226) | (0.061) | (0.135) | (0.098) | (0.081) |
| Log(Mean_Experience) | 0.304$^{***}$ | 0.289$^{***}$ | 0.105 | 0.106 | 0.039$^{***}$ | 0.063$^{***}$ |
| | (0.022) | (0.020) | (0.058) | (0.059) | (0.007) | (0.006) |
| Previous Collaboration | | -0.339$^{***}$ | | -0.204$^{***}$ | | -0.152$^{***}$ |
| | | (0.050) | | (0.059) | | (0.011) |
| Proportion of New Innovators | | -0.654$^{***}$ | | -0.091 | | -0.032$^{*}$ |
| | | (0.041) | | (0.071) | | (0.016) |
| Constant | 1.297$^{***}$ | 1.674$^{***}$ | 0.312 | 0.028 | 0.329$^{***}$ | 0.242 |
| | (0.092) | (0.118) | (0.327) | (0.355) | (0.060) | (0.171) |
| Ln(Alpha) | 0.445$^{***}$ | 0.438$^{***}$ | 0.766$^{***}$ | 0.705$^{***}$ | -2.890$^{***}$ | -3.016$^{***}$ |
| | (0.060) | (0.060) | (0.080) | (0.088) | (0.166) | (0.169) |
| Team Size FE | YES | YES | YES | YES | YES | YES |
| Year FE | YES | YES | YES | YES | YES | YES |
| Discipline/Field/Industry FE | YES | YES | YES | YES | YES | YES |
| Other Controls | | | | Claims, Family, References | | Location, Demographics |
| Pseudo R2 | 0.017 | 0.018 | 0.017 | 0.027 | 0.044 | 0.050 |
| Observation *(N)* | 20405492 | 20405492 | 2524918 | 2524918 | 10180 | 10180 |

$^{*}$ $p < 0.05$, $^{**}$ $p < 0.01$, $^{***}$ $p < 0.001$. Standard errors clustered at discipline/tech field/industry level.

Table S4. Regression (continuing the table above).

| | Model 7 | Model 8 | Model 9 | Model 10 | Model 11 | Model 12 |
|---|---|---|---|---|---|---|
| | **Film**<br>OLS<br>DV: IMDB Rating | | **Wikipedia: First Record**<br>OLS<br>DV: Wiki Page Quality Grading | | **Wikipedia: Multi-Year Records**<br>OLS with High Dimensional FE (reghdfe)<br>DV: Wiki Page Quality Grading | |
| Log(Perspective Diversity) | 0.973$^{***}$ | 1.081$^{***}$ | 0.627$^{***}$ | 0.603$^{***}$ | 0.313$^{***}$ | 0.318$^{***}$ |
| | (0.090) | (0.096) | (0.001) | (0.001) | (0.001) | (0.001) |
| Log(Background Diversity) | -1.490$^{***}$ | -1.711$^{***}$ | -2.129$^{***}$ | -2.139$^{***}$ | -1.117$^{***}$ | -1.127$^{***}$ |
| | (0.198) | (0.217) | (0.003) | (0.003) | (0.004) | (0.004) |
| Log(Mean_Experience) | -0.025$^{*}$ | -0.024$^{**}$ | -0.021$^{***}$ | -0.021$^{***}$ | -0.003$^{***}$ | -0.005$^{***}$ |
| | (0.009) | (0.008) | (0.000) | (0.000) | (0.000) | (0.000) |
| Previous Collaboration | | 0.044 | | -0.037$^{***}$ | | 0.005$^{***}$ |
| | | (0.027) | | (0.001) | | (0.000) |
| Proportion of New Innovators | | -0.024 | | -0.172$^{***}$ | | -0.144$^{***}$ |
| | | (0.055) | | (0.002) | | (0.001) |
| Constant | 6.088$^{***}$ | 7.947$^{***}$ | 1.750$^{***}$ | 1.781$^{***}$ | 1.374$^{***}$ | 1.415$^{***}$ |
| | (0.085) | (0.342) | (0.003) | (0.003) | (0.003) | (0.003) |
| Team Size FE | YES | YES | YES | YES | YES | YES |
| Year FE | YES | YES | YES | YES | YES | YES |
| Genre/Page ID FE | YES | YES | | | YES | YES |
| Other Controls | | Country, Language | | | | |
| Adjusted R2 | 0.128 | 0.215 | 0.227 | 0.228 | 0.708 | 0.709 |
| Observation *(N)* | 29797 | 29797 | 5058522 | 5058522 | 9676033 | 9676033 |

$^{*}$ $p < 0.05$, $^{**}$ $p < 0.01$, $^{***}$ $p < 0.001$. Standard errors clustered at genre/page level.

Table S5. Marginal Diversity Contribution Influences Team Roles for Scientists (One vs. One, Using Direct Sup Role as Baseline).

| | Model 1 | Model 2 | Model 3 | Model 4 | Model 5 | Model 6 | Model 7 | Model 8 |
|---|---|---|---|---|---|---|---|---|
| | **DV: Probability of Most Important Roles In Scientific Collaboration** | | | | | | | |
| | **Leading Role vs. Direct Sup Role** | | | | **Indirect Sup Role vs. Direct Sup Role** | | | |
| Marginal Perspective Diversity Contribution (Proportion) | 0.030*** | 0.090*** | 0.040*** | 0.041*** | -0.070*** | -0.009 | -0.023* | -0.017 |
| | (0.007) | (0.011) | (0.006) | (0.006) | (0.009) | (0.007) | (0.010) | (0.010) |
| Marginal Background Diversity Contribution (Proportion) | -0.098*** | -0.133*** | -0.063*** | -0.070*** | 0.047*** | -0.010 | 0.018* | 0.009 |
| | (0.008) | (0.013) | (0.005) | (0.005) | (0.004) | (0.005) | (0.007) | (0.006) |
| Log(Team Perspective Diversity) | 0.078*** | 0.092*** | -0.003 | 0.005 | -0.067*** | -0.049*** | -0.070*** | -0.070*** |
| | (0.009) | (0.008) | (0.009) | (0.008) | (0.006) | (0.005) | (0.015) | (0.016) |
| Log(Team Background Diversity) | -0.235** | -0.170* | 0.077*** | 0.118*** | 0.076* | 0.133** | 0.120 | 0.160* |
| | (0.076) | (0.062) | (0.016) | (0.019) | (0.032) | (0.037) | (0.059) | (0.062) |
| Log(Mean_Experience) | | 0.046*** | | 0.021*** | | 0.033*** | | 0.008*** |
| | | (0.002) | | (0.000) | | (0.002) | | (0.000) |
| Previous Collaboration | | -0.041*** | | 0.027*** | | -0.016* | | 0.015 |
| | | (0.005) | | (0.005) | | (0.007) | | (0.008) |
| Proportion of New Innovators | | 0.043*** | | 0.074*** | | 0.041*** | | -0.003 |
| | | (0.003) | | (0.005) | | (0.008) | | (0.020) |
| Constant | 0.644*** | 0.401*** | 0.701*** | 0.538*** | 0.216*** | 0.051*** | 0.245*** | 0.190*** |
| | (0.005) | (0.015) | (0.002) | (0.004) | (0.002) | (0.004) | (0.010) | (0.016) |
| **Innovator-Individual FE** | | | YES | YES | | | YES | YES |
| Team Size FE | YES | YES | YES | YES | YES | YES | YES | YES |
| Year FE | YES | YES | YES | YES | YES | YES | YES | YES |
| Discipline FE | YES | YES | YES | YES | YES | YES | YES | YES |
| Adjusted $R^2$ | 0.090 | 0.155 | 0.312 | 0.314 | 0.015 | 0.065 | 0.262 | 0.263 |
| Observation *(N)* | 480111 | 480111 | 223538 | 223538 | 215294 | 215294 | 62960 | 62960 |

* $p < 0.05$, ** $p < 0.01$, *** $p < 0.001$. Standard errors clustered at discipline level.

Table S6. Regression for Knowledge Integration.

| | Model 1 | Model 2 | Model 3 | Model 4 | Model 5 | Model 6 | Model 7 | Model 8 | Model 9 | Model 10 |
|---|---|---|---|---|---|---|---|---|---|---|
| | **Science** | | **Technology** | | **Entrepreneurship** | | **Film** | | **Wikipedia (Multi-Records)** | |
| Dependent Variable | Knowledge Integration | 5Y Forward Citations | Knowledge Integration | 5Y Forward Citations | Knowledge Integration | Total Rounds Funding | Knowledge Integration | IMDB Rating | Knowledge Integration | Wiki Page Quality Grading |
| Log(Perspective Diversity) | **0.172**$^{***}$ | 0.427$^{***}$ | **0.280**$^{***}$ | 0.067 | **0.329**$^{***}$ | 0.234$^{***}$ | **0.120**$^{***}$ | 0.972$^{***}$ | **0.063**$^{***}$ | 0.314$^{***}$ |
| | **(0.017)** | (0.105) | **(0.011)** | (0.042) | **(0.037)** | (0.050) | **(0.009)** | (0.081) | **(0.001)** | (0.001) |
| Log(Background Diversity) | **-0.634**$^{***}$ | -0.811$^{***}$ | **-0.902**$^{***}$ | -0.805$^{***}$ | **-1.945**$^{***}$ | -0.006 | **-0.405**$^{***}$ | -1.510$^{***}$ | **-0.409**$^{***}$ | -1.114$^{***}$ |
| | **(0.013)** | (0.165) | **(0.024)** | (0.107) | **(0.219)** | (0.128) | **(0.026)** | (0.180) | **(0.002)** | (0.005) |
| Log(Mean_Experience) | 0.065$^{***}$ | 0.256$^{***}$ | 0.043$^{***}$ | 0.100 | 0.053$^{***}$ | 0.031$^{***}$ | 0.127$^{***}$ | -0.066$^{***}$ | 0.021$^{***}$ | -0.007$^{***}$ |
| | (0.003) | (0.014) | (0.004) | (0.057) | (0.002) | (0.007) | (0.003) | (0.013) | (0.000) | (0.000) |
| Previous Collaboration | 0.192$^{***}$ | -0.429$^{***}$ | 0.324$^{***}$ | -0.244$^{***}$ | 0.158$^{***}$ | -0.297$^{***}$ | -0.012$^{**}$ | 0.047 | 0.208$^{***}$ | -0.011$^{***}$ |
| | (0.011) | (0.046) | (0.011) | (0.071) | (0.006) | (0.010) | (0.003) | (0.026) | (0.000) | (0.001) |
| Proportion of New Innovators | -0.198$^{***}$ | -0.562$^{***}$ | -0.477$^{***}$ | -0.030 | -0.036 | 0.006 | -0.554$^{***}$ | 0.147$^{*}$ | -0.259$^{***}$ | -0.125$^{***}$ |
| | (0.013) | (0.054) | (0.020) | (0.049) | (0.017) | (0.008) | (0.006) | (0.065) | (0.001) | (0.001) |
| **Knowledge Integration** | | **0.484**$^{**}$ | | **0.127**$^{*}$ | | **0.709**$^{***}$ | | **0.329**$^{**}$ | | **0.077**$^{***}$ |
| | | **(0.150)** | | **(0.055)** | | **(0.015)** | | **(0.114)** | | **(0.001)** |
| Constant | 0.006 | 1.665$^{***}$ | 0.066 | 0.028 | 0.343$^{*}$ | -0.024 | 0.683$^{***}$ | 7.577$^{***}$ | 0.348$^{***}$ | 1.379$^{***}$ |
| | (0.016) | (0.110) | (0.151) | (0.363) | (0.129) | (0.162) | (0.021) | (0.270) | (0.002) | (0.003) |
| Team Size FE | YES | YES | YES | YES | YES | YES | YES | YES | YES | YES |
| Year FE | YES | YES | YES | YES | YES | YES | YES | YES | YES | YES |
| Discipline/Field/Industry/Genre/Page ID | YES | YES | YES | YES | YES | YES | YES | YES | YES | YES |
| Other Controls | | | Claims, Family, References | | Location, Demographics | | Country, Language | | | |
| Adjusted R2/Pseudo R2 | 0.341 | 0.019 | 0.522 | 0.027 | 0.294 | 0.066 | 0.675 | 0.214 | 0.506 | 0.706 |
| Observation *(N)* | 20405492 | 20405492 | 2524918 | 2524918 | 10181 | 10180 | 29786 | 29786 | 9433618 | 9433618 |

$^{*}$ $p < 0.05$, $^{**}$ $p < 0.01$, $^{***}$ $p < 0.001$. Standard errors clustered at discipline/tech field/industry/genre/page level.

Table S7. Regression for Knowledge Speculation.

| | Model 1 | Model 2 | Model 3 | Model 4 | Model 5 | Model 6 | Model 7 | Model 8 | Model 9 | Model 10 |
|---|---|---|---|---|---|---|---|---|---|---|
| | **Science** | | **Technology** | | **Entrepreneurship** | | **Film** | | **Wikipedia (Multi-Records)** | |
| Dependent Variable | Knowledge Speculation | 5Y Forward Citations | Knowledge Speculation | 5Y Forward Citations | Knowledge Speculation | Total Rounds Funding | Knowledge Speculation | IMDB Rating | Knowledge Speculation | Wiki Page Quality Grading |
| Log(Perspective Diversity) | **-0.294**$^{***}$ | 0.383$^{***}$ | **-0.313**$^{***}$ | 0.069 | **-0.221**$^{***}$ | 0.318$^{***}$ | **-0.163**$^{***}$ | 0.992$^{***}$ | **-0.138**$^{***}$ | 0.314$^{***}$ |
| | **(0.027)** | (0.098) | **(0.015)** | (0.035) | **(0.039)** | (0.051) | **(0.007)** | (0.087) | **(0.001)** | (0.001) |
| Log(Background Diversity) | **0.761**$^{***}$ | -0.804$^{***}$ | **0.930**$^{***}$ | -0.824$^{***}$ | **1.074**$^{***}$ | -0.668$^{***}$ | **0.069**$^{**}$ | -1.635$^{***}$ | **0.555**$^{***}$ | -1.127$^{***}$ |
| | **(0.040)** | (0.176) | **(0.035)** | (0.130) | **(0.176)** | (0.090) | **(0.020)** | (0.221) | **(0.003)** | (0.005) |
| Log(Mean_Experience) | -0.102$^{***}$ | 0.245$^{***}$ | -0.054$^{***}$ | 0.100 | -0.040$^{***}$ | 0.043$^{***}$ | -0.114$^{***}$ | -0.038$^{**}$ | -0.045$^{***}$ | -0.007$^{***}$ |
| | (0.005) | (0.014) | (0.004) | (0.059) | (0.003) | (0.005) | (0.005) | (0.011) | (0.000) | (0.000) |
| Previous Collaboration | -0.039$^{**}$ | -0.356$^{***}$ | -0.113$^{***}$ | -0.215$^{***}$ | 0.030$^{**}$ | -0.139$^{***}$ | 0.066$^{***}$ | 0.051 | -0.124$^{***}$ | 0.000 |
| | (0.012) | (0.045) | (0.009) | (0.059) | (0.007) | (0.011) | (0.003) | (0.026) | (0.000) | (0.001) |
| Proportion of New Innovators | 0.130$^{***}$ | -0.609$^{***}$ | 0.269$^{***}$ | -0.065 | -0.183$^{***}$ | -0.145$^{***}$ | 0.077$^{***}$ | -0.026 | 0.279$^{***}$ | -0.136$^{***}$ |
| | (0.016) | (0.045) | (0.009) | (0.066) | (0.010) | (0.012) | (0.009) | (0.058) | (0.001) | (0.001) |
| **Knowledge Speculation** | | **-0.428**$^{***}$ | | **-0.102**$^{***}$ | | **-0.707**$^{***}$ | | **-0.120** | | **-0.034**$^{***}$ |
| | | **(0.104)** | | **(0.025)** | | **(0.012)** | | **(0.074)** | | **(0.000)** |
| Constant | 0.938$^{***}$ | 2.065$^{***}$ | 0.754$^{***}$ | 0.111 | 0.294$^{**}$ | 0.422$^{*}$ | 0.263$^{***}$ | 7.833$^{***}$ | 0.872$^{***}$ | 1.435$^{***}$ |
| | (0.023) | (0.086) | (0.151) | (0.358) | (0.062) | (0.178) | (0.036) | (0.342) | (0.003) | (0.003) |
| Team Size FE | YES | YES | YES | YES | YES | YES | YES | YES | YES | YES |
| Year FE | YES | YES | YES | YES | YES | YES | YES | YES | YES | YES |
| Discipline/Field/Industry/Genre/Page ID | YES | YES | YES | YES | YES | YES | YES | YES | YES | YES |
| Other Controls | | | Claims, Family, References | | Location, Demographics | | Country, Language | | | |
| Adjusted R2/Pseudo R2 | 0.240 | 0.019 | 0.234 | 0.027 | 0.204 | 0.058 | 0.487 | 0.213 | 0.398 | 0.705 |
| Observation *(N)* | 20405492 | 20405492 | 2524918 | 2524918 | 10181 | 10180 | 29786 | 29786 | 9433618 | 9433618 |

$^{*}$ $p < 0.05$, $^{**}$ $p < 0.01$, $^{***}$ $p < 0.001$. Standard errors clustered at discipline/tech field/industry/genre/page level.

Table S8. Startup Case and Movie Case with Alternative Dependent Variables.

| | Model 1 | Model 2 | Model 3 | Model 4 | Model 5 | Model 6 | Model 7 | Model 8 |
|---|---|---|---|---|---|---|---|---|
| | **Entrepreneurship Case IPO or Acquisition (Dummy)** | | **Film Case Any Award (Dummy)** | | **Wikipedia Case (First Record) Page Views (Log Transformation)** | | **Wikipedia Case (Multi-Records) Page Views (Log Transformation)** | |
| Log(Perspective Diversity) | 0.553$^{***}$ | 0.543$^{***}$ | 2.657$^{***}$ | 2.947$^{***}$ | 1.816$^{***}$ | 1.599$^{***}$ | 0.784$^{***}$ | 0.827$^{***}$ |
| | (0.076) | (0.070) | (0.214) | (0.128) | (0.006) | (0.006) | (0.004) | (0.004) |
| Log(Background Diversity) | -2.154$^{***}$ | -2.350$^{***}$ | -3.479$^{***}$ | -4.276$^{***}$ | -1.919$^{***}$ | -2.007$^{***}$ | 0.140$^{***}$ | -0.092$^{***}$ |
| | (0.288) | (0.327) | (0.586) | (0.445) | (0.014) | (0.014) | (0.025) | (0.025) |
| Log(Mean_Experience) | 0.198$^{***}$ | 0.198$^{***}$ | -0.061$^{***}$ | -0.001 | -0.084$^{***}$ | -0.080$^{***}$ | 0.005$^{***}$ | -0.003$^{***}$ |
| | (0.029) | (0.020) | (0.010) | (0.020) | (0.001) | (0.001) | (0.000) | (0.000) |
| Previous Collaboration | | -0.161$^{*}$ | | -0.108$^{*}$ | | -0.282$^{***}$ | | -0.049$^{***}$ |
| | | (0.071) | | (0.054) | | (0.004) | | (0.002) |
| Proportion of New Innovators | | 2.494$^{***}$ | | -0.223$^{*}$ | | -0.752$^{***}$ | | -0.680$^{***}$ |
| | | (0.137) | | (0.095) | | (0.008) | | (0.004) |
| Constant | -2.354$^{***}$ | -0.796 | -2.382$^{***}$ | -2.722$^{***}$ | 6.797$^{***}$ | 6.176$^{***}$ | 8.520$^{***}$ | 7.926$^{***}$ |
| | (0.094) | (0.487) | (0.316) | (0.545) | (0.007) | (0.010) | (0.005) | (0.006) |
| Team Size FE | YES | YES | YES | YES | YES | YES | YES | YES |
| Year FE | YES | YES | YES | YES | YES | YES | YES | YES |
| Industry/Genre/Page FE | YES | YES | YES | YES | | | YES | YES |
| Other Controls | Location, Demographics | | Country, Language | | | | | |
| Adjusted R2/Pseudo R2 | 0.085 | 0.092 | 0.139 | 0.168 | 0.183 | 0.187 | 0.876 | 0.877 |
| Observation *(N)* | 10157 | 10137 | 29796 | 29710 | 2331545 | 2331545 | 5010990 | 5010990 |

$^{*}$ $p < 0.05$, $^{**}$ $p < 0.01$, $^{***}$ $p < 0.001$. Standard errors clustered at industry/genre level.

Table S9. Technology Case with Alternative Space (Science as the Underlying Space)

| | Model 1 | Model 2 | Model 3 | Model 4 | Model 5 | Model 6 |
|---|---|---|---|---|---|---|
| | Original Model | | Mechanism: Knowledge Integration | | Mechanism: Knowledge Speculation | |
| | **Citation** | **Citation** | **Knowledge Integration** | **Citation** | **Knowledge Speculation** | **Citation** |
| **Knowledge Integration** | | | | $0.161^{*}$ | | |
| | | | | (0.070) | | |
| **Knowledge Speculation** | | | | | | $-0.121^{***}$ |
| | | | | | | (0.032) |
| Log(Perspective Diversity) | 0.087 | 0.044 | $0.255^{***}$ | 0.007 | $-0.284^{***}$ | 0.012 |
| | (0.070) | (0.061) | (0.013) | (0.075) | (0.015) | (0.067) |
| Log(Background Diversity) | $-1.126^{***}$ | $-1.160^{***}$ | $-1.204^{***}$ | $-0.972^{***}$ | $1.249^{***}$ | $-1.012^{***}$ |
| | (0.128) | (0.196) | (0.058) | (0.160) | (0.073) | (0.187) |
| Log(Mean_Experience) | 0.100 | 0.093 | $0.038^{***}$ | 0.086 | $-0.050^{***}$ | 0.086 |
| | (0.054) | (0.053) | (0.004) | (0.051) | (0.004) | (0.052) |
| Previous Collaboration | | $-0.162^{**}$ | $0.348^{***}$ | $-0.215^{**}$ | $-0.135^{***}$ | $-0.177^{***}$ |
| | | (0.050) | (0.011) | (0.067) | (0.010) | (0.051) |
| Proportion of New Innovators | | -0.123 | $-0.489^{***}$ | -0.045 | $0.279^{***}$ | -0.092 |
| | | (0.092) | (0.019) | (0.060) | (0.008) | (0.085) |
| Constant | 0.452 | 0.128 | 0.052 | 0.127 | $0.766^{***}$ | 0.224 |
| | (0.456) | (0.481) | (0.149) | (0.486) | (0.150) | (0.482) |
| Ln(Alpha) | $0.837^{***}$ | $0.773^{***}$ | | $0.772^{***}$ | | $0.772^{***}$ |
| | (0.083) | (0.092) | | (0.092) | | (0.092) |
| Team Size FE | YES | YES | YES | YES | YES | YES |
| Year FE | YES | YES | YES | YES | YES | YES |
| Discipline/Field/Industry/Genre/Page ID FE | YES | YES | YES | YES | YES | YES |
| Other Controls | | YES | YES | YES | YES | YES |
| Adjusted R2/Pseudo R2 | 0.018 | 0.029 | 0.522 | 0.029 | 0.229 | 0.029 |
| Observation *(N)* | 2834716 | 2834716 | 2834716 | 2834716 | 2834716 | 2834716 |

$^{*}$ $p < 0.05$, $^{**}$ $p < 0.01$, $^{***}$ $p < 0.001$. Standard errors clustered at industry/genre level.

Table S10. Regression with Team Centroid-Task Distance Controlled.

| | Model 1<br>**Science (Citation)** | Model 2<br>**Technology (Citation)** | Model 3<br>**Entrepreneurship (Fundings)** | Model 4<br>**Film (Rating)** | Model 5<br>**Wikipedia (First-record; Page Quality Grading)** | Model 6<br>**Wikipedia (Multi-records; Page Quality Grading)** |
|---|---|---|---|---|---|---|
| | Negative Binomial Regression | | | | OLS | |
| Log(Perspective Diversity) | 0.210$^{***}$ | 0.001 | 0.221$^{**}$ | 0.575$^{***}$ | 0.206$^{***}$ | 0.143$^{***}$ |
| | (0.057) | (0.054) | (0.075) | (0.078) | (0.001) | (0.001) |
| Log(Background Diversity) | -0.521$^{***}$ | -0.664$^{***}$ | -0.770$^{***}$ | -1.051$^{***}$ | -1.594$^{***}$ | -0.944$^{***}$ |
| | (0.062) | (0.099) | (0.099) | (0.164) | (0.003) | (0.004) |
| Log(Mean_Experience) | 0.288$^{***}$ | 0.106 | 0.046$^{***}$ | -0.045$^{***}$ | -0.018$^{***}$ | -0.005$^{***}$ |
| | (0.019) | (0.059) | (0.008) | (0.008) | (0.000) | (0.000) |
| Previous Collaboration | -0.347$^{***}$ | -0.212$^{***}$ | -0.133$^{***}$ | 0.040 | -0.022$^{***}$ | 0.002$^{***}$ |
| | (0.050) | (0.060) | (0.010) | (0.027) | (0.001) | (0.000) |
| Proportion of New Innovators | -0.638$^{***}$ | -0.078 | -0.029$^{*}$ | -0.038 | -0.058$^{***}$ | -0.072$^{***}$ |
| | (0.042) | (0.071) | (0.014) | (0.057) | (0.001) | (0.001) |
| Constant | 1.005$^{***}$ | -0.237 | -0.286 | 7.089$^{***}$ | 0.399$^{***}$ | 0.417$^{***}$ |
| | (0.235) | (0.513) | (0.188) | (0.296) | (0.004) | (0.005) |
| Ln(Alpha) | 0.437$^{***}$ | 0.705$^{***}$ | -3.002$^{***}$ | | | |
| | (0.060) | (0.088) | (0.167) | | | |
| **Team Centroid-Task Dist** | 0.810$^{***}$ | 0.421$^{***}$ | 0.662$^{***}$ | 0.825$^{***}$ | 1.623$^{***}$ | 1.128$^{***}$ |
| | (0.210) | (0.111) | (0.111) | (0.083) | (0.003) | (0.005) |
| Team Size FE | YES | YES | YES | YES | YES | YES |
| Year FE | YES | YES | YES | YES | YES | YES |
| Discipline/Field/Industry/Genre/Page ID FE | YES | YES | YES | YES | YES | YES |
| Other Controls | | Claims, Family, References | Location, Demographics | Country, Language | | |
| Adjusted R2/Pseudo R2 | 0.018 | 0.027 | 0.050 | 0.220 | 0.280 | 0.714 |
| Observation *(N)* | 20405492 | 2524921 | 10180 | 29796 | 5058522 | 9676033 |

$^{*}$ $p < 0.05$, $^{**}$ $p < 0.01$, $^{***}$ $p < 0.001$. Standard errors clustered at industry/genre level.

Table S11. Regressions for Experience Convergence Effects.

| | Model 1<br>**Science (Citation)** | Model 2<br>**Technology (Citation)** | Model 3<br>**Entrepreneurship (Fundings)** | Model 4<br>**Film (Rating)** | Model 5<br>**Wikipedia (First-record; Page Quality Grading)** | Model 6<br>**Wikipedia (Multi-records; Page Quality Grading)** |
|---|---|---|---|---|---|---|
| | Negative Binomial Regression | | | | OLS | |
| **Experience Convergence** | 0.525$^{***}$ | 0.320$^{***}$ | -0.009 | -0.153$^{***}$ | -0.286$^{***}$ | -0.048$^{***}$ |
| | (0.125) | (0.021) | (0.063) | (0.035) | (0.008) | (0.007) |
| Log(Perspective Diversity) | 0.500$^{***}$ | 0.098$^{**}$ | 0.438$^{***}$ | 1.014$^{***}$ | 0.602$^{***}$ | 0.318$^{***}$ |
| | (0.129) | (0.033) | (0.070) | (0.097) | (0.001) | (0.001) |
| Log(Background Diversity) | -1.116$^{***}$ | -0.932$^{***}$ | -1.268$^{***}$ | -1.646$^{***}$ | -2.135$^{***}$ | -1.128$^{***}$ |
| | (0.224) | (0.136) | (0.082) | (0.231) | (0.003) | (0.004) |
| Log(Mean_Experience) | 0.291$^{***}$ | 0.106 | 0.063$^{***}$ | -0.025$^{**}$ | -0.022$^{***}$ | -0.006$^{***}$ |
| | (0.019) | (0.059) | (0.006) | (0.008) | (0.000) | (0.000) |
| Previous Collaboration | -0.347$^{***}$ | -0.211$^{***}$ | -0.152$^{***}$ | 0.044 | -0.036$^{***}$ | 0.005$^{***}$ |
| | (0.052) | (0.059) | (0.011) | (0.026) | (0.001) | (0.000) |
| Proportion of New Innovators | -0.645$^{***}$ | -0.086 | -0.032 | -0.035 | -0.172$^{***}$ | -0.144$^{***}$ |
| | (0.040) | (0.071) | (0.017) | (0.059) | (0.002) | (0.001) |
| Constant | 1.666$^{***}$ | 0.022 | 0.242 | 7.807$^{***}$ | 1.782$^{***}$ | 1.416$^{***}$ |
| | (0.117) | (0.350) | (0.171) | (0.335) | (0.003) | (0.003) |
| Ln(Alpha) | 0.437$^{***}$ | 0.705$^{***}$ | -3.016$^{***}$ | | | |
| | (0.061) | (0.088) | (0.169) | | | |
| Team Size FE | YES | YES | YES | YES | YES | YES |
| Year FE | YES | YES | YES | YES | YES | YES |
| Discipline/Field/Industry/Genre/Page ID FE | YES | YES | YES | YES | YES | YES |
| Other Controls | | Claims, Family, References | Location, Demographics | Country, Language | | |
| Adjusted R2/Pseudo R2 | 0.018 | 0.027 | 0.050 | 0.213 | 0.229 | 0.709 |
| Observation *(N)* | 20405492 | 2524918 | 10180 | 29796 | 5058522 | 9676033 |

$^{*}$ $p < 0.05$, $^{**}$ $p < 0.01$, $^{***}$ $p < 0.001$. Standard errors clustered at industry/genre level.

Table S12. Regressions without Any New Innovator.

| | Model 1 **Science (Citation)** | Model 2 **Technology (Citation)** | Model 3 **Entrepreneurship (Fundings)** | Model 4 **Film (Rating)** | Model 5 **Wikipedia (First-record; Page Quality Grading)** | Model 6 **Wikipedia (Multi-records; Page Quality Grading)** |
|---|---|---|---|---|---|---|
| | Negative Binomial Regression | | | | OLS | |
| Log(Perspective Diversity) | 0.337** | 0.045 | 0.449*** | 1.130*** | 0.326*** | 0.190*** |
| | (0.116) | (0.039) | (0.068) | (0.083) | (0.002) | (0.002) |
| Log(Background Diversity) | -1.282*** | -0.963*** | -1.277*** | -1.876*** | -2.163*** | -1.501*** |
| | (0.268) | (0.149) | (0.086) | (0.189) | (0.003) | (0.009) |
| Log(Mean_Experience) | 0.308*** | 0.112 | 0.065*** | -0.040** | -0.014*** | -0.004*** |
| | (0.020) | (0.061) | (0.006) | (0.013) | (0.000) | (0.000) |
| Previous Collaboration | -0.347*** | -0.199*** | -0.153*** | 0.024 | -0.024*** | -0.002** |
| | (0.038) | (0.059) | (0.011) | (0.027) | (0.002) | (0.001) |
| Constant | 1.574*** | 0.188 | 0.203 | 8.054*** | 2.035*** | 1.615*** |
| | (0.100) | (0.349) | (0.163) | (0.453) | (0.006) | (0.009) |
| Ln(Alpha) | 0.404*** | 0.727*** | -2.993*** | | | |
| | (0.047) | (0.088) | (0.169) | | | |
| Team Size FE | YES | YES | YES | YES | YES | YES |
| Year FE | YES | YES | YES | YES | YES | YES |
| Discipline/Field/Industry/Genre/Page ID FE | YES | YES | YES | YES | YES | YES |
| Other Controls | | Claims, Family, References | Location, Demographics | Country, Language | | |
| Adjusted R2/Pseudo R2 | 0.016 | 0.027 | 0.050 | 0.216 | 0.217 | 0.727 |
| Observation *(N)* | 9613720 | 1690701 | 10052 | 19549 | 2267884 | 2386650 |

$^{*}$ $p < 0.05$, $^{**}$ $p < 0.01$, $^{***}$ $p < 0.001$. Standard errors clustered at industry/genre level.

Table S13. Regression with Team Experience SD Controlled.

| | Model 1 **Science (Full Sample)** | Model 2 **Science (Cont. Statement Sample)** | Model 2 **Technology (Citation)** | Model 3 **Entrepreneurship (Fundings)** | Model 4 **Film (Rating)** | Model 5 **Wikipedia (First-record; Page Quality Grading)** | Model 6 **Wikipedia (Multi-records; Page Quality Grading)** |
|---|---|---|---|---|---|---|---|
| | Negative Binomial Regression | | | | | OLS | |
| Log(Perspective Diversity) | 0.370*** | 0.504*** | 0.097** | 0.543*** | 1.079*** | 0.605*** | 0.319*** |
| | (0.031) | (0.128) | (0.034) | (0.069) | (0.094) | (0.001) | (0.001) |
| Log(Background Diversity) | -1.560*** | -1.105*** | -0.915*** | -0.804*** | -1.701*** | -2.137*** | -1.124*** |
| | (0.172) | (0.225) | (0.135) | (0.090) | (0.206) | (0.003) | (0.004) |
| Log(Mean_Experience) | 0.109*** | 0.321*** | 0.096 | 0.124*** | -0.005 | -0.020*** | -0.005*** |
| | (0.012) | (0.021) | (0.052) | (0.005) | (0.023) | (0.000) | (0.000) |
| SD(Experience) | 0.000 | -0.000*** | 0.000 | -0.001*** | -0.005 | -0.000*** | -0.000*** |
| | (0.000) | (0.000) | (0.000) | (0.000) | (0.004) | (0.000) | (0.000) |
| L Ratio | -0.190*** | | | | | | |
| | (0.052) | | | | | | |
| Previous Collaboration | -0.392*** | -0.342*** | -0.204*** | -0.112*** | 0.041 | -0.035*** | 0.004*** |
| | (0.033) | (0.050) | (0.058) | (0.012) | (0.029) | (0.001) | (0.000) |
| Proportion of New Innovators | -0.664*** | -0.629*** | -0.102 | 0.060* | -0.016 | -0.170*** | -0.142*** |
| | (0.059) | (0.044) | (0.076) | (0.026) | (0.052) | (0.002) | (0.001) |
| Constant | 4.285*** | 1.572*** | 0.029 | -0.013 | 7.952*** | 1.769*** | 1.410*** |
| | (0.077) | (0.127) | (0.356) | (0.170) | (0.347) | (0.003) | (0.003) |
| Ln(Alpha) | -0.186*** | 0.437*** | 0.705*** | -3.080*** | | | |
| | (0.030) | (0.060) | (0.088) | (0.159) | | | |
| Team Size FE | YES | YES | YES | YES | YES | YES | YES |
| Year FE | YES | YES | YES | YES | YES | YES | YES |
| Discipline/Field/Industry/Genre/Page ID FE | YES | YES | YES | YES | YES | YES | YES |
| Other Controls | | | YES | YES | YES | | |
| Adjusted R2/Pseudo R2 | 0.037 | 0.018 | 0.027 | 0.054 | 0.215 | 0.229 | 0.709 |
| Observation *(N)* | 76934 | 20405492 | 2524918 | 9711 | 29797 | 5058522 | 9676033 |

$^{*}$ $p < 0.05$, $^{**}$ $p < 0.01$, $^{***}$ $p < 0.001$. Standard errors clustered at industry/genre level.

Table S14. Validation of LLM Evaluators Using ICLR 2025 Reviewer Scores

| | Pairwise Comparison Accuracy | | |
|---|---|---|---|
| Model | Low vs. Mid | Mid vs. High | **Low vs. High** |
| GPT-4o | 60.7% | 58.1% | **69.8%** |
| GPT-4o-mini | 54.8% | 51.3% | **58.3%** |

Accuracy in selecting the paper with a higher average human review rating from pairs. Papers are categorized into three groups based on their average human review ratings: low-rating (below 5, below marginal acceptance), middle-rating (5 to less than 6, around marginal acceptance), and high-rating (6 or above, above marginal acceptance).

Table S15. LLM Experiment Results: 95% Bootstrap Confidence Intervals for All Performance Metrics

| LLM Agents | Condition | Scientific Scores | Writing Scores | Knowledge Integration | Knowledge Speculation | Distance to Assigned Topic |
|---|---|---|---|---|---|---|
| Gemma | P1B0 | [7.695, 8.133] | [7.610, 8.076] | [0.224, 0.259] | [0.631, 0.667] | [0.224, 0.250] |
| | P1B1 | [7.171, 7.590] | [7.571, 7.914] | [0.157, 0.179] | [0.714, 0.737] | [0.244, 0.275] |
| | P0B0 | [6.867, 7.314] | [6.876, 7.352] | [0.166, 0.187] | [0.673, 0.709] | [0.311, 0.392] |
| | P0B1 | [7.181, 7.676] | [7.105, 7.571] | [0.097, 0.119] | [0.718, 0.743] | [0.284, 0.336] |
| Mistral | P1B0 | [7.048, 7.533] | [7.543, 7.990] | [0.127, 0.193] | [0.621, 0.689] | [0.395, 0.467] |
| | P1B1 | [6.095, 6.752] | [6.124, 6.705] | [0.076, 0.122] | [0.685, 0.721] | [0.454, 0.587] |
| | P0B0 | [6.210, 6.829] | [6.771, 7.333] | [0.108, 0.160] | [0.612, 0.666] | [0.629, 0.762] |
| | P0B1 | [6.429, 6.971] | [7.010, 7.571] | [0.030, 0.065] | [0.664, 0.722] | [0.515, 0.651] |
| QwQ | P1B0 | [7.124, 7.619] | [7.419, 7.857] | [0.230, 0.261] | [0.643, 0.675] | [0.110, 0.141] |
| | P1B1 | [6.762, 7.334] | [6.676, 7.190] | [0.189, 0.217] | [0.676, 0.708] | [0.130, 0.154] |
| | P0B0 | [6.419, 6.905] | [7.190, 7.724] | [0.198, 0.228] | [0.620, 0.654] | [0.192, 0.222] |
| | P0B1 | [6.781, 7.352] | [6.848, 7.410] | [0.113, 0.139] | [0.700, 0.719] | [0.160, 0.189] |

**Supplementary References**